\documentclass{article}

\usepackage{microtype}
\usepackage{graphicx}
\usepackage{subcaption}
\usepackage{booktabs} 
\usepackage{eso-pic}

\usepackage{hyperref}

\usepackage[preprint]{icml2026}

\usepackage{amsmath}
\usepackage{amssymb}
\usepackage{mathtools}
\usepackage{amsthm}

\usepackage[capitalize,noabbrev]{cleveref}

\theoremstyle{plain}

\theoremstyle{definition}

\theoremstyle{remark}

\usepackage{enumitem}

\usepackage[textsize=tiny]{todonotes}
\usepackage{float}
\icmltitlerunning{Transient Emergence of Perceptual Geometry in LLM Representations}

\begin{document}
\AddToShipoutPictureFG*{
    \AtPageUpperLeft{
        \hspace*{0.44\paperwidth}
        \raisebox{-2.0cm}{
            \includegraphics[width=3.2cm]{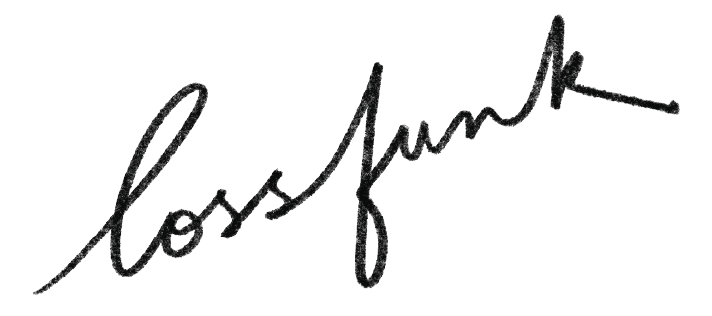}
        }
    }
}

\twocolumn[
  \icmltitle{Geometry of Human Perceptual Domains Emerges Transiently \\
  in LLM Representations}



\icmlsetsymbol{equal}{*}
  \begin{icmlauthorlist}
    \icmlauthor{Simardeep Singh}{equal,iitr}
    \icmlauthor{Paras Chopra}{equal,lossfunk}
  \end{icmlauthorlist}
  \icmlaffiliation{iitr}{Indian Institute of Technology Roorkee, Roorkee, India}
  \icmlaffiliation{lossfunk}{Lossfunk, India}
  \icmlcorrespondingauthor{Simardeep Singh}{simardeep\_s@mt.iitr.ac.in}
  \icmlcorrespondingauthor{Paras Chopra}{paras@lossfunk.com}

  \icmlkeywords{Machine Learning, ICML}

  \vskip 0.3in
]



\printAffiliationsAndNotice{}  

\begin{abstract}
While large language models (LLMs) are trained purely on textual data, prior work has shown that their internal representations can exhibit rich geometric structure in embedding space. Building on this line of work, we investigate whether such structure is similar to human perceptual organisation across different domains (e.g., color, pitch, emotion, and taste). Specifically, we study the layer-wise emergence of intrinsic geometrical structure corresponding to perceptual modalities within the residual streams of multiple open-weight transformer architectures. Our results reveal three key findings. First, we observe the emergence of layer-wise geometric structure across multiple perceptual domains, despite the absence of any direct perceptual supervision during training. Second, these perceptual domains exhibit distinct emergence profiles, with both geometric structure and its alignment with human baselines following domain- and model-specific trajectories across depth. Third, this emergence follows a consistent representational trajectory: geometry is weak or diffuse in early layers, becomes progressively organised in intermediate layers, and is attenuated in later layers, suggesting that perceptual geometry arises transiently as part of the model’s internal transformation pipeline. This provides new insight into how and where human-like perceptual geometry arises in LLMs, offering a principled pathway for mechanistic analysis of internal representations.

\end{abstract}

\section{Introduction}

LLMs have demonstrated remarkable capabilities in capturing relational geometric structures from purely textual data. Recent work has shown that LLM representations exhibit structured geometry across a range of concepts. For instance, cyclical domains such as days of the week, months, or hues are often organized along circular manifolds, reflecting their inherent periodic structure  \cite{engels2025, modell2025}. \cite{gurnee2024, gurnee2026} show that LLM representations corresponding to continuums like years of history or character counts are organized in compact 1D manifolds. A unifying explanation for these geometric patterns is provided by \cite{karkada2026}, who argue that such structures emerge naturally from translation symmetries in language co-occurrence statistics. Importantly, this phenomenon extends beyond simple conceptual domains to more complex scientific concepts. For example, representations of the periodic table have been shown to organize into a structured 3D spiral, where intermediate layers encode continuous relational properties and deeper layers sharpen categorical distinctions \cite{lei2025}.
\begin{figure}[H]
    \centering
    \includegraphics[width=0.70\linewidth]{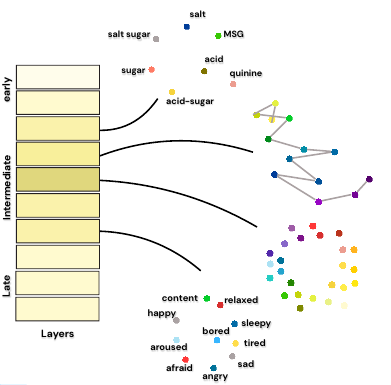}
    \caption{Perceptual geometric structures emerge within LLM representations across sensory and affective domains (taste, pitch, color, and emotion). Simpler structures (taste) peak earlier, followed progressively by pitch, color, and emotion.  Across domains, geometry is weak in early layers, organizes in intermediate layers, and attenuates in later layers.}
    \label{fig:overview}
\end{figure}
\begin{figure*}[t]
\centering
\setlength{\tabcolsep}{2pt}

\begin{tabular}{ccc}
    \includegraphics[width=0.25\textwidth,height=0.22\textwidth]{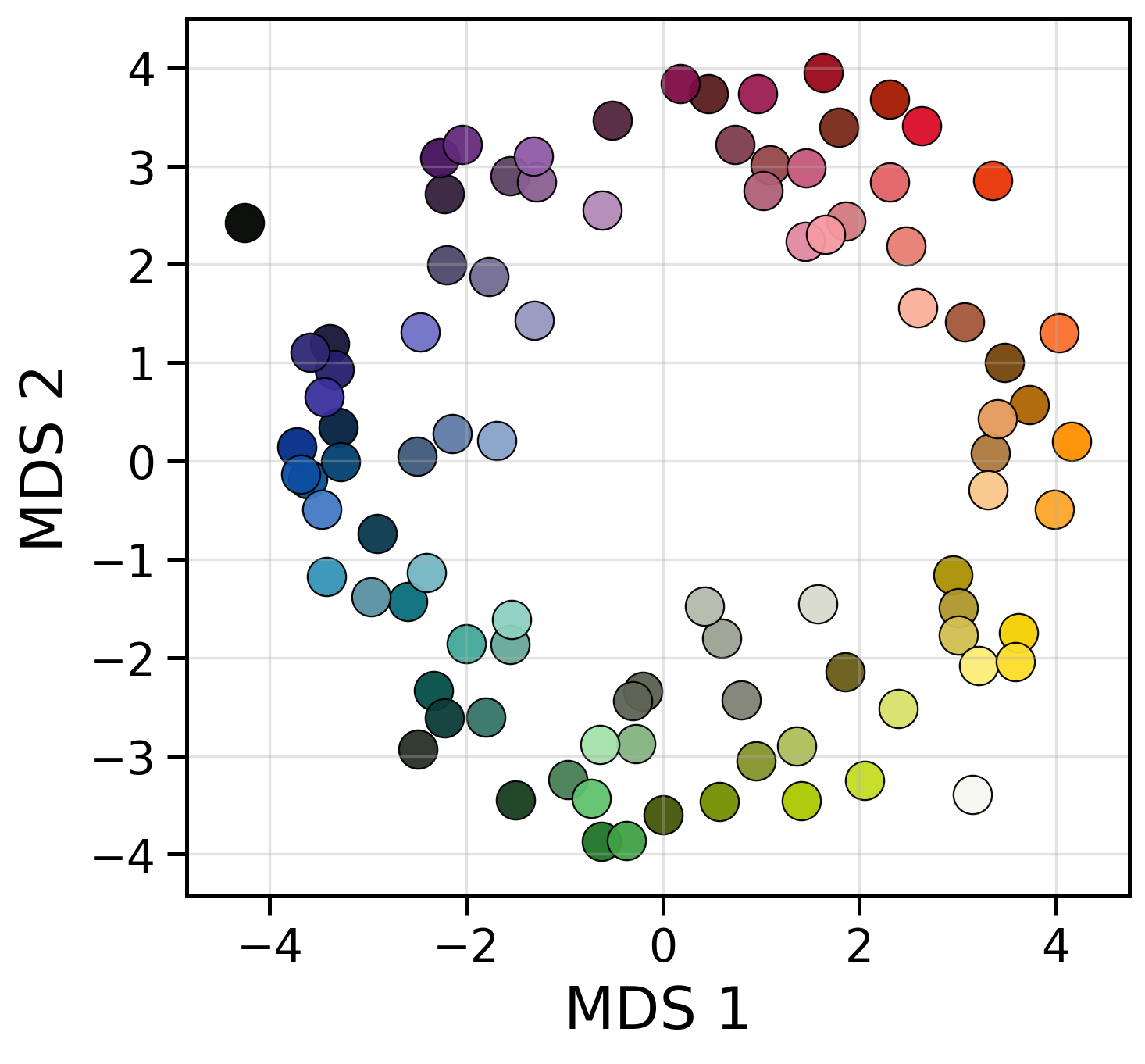} &
    \includegraphics[width=0.25\textwidth,height=0.22\textwidth]{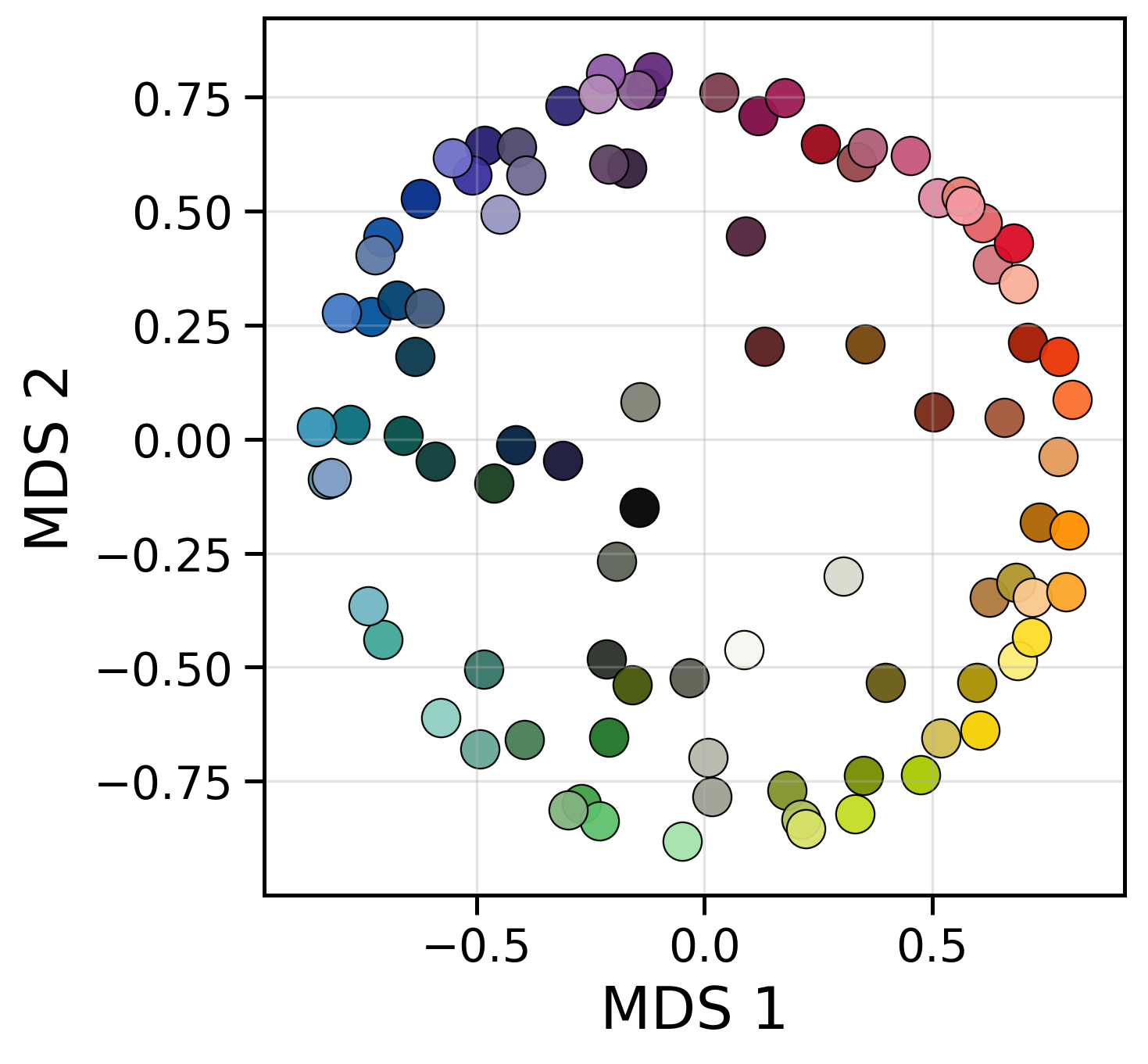} &
    \includegraphics[width=0.42\textwidth,height=0.18\textwidth]{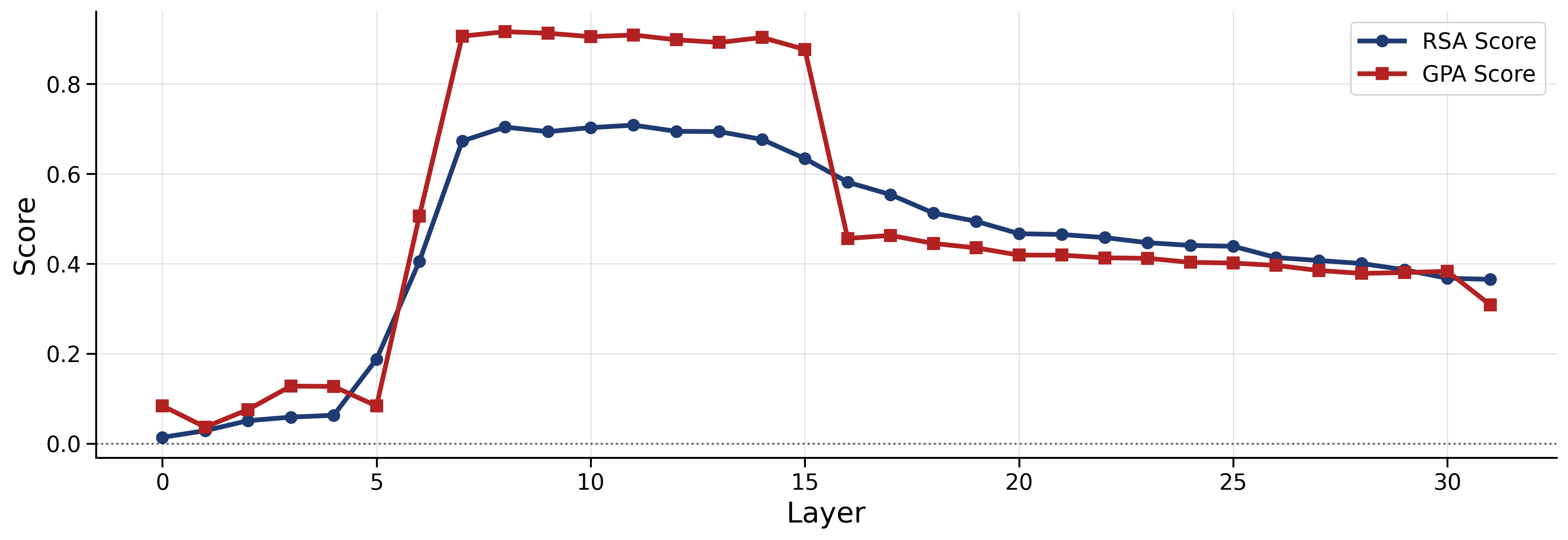} \\

    \includegraphics[width=0.30\textwidth,height=0.22\textwidth]{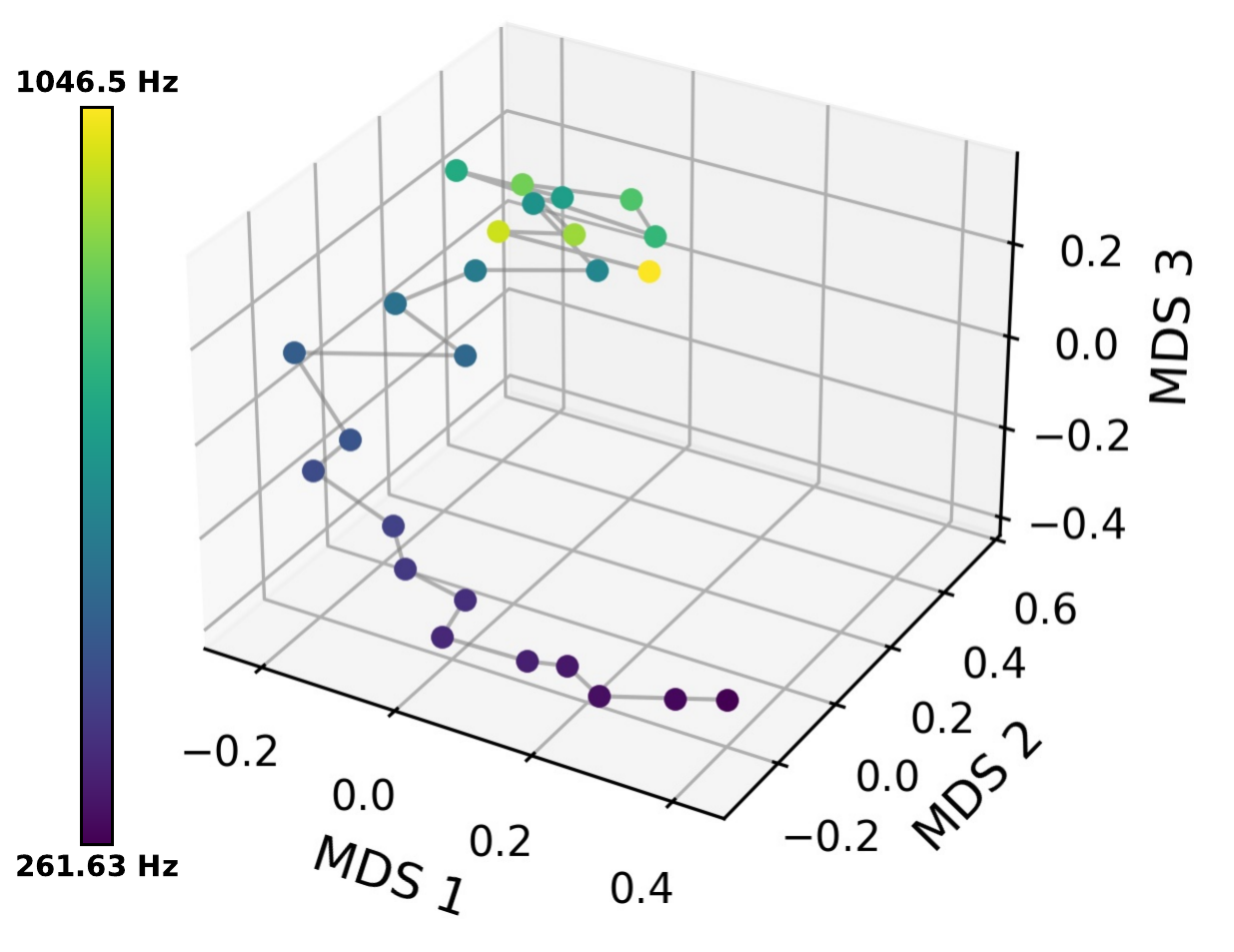} &
    \includegraphics[width=0.25\textwidth,height=0.22\textwidth]{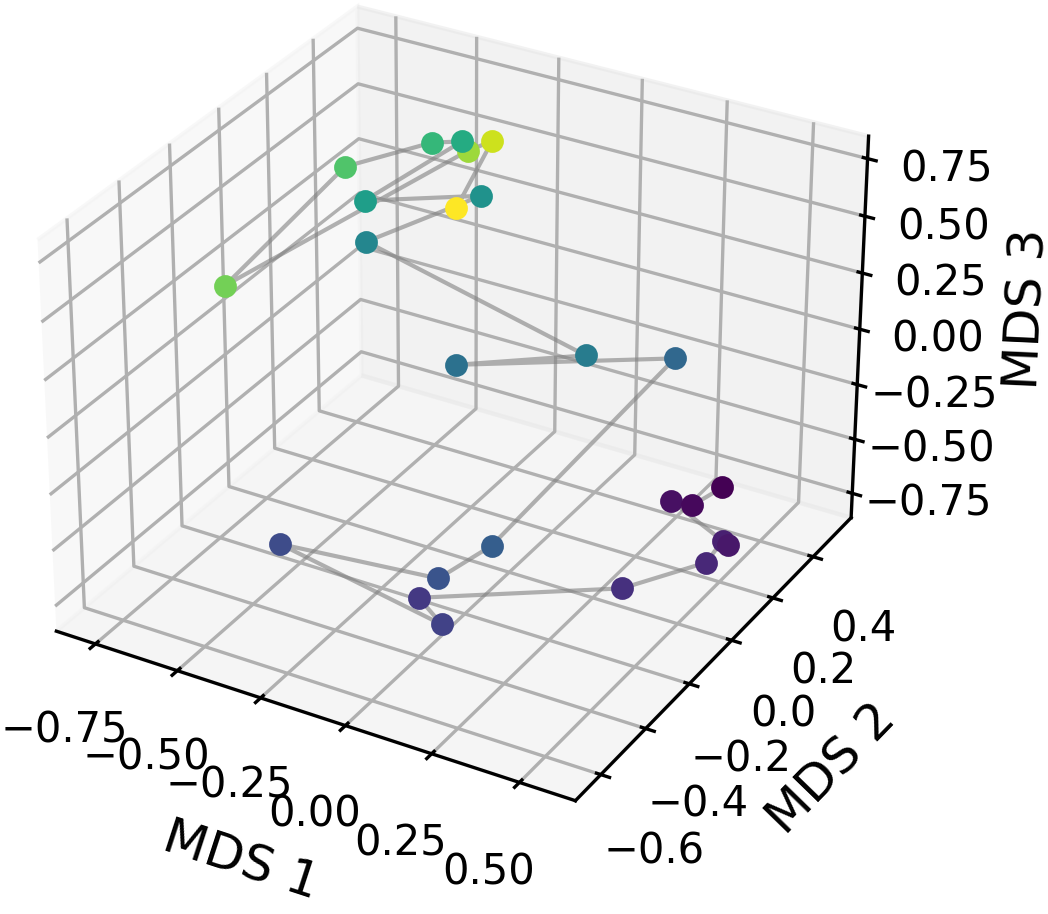} &
    \includegraphics[width=0.42\textwidth,height=0.18\textwidth]{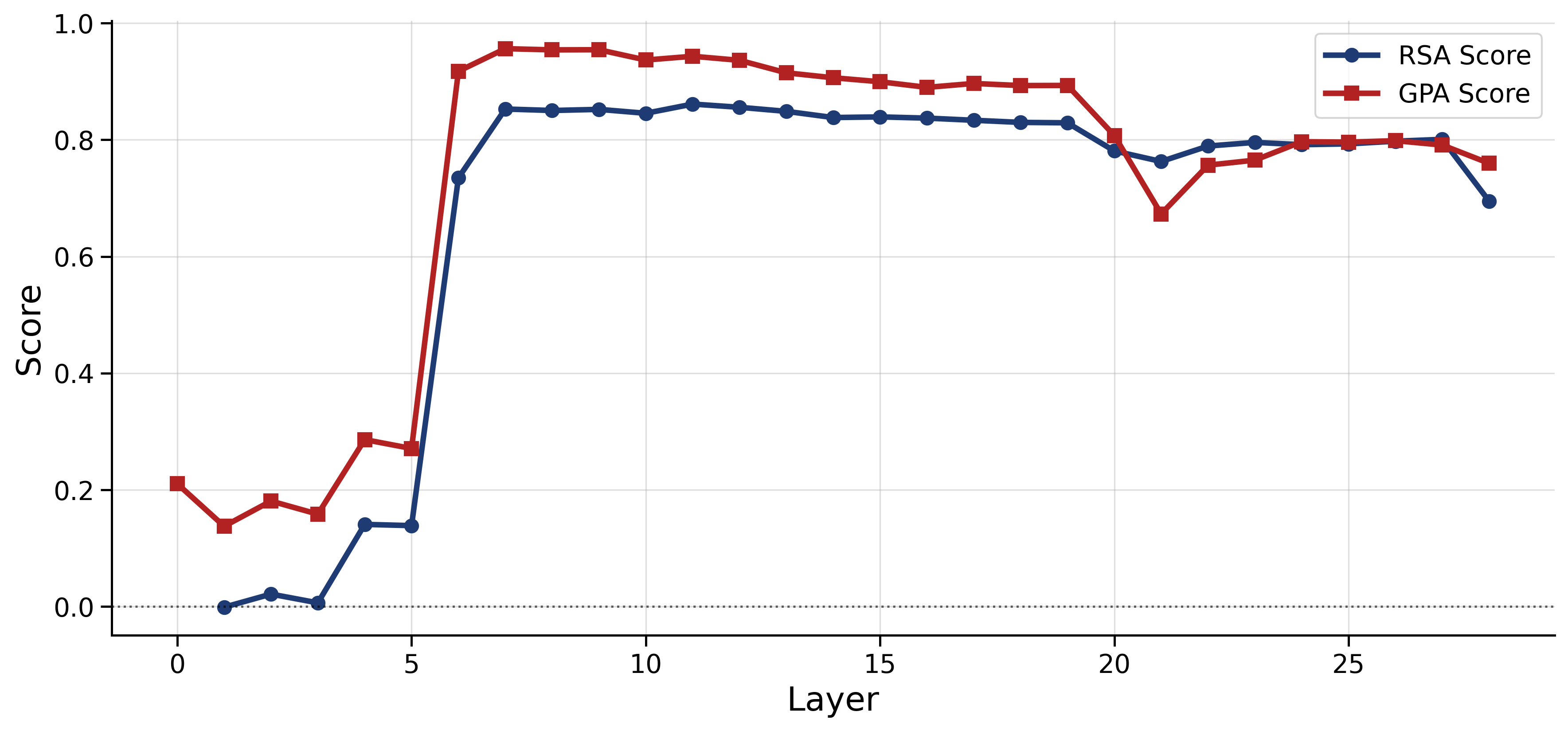} \\

    \includegraphics[width=0.25\textwidth,height=0.22\textwidth]{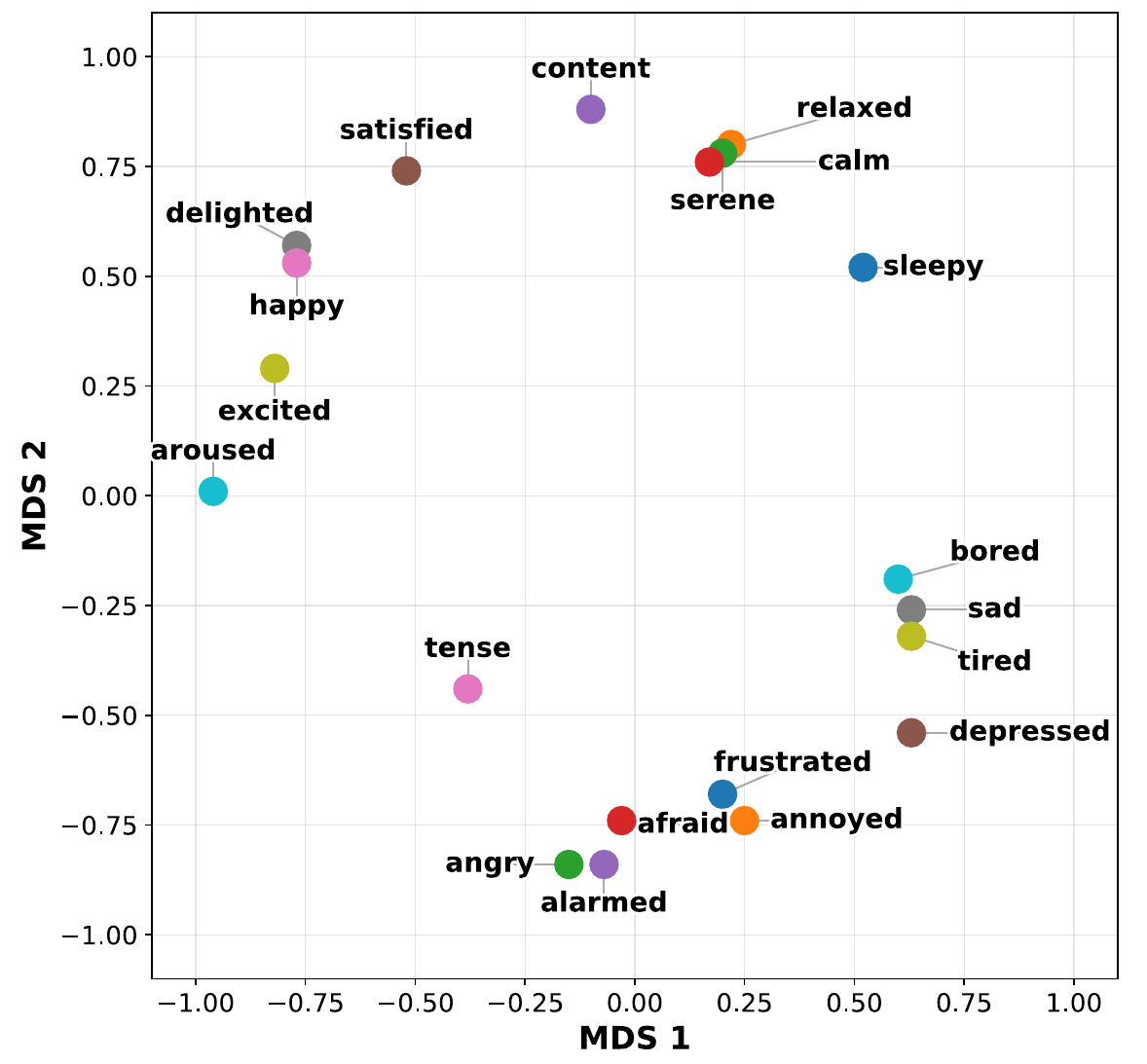} &
    \includegraphics[width=0.25\textwidth,height=0.22\textwidth]{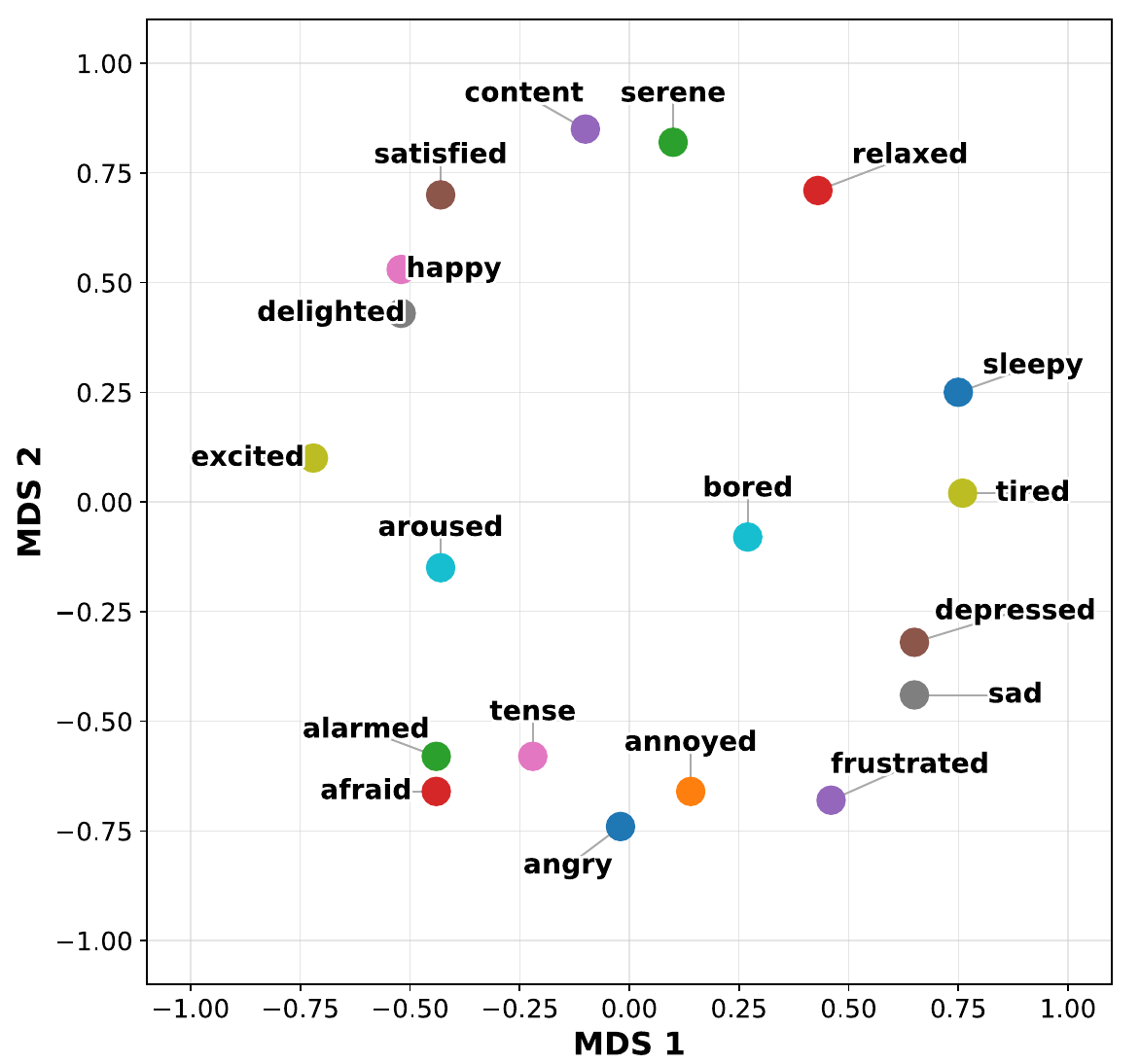} &
    \includegraphics[width=0.42\textwidth,height=0.18\textwidth]{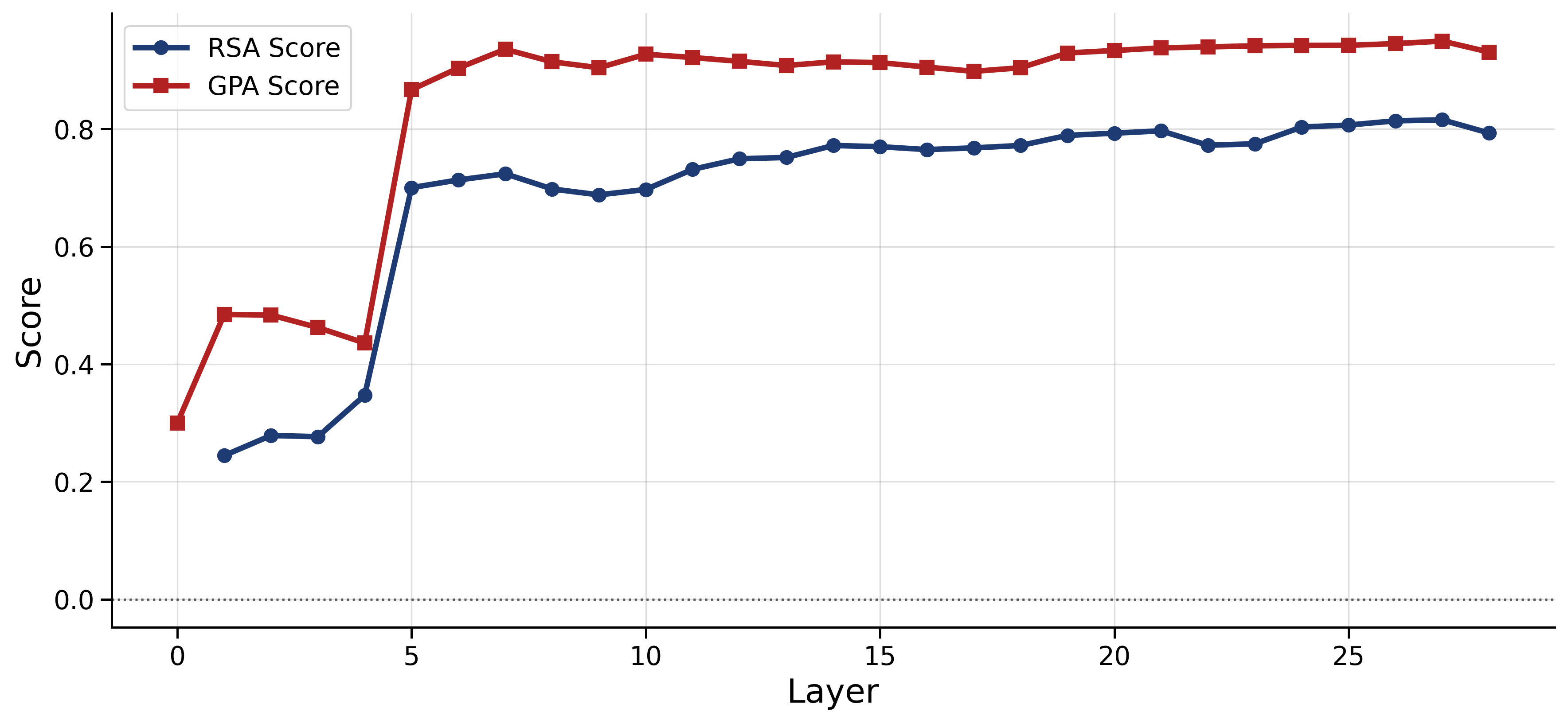} \\
    
\end{tabular}

\caption{
Overview of perceptual geometry emergence across modalities. Each row corresponds to a perceptual domain—color (top, LLaMA-3-8B), pitch (middle, Qwen-3-4B), and emotion (bottom, Gemma-7B), with the clearest representative model shown for each. Columns show the human perceptual baseline (left), the peak-alignment geometric representation from the model (middle), and the corresponding layer-wise alignment profile (right).
}
\label{fig:intro_overview}
\end{figure*}
Despite lacking direct sensory grounding, these models often exhibit behaviors that suggest an implicit understanding of perceptual and affective concepts such as color, pitch, and emotion. 
Prior work suggests that language itself may be closely linked to the organization of subjective experience, with internal representations in neural systems reflecting structured relationships between qualia and linguistic concepts \cite{taniguchi2025}. This raises a fundamental question: \textit{to what extent do internal representations in LLMs reflect human perceptual organization?}

Mechanistic interpretability provides a framework for addressing such questions by analyzing how information is represented and transformed across intermediate layers of neural networks \cite{singh2024, dar2023,alain2018} and has shown that transformer representations evolve across layers, with early layers encoding lexical features, and middle layers capturing abstract semantic relationships \cite{skean2024,skean2025}.

Emerging studies have begun to explore the relationship between language models and human perceptual structures. Evidence shows that LLMs can reproduce human sensory judgments across modalities and recover known perceptual structures such as the color wheel and pitch spiral through model outputs \cite{marjieh2023}. Similarly, \cite{abdou2021} demonstrates alignment between Language models and color spaces like CIELAB, highlighting that perceptual structure can, to some extent, be inferred from language. However, these approaches primarily operate either at the level of model outputs or task specific probing and do not directly examine how perceptual structure emerges and evolves across layers within the model.

We present three main contributions:
\begin{itemize}
\item  \textbf{Layer-wise emergence of perceptual geometry:} We demonstrate that LLMs develop structured geometric representations corresponding to multiple perceptual domains (color, pitch, taste, and emotion) within their residual streams, and quantitatively compare these representations with human perceptual baselines.
\item \textbf{Modality-specific emergence profiles:} Different perceptual domains exhibit distinct emergence profiles, with both geometric structure and its alignment with human perceptual baselines following domain- and model-specific trajectories across model depth.
\item \textbf{Transient perceptual encoding:} Perceptual structure follows a consistent depth-wise trajectory across models: it is weak or diffuse in early layers, becomes progressively organised in intermediate layers, and attenuates in later layers, indicating that perceptual geometry is encoded transiently rather than uniformly across depth.
\end{itemize}
\begin{figure}[H]
\centering
\setlength{\tabcolsep}{1.5pt}

\begin{tabular}{cc}
    \includegraphics[width=0.25
    \textwidth,height=0.21\textwidth]{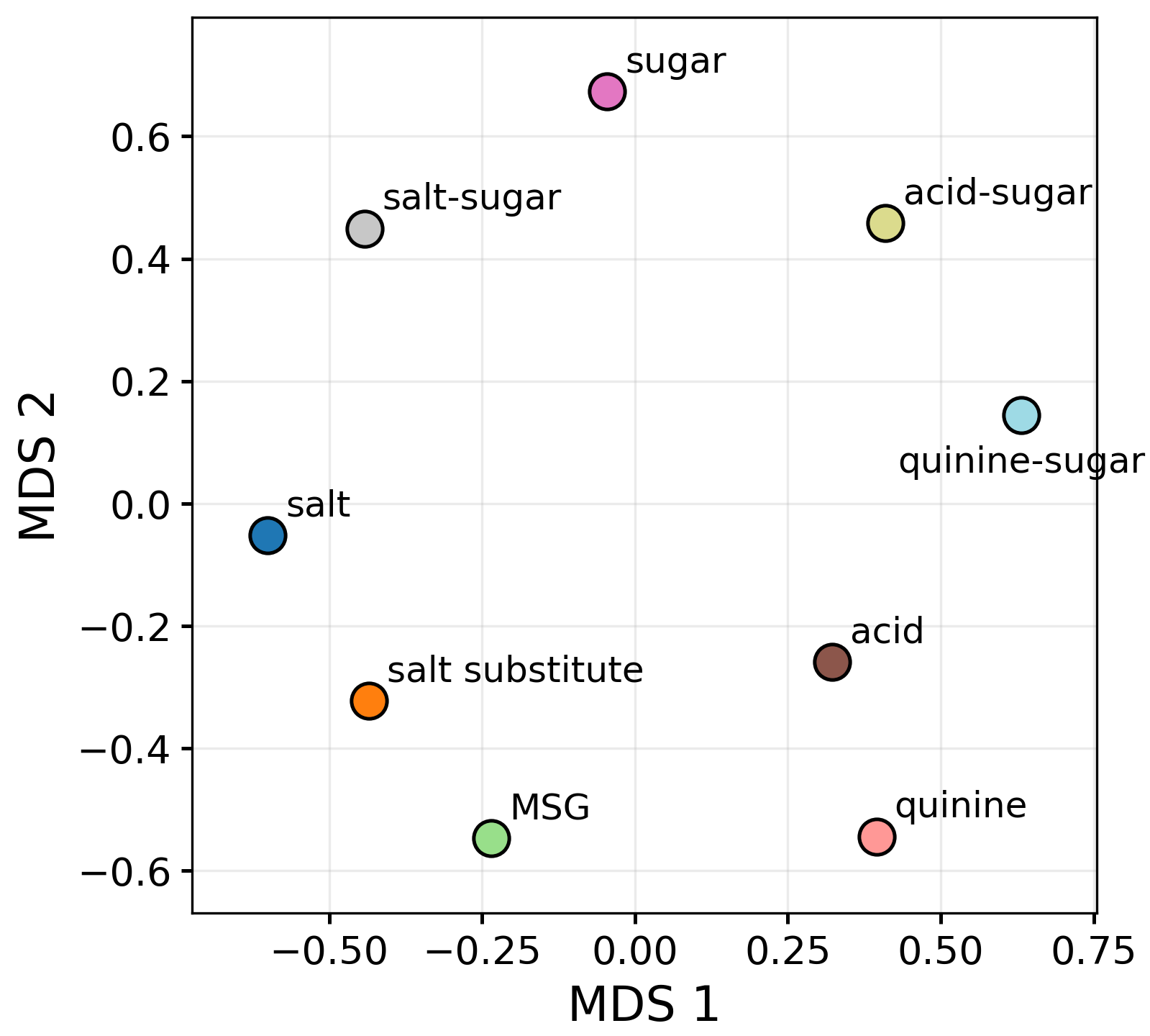} &
    \includegraphics[width=0.25\textwidth,height=0.22\textwidth]{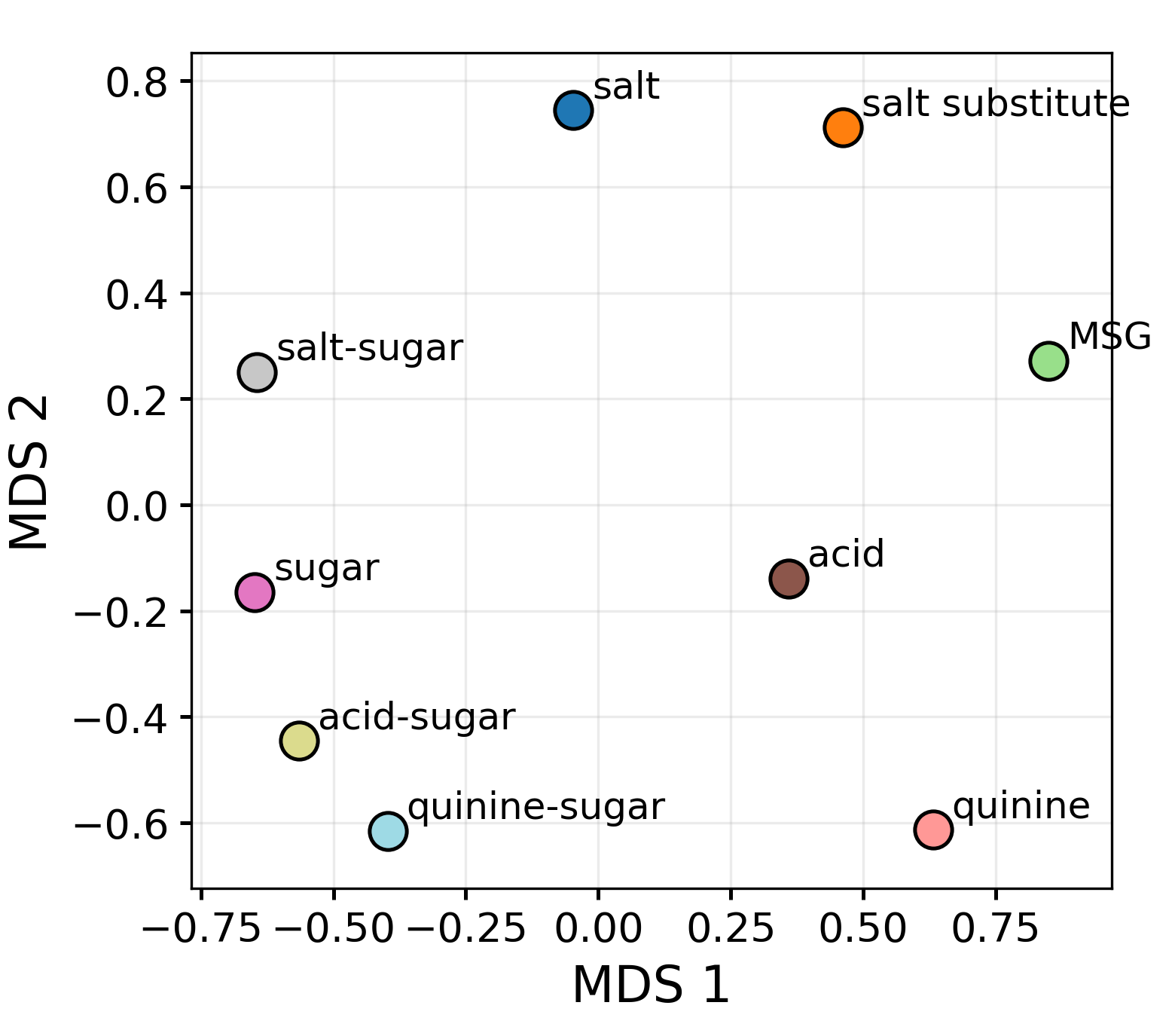} \\

    \multicolumn{2}{c}{
        \includegraphics[width=0.42\textwidth,height=0.18\textwidth]{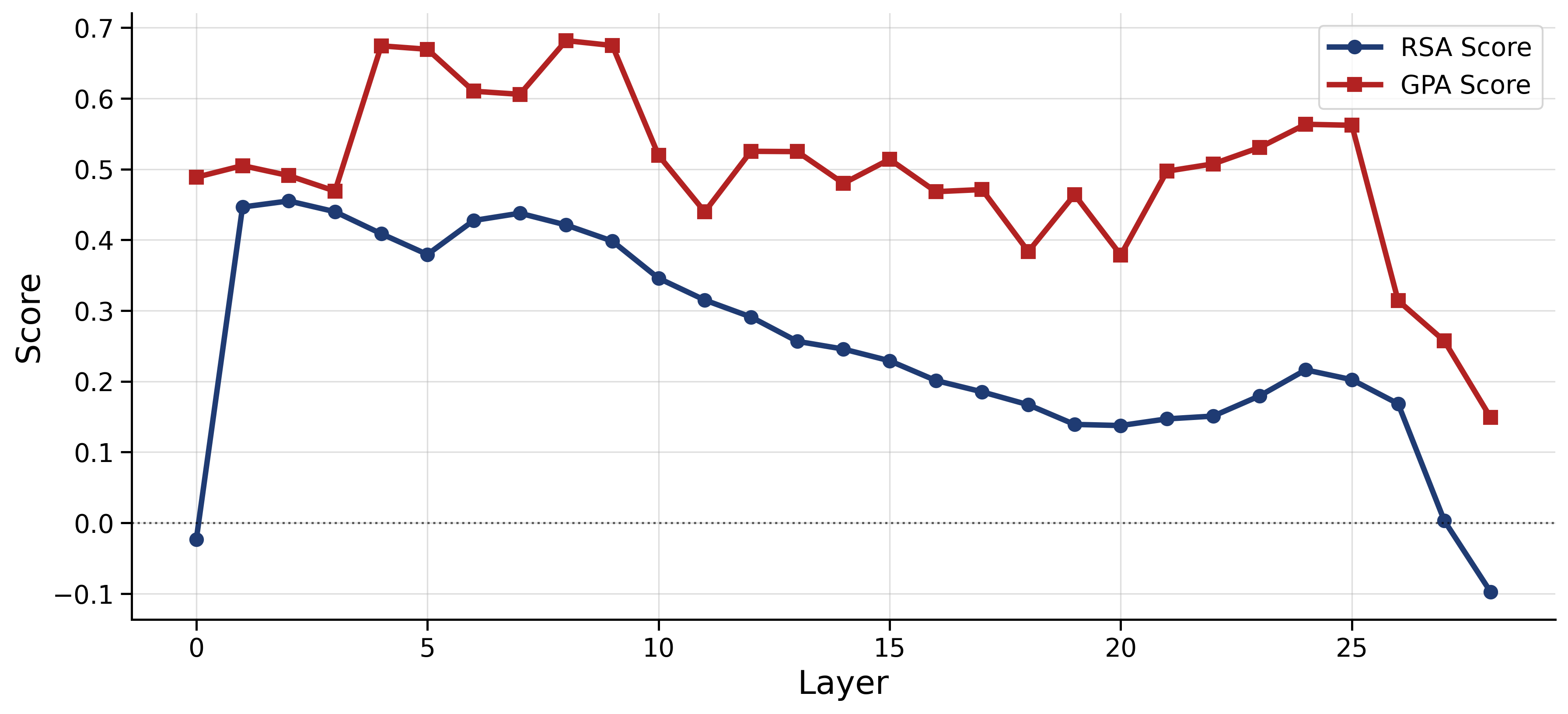}
    } \\
\end{tabular}

\caption{
Comparison of the human perceptual map (top left) and peak-layer LLM representation for taste in Gemma-7B (top right). Bottom: layer-wise alignment scores (RSA and GPA).
}
\label{fig:taste}
\end{figure}
\section{Methodology}

\subsection{Overview}

We investigate whether perceptual geometric structures exist within the embedding space of LLMs which aligns with the human similarity judgements. Our approach is fully \textit{intrinsic}, requiring no probing or additional training. Given a set of stimuli (e.g., colors, emotions, pitch values), we extract layer-wise representations from pretrained LLMs and construct geometric maps, which are then quantitatively compared with human perceptual baselines.

Formally, let a set of stimuli be denoted as $\{x_i\}_{i=1}^{N}$, where each stimulus is associated with a minimal structured prompt $P_i$ (see Appendix~\ref{app:prompts} and \ref{app:prompt-completions} for all prompt templates and their rationale). For example, for color stimuli, a prompt may take the form: \textit{``The description of color given as \texttt{\#9B081A}''}.  Each prompt is passed through a model $m$ to obtain last token hidden state activation at every transformer layer $l$, $\mathbf{h}_i^{(l,m)} \in \mathbb{R}^d$.

Then, for each layer $l$, we construct a geometric representation of the stimulus set through the following steps:

\textbf{Dissimilarity Computation}: Given embeddings $\{\mathbf{h}_i\}_{i=1}^{N}$, we compute pairwise cosine dissimilarities:
\begin{equation}
d_{ij} = 1 - \frac{\mathbf{h}_i \cdot \mathbf{h}_j}{\|\mathbf{h}_i\|\|\mathbf{h}_j\|}.
\end{equation}

This gives a symmetric dissimilarity matrix $\mathbf{D} \in \mathbb{R}^{N \times N}$. Cosine distance isolates directional structure and is invariant to scale differences across layers.

\textbf{Multidimensional Scaling (MDS)}: To obtain a low-dimensional geometric embedding, we MDS \cite{Kruskal1964}. Given $\mathbf{D}$, MDS finds coordinates $\{\mathbf{y}_i \in \mathbb{R}^2\}$ by minimizing the following stress function:
\begin{equation}
\min_{\{\mathbf{y}_i\}} \sum_{i < j} \left( d_{ij} - \|\mathbf{y}_i - \mathbf{y}_j\| \right)^2.
\end{equation}

This produces a 2D representation that preserves pairwise dissimilarities, forming a qualia map for each layer. These model representations were compared against human similarity judgments (see Appendix~\ref{app:data} for dataset details). The resulting human dissimilarity matrices from the datasets were projected into a low-dimensional space using MDS to construct the baseline perceptual geometry.

For quantifying the alignment between LLM and human representations we use the following metrics:

\textbf{Representational Similarity Analysis (RSA)}: RSA measures the correlation between two dissimilarity matrices. Let $\mathbf{D}^{\text{LLM}}$ and $\mathbf{D}^{\text{human}}$ denote the model and human dissimilarity matrices. We compute RSA as the Spearman rank correlation between the vectorized upper-triangular entries of $\mathbf{D}^{\text{LLM}}$ and $\mathbf{D}^{\text{human}}$.

\textbf{Generalized Procrustes Analysis (GPA)}: GPA measures geometric alignment between two geometric maps by finding the optimal orthogonal transformation:
\begin{equation}
\min_{\mathbf{R}} \|\mathbf{Y}^{\text{LLM}}\mathbf{R} - \mathbf{Y}^{\text{human}}\|_F^2.
\end{equation}
The final score is computed as $\text{GPA} = 1 - \frac{\text{residual}}{2}$. Higher values indicate stronger geometric alignment.

\subsection{Layer-wise Profiling}

We apply the above pipeline across all layers $l \in \{0, \dots, L\}$ for each model $m$, producing a sequence of geometric maps $\{\mathbf{Y}^{(l,m)}\}$. Each layer is independently compared with the human baseline using RSA and GPA, yielding a layer-wise alignment profile. This allows us to trace how perceptual structure evolves across depth rather than treating representation quality as a single endpoint. By evaluating each layer independently, we can identify where perceptual organisation first becomes measurable, how it strengthens through intermediate processing, and whether it is preserved or transformed in later layers.

\section{Results and Discussion}
We analyse the emergence of perceptual structure across four modalities: color, emotion, pitch, and taste. For each modality, we examine both qualitative geometry (via MDS maps) and quantitative alignment (via RSA and GPA) across layers. Figure~\ref{fig:intro_overview} summarizes the central result for three representative modalities—color, emotion, and pitch by showing the human perceptual geometry, the peak-alignment model geometry, and the corresponding layer-wise alignment profile. Full layer-wise geometric trajectories, including early and late-layer maps as well as complementary Isomap visualizations, are provided in Appendix~\ref{subsec:Layerwise}.

For color, the peak-layer representation in Figure~\ref{fig:intro_overview} exhibits a clear circular manifold closely resembling the human perceptual color wheel, indicating that the model recovers a smooth chromatic geometry from purely linguistic supervision. This stage corresponds to peak RSA and GPA values, where the model reorganizes raw lexical statistics into a structured geometry without any explicit supervision. The corresponding layer-wise profile shows that this structure emerges gradually, peaks in the early-middle layers, and weakens thereafter. This depth-wise pattern is broadly consistent across architectures (Figure~\ref{fig:appendix_color_metrics}): LLaMA-3-8B, LLaMA-3.2-3B, and Gemma-7B each exhibit a clear rise–peak–fall profile, where color alignment peaks at an intermediate layer followed by late-layer  attenuation. Early and last layer maps (Appendix~\ref{subsec:Layerwise}) show that this intermediate organization is preceded by fragmented local structure and followed by geometric degradation in later layers as the model increasingly specialize for task specific representations. Qwen-3-4B follows the same overall trajectory but with a shifted late-layer profile: after the intermediate peak, alignment briefly rebounds in the deeper layers before degrading again toward the final layers. Taken together, these results suggest that perceptual color structure is transient, emerging most coherently in intermediate representations before weakening in later layers.

For emotion, the peak-layer representation recovers a well-organized affective manifold aligned with the human valence--arousal structure \cite{russell}. In contrast to color, the associated layer-wise profile shows that this alignment is not only strong at its peak but remains comparatively stable across later layers, indicating more persistent retention of emotional geometry through depth. Our observation of a low-dimensional affective space aligns with concurrent findings by \cite{choi2026,sun2026}; however, our analysis additionally localizes where this structure becomes most coherent and how its stability differs from other perceptual domains. Detailed layer-wise maps (Figures~\ref{fig:appendix_emo_maps_2d_mds} and~\ref{fig:appendix_emo_maps_2d_iso}) further illustrate this progressive organization and relative preservation.

For pitch, the peak-layer geometry reveals a smooth arc-like organization consistent with the continuous and ordinal nature of human pitch perception. The layer-wise profile shows a more localized emergence pattern in which relational structure becomes coherent in intermediate layers before progressively deforming at greater depth. The full sequence in Figures~\ref{fig:appendix_pitch_maps_2d_gemma} and~\ref{fig:appendix_pitch_maps_3d_qwen} shows that early layers exhibit only weak partial ordering, intermediate layers undergo a clear structural transition in which frequencies become arranged along a arc-like manifold, and later layers progressively deform this relational organization.
The absence of discrete clusters and the emergence of a smooth trajectory suggest that pitch is represented in a relational, continuous manner rather than as a set of categorical groupings.

Taste representations also recover a clearly organized geometry that closely matches the human perceptual structure (Figure~\ref{fig:taste}). The peak-layer map recovers a qualitatively well-formed taste manifold, with the relative ordering of primary tastes and mixtures broadly matching the human perceptual arrangement (Figures~\ref{fig:appendix_taste_maps_2d_mds},~\ref{fig:appendix_taste_maps_2d_isomap}). This is reflected in strong geometric alignment (High GPA scores) at peak layers, indicating that the model captures the overall global structure of taste relations. However, taste differs from the other perceptual domains in its lower RSA scores and noisier layer-wise profiles (Figures~\ref{fig:appendix_taste_metrics}) , indicating that this structure is less stable and less precise at the level of fine-grained pairwise relations. The layer-wise maps and metric profiles show that taste geometry emerges early, but degrades rapidly thereafter, with a markedly noisier and less stable trajectory across subsequent layers than observed in the other perceptual domains.

\section{Limitations}

Our analysis is primarily descriptive: while we identify where perceptual geometry emerges and how it evolves across depth, we do not explain the mechanisms that give rise to these manifolds. This includes architecture-specific deviations—such as the late layer rebound observed in Qwen’s color trajectory, which we can localize descriptively but do not yet explain mechanistically. Understanding why such structure forms remains an important direction for future mechanistic work. Second, our study is limited to four perceptual domains. Although the observed trends are broadly consistent, broader evaluation across domains, architectures, and prompting conditions is needed to establish generality. We further restrict evaluation to RSA and GPA; while these capture complementary aspects of representational geometry, other metrics may reveal additional aspects of organization. Finally, geometric alignment with human baselines should not be interpreted as evidence of human-like perception, but only as evidence of structurally similar representational organization.


\bibliography{main_paper}
\bibliographystyle{icml2026}

\newpage
\appendix
\onecolumn
\section{Appendix}

\subsection{Prompts Used}
\label{app:prompts}

We use a set of minimal, structured prompts to extract representations corresponding to each domain. The list of prompts used across modalities is provided below:

\textbf{Color:} ``The description of the color given as \texttt{\#HEXCODE}'' \\
\textbf{Emotion:} ``Describe the person who is feeling \texttt{emotion}'' \\
\textbf{Taste:} ``The description of taste on tongue given by \texttt{taste}'' \\
\textbf{Pitch:} ``The description of sound of musical tone \texttt{VALUE Hz}''

We intentionally employ minimal and structurally simple prompts to avoid injecting additional semantic bias or task-specific reasoning into the representations. This ensures that the extracted embeddings primarily reflect the model’s intrinsic encoding of the stimulus rather than artifacts introduced by external or additional linguistic context.

\subsection{Prompt Completion Examples}
\label{app:prompt-completions}

To complement the minimal prompt templates described above, we provide representative prompt-completion examples generated by the LLMs for each modality. Each example contains the queried prompt for a given stimulus followed by the top descriptor phrases generated by Qwen3-8B, LLaMA-3-8B, and Gemma-7B. These examples verify that the prompts consistently elicit coherent and modality-relevant descriptions despite their intentionally minimal structure. They also ground the embedding analysis by showing the natural language completions from which latent representations are extracted, demonstrating that the resulting embeddings arise from semantically meaningful and stimulus-consistent textual descriptions rather than arbitrary vector artifacts.

\begin{center}
    \setlength{\tabcolsep}{1.5pt}
    \begin{tabular}{cc}
        \includegraphics[width=0.50\textwidth,height=0.55\textwidth]{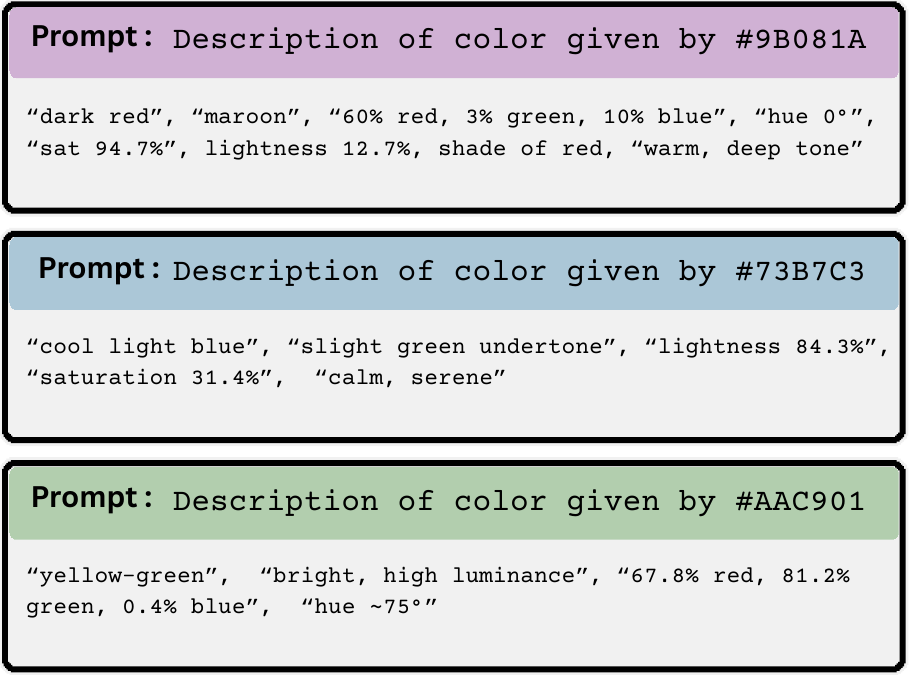} &
        \includegraphics[width=0.50\textwidth,height=0.55\textwidth]{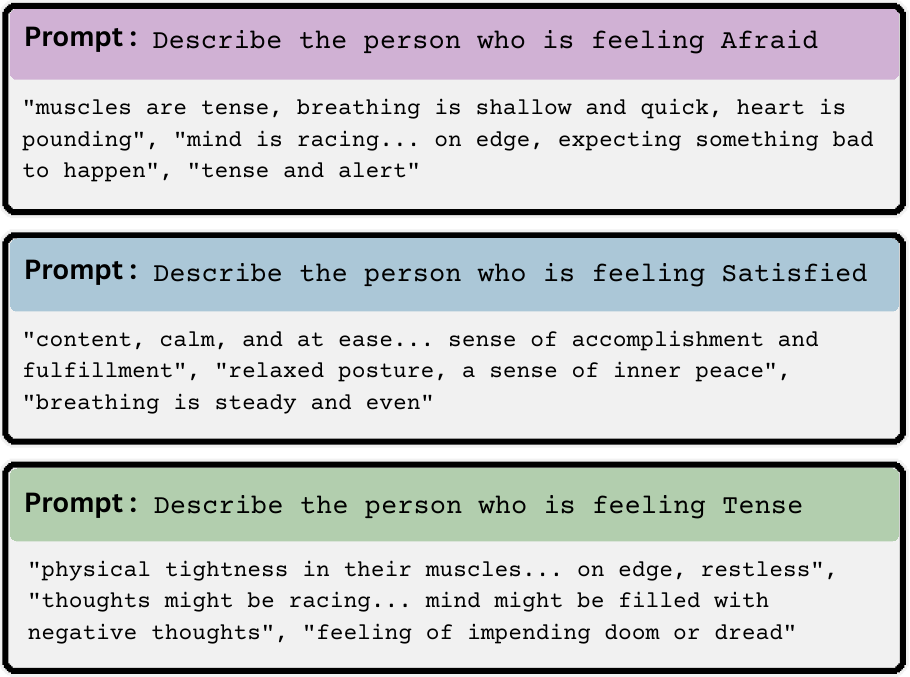}
    \end{tabular}

    \vspace{0.3em}
    \captionof{figure}{Representative prompt completions for color and emotion. Each panel shows the queried prompt together with the top descriptor phrases produced by Qwen3-8B, LLaMA-3-8B, and Gemma-7B.}
    \label{fig:appendix_prompt_pitch_taste}
\end{center}

\begin{center}
    \setlength{\tabcolsep}{1.5pt}
    \begin{tabular}{cc}
        \includegraphics[width=0.50\textwidth,height=0.55\textwidth]{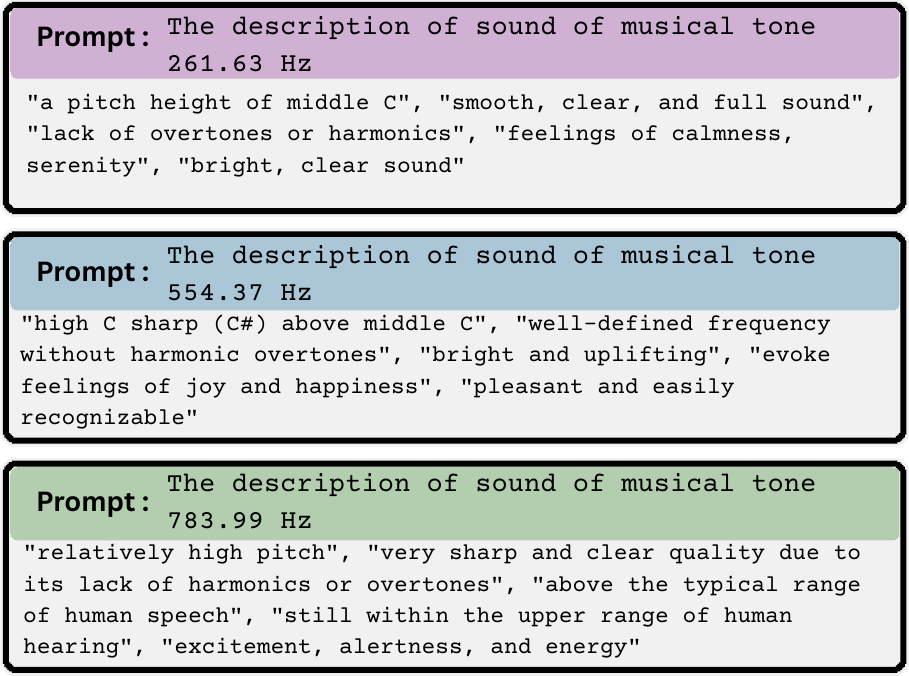} &
        \includegraphics[width=0.50\textwidth,height=0.55\textwidth]{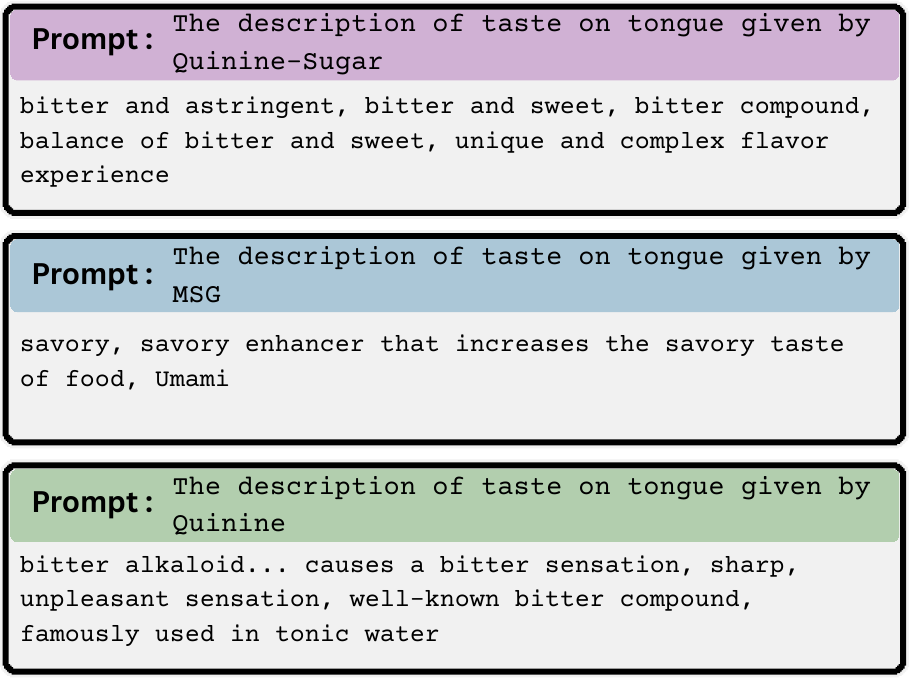}
    \end{tabular}

    \vspace{0.3em}
    \captionof{figure}{Representative prompt completions for pitch and taste. Each panel shows the queried prompt together with the top descriptor phrases produced by Qwen3-8B, LLaMA-3-8B, and Gemma-7B.}
    \label{fig:appendix_prompt_pitch_taste}
\end{center}

\subsection{Human Baseline Dataset Collection}
\label{app:data}

The datasets used in this work were collected from established sources of human perceptual similarity judgments. For color, pitch, and taste, we use datasets compiled in \cite{marjieh2023}, which aggregate similarity judgments and confusion matrices from multiple psychological studies and provide normalized dissimilarity matrices. For emotion, we use the ANEW benchmark database \cite{bradley1999affective}, which provides human ratings along valence, arousal, and dominance (VAD) dimensions. These coordinates are converted into pairwise dissimilarities using cosine distance in the VAD space to construct the baseline dissimilarity matrix. For each modality, the resulting dissimilarity matrices are projected into low-dimensional space using MDS to obtain the corresponding human baseline geometry.

\section{Additional Results}

\subsection{Layer-wise Geometric Emergence (MDS and Isomap)}
\label{subsec:Layerwise}

While MDS provides a faithful reconstruction of pairwise dissimilarities, it is important to verify that observed structures are not artifacts of the embedding method. Therefore, we additionally employ Isomap, a nonlinear dimensionality reduction technique that preserves geodesic distances on the underlying manifold. Consistency between MDS and Isomap visualizations provides stronger evidence that the observed geometric structures reflect intrinsic properties of the representation space rather than projection artifacts.

\begin{center}
    \setlength{\tabcolsep}{2pt}
    \begin{tabular}{cc}
        \includegraphics[width=0.28\textwidth,height=0.28\textwidth]{Figures/Color/Color-main/updated-llama8b/human.png} &
        \includegraphics[width=0.28\textwidth,height=0.28\textwidth]{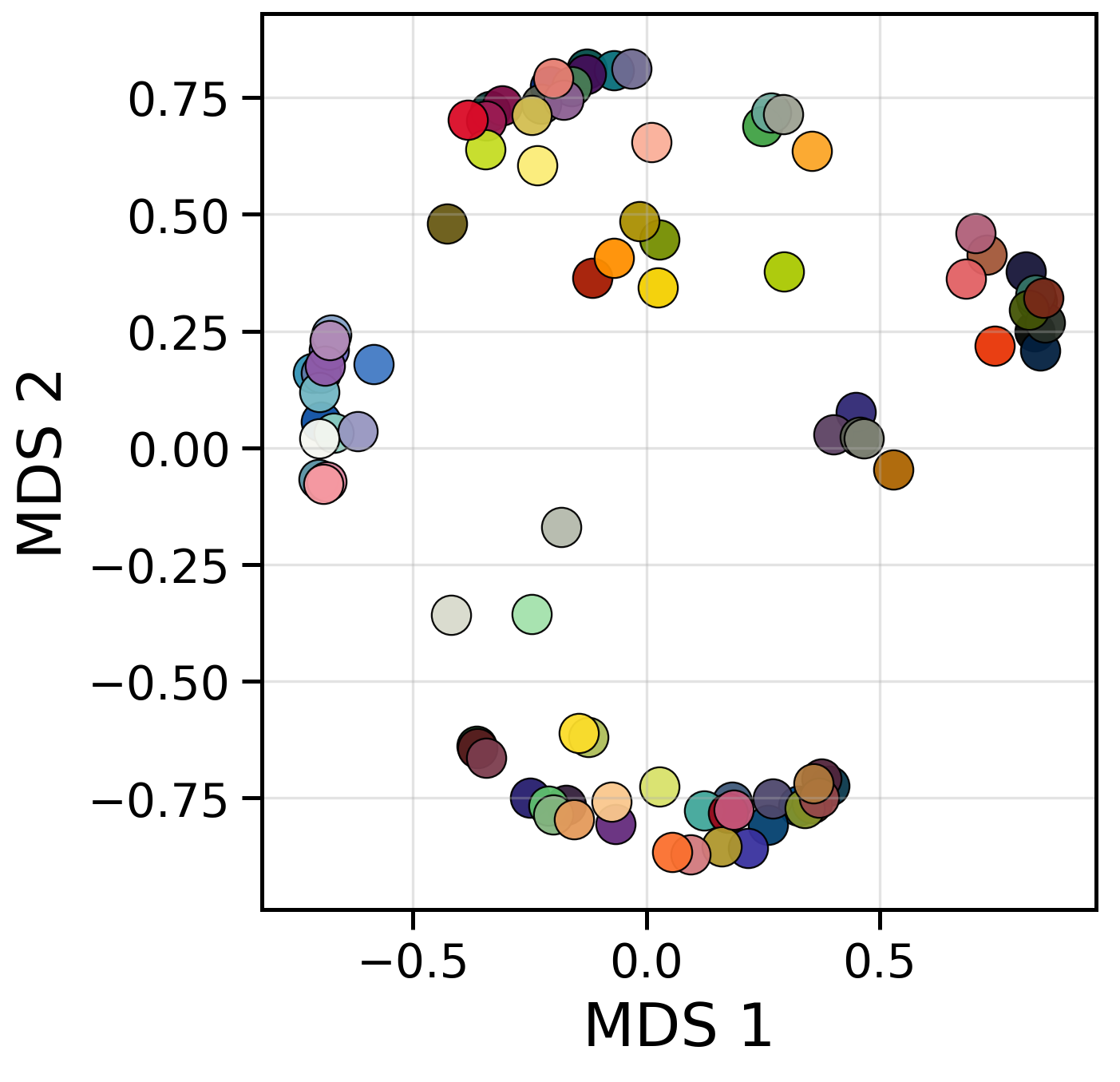} \\

        \includegraphics[width=0.28\textwidth,height=0.28\textwidth]{Figures/Color/Color-main/updated-llama8b/layer_08.png} &
        \includegraphics[width=0.28\textwidth,height=0.28\textwidth]{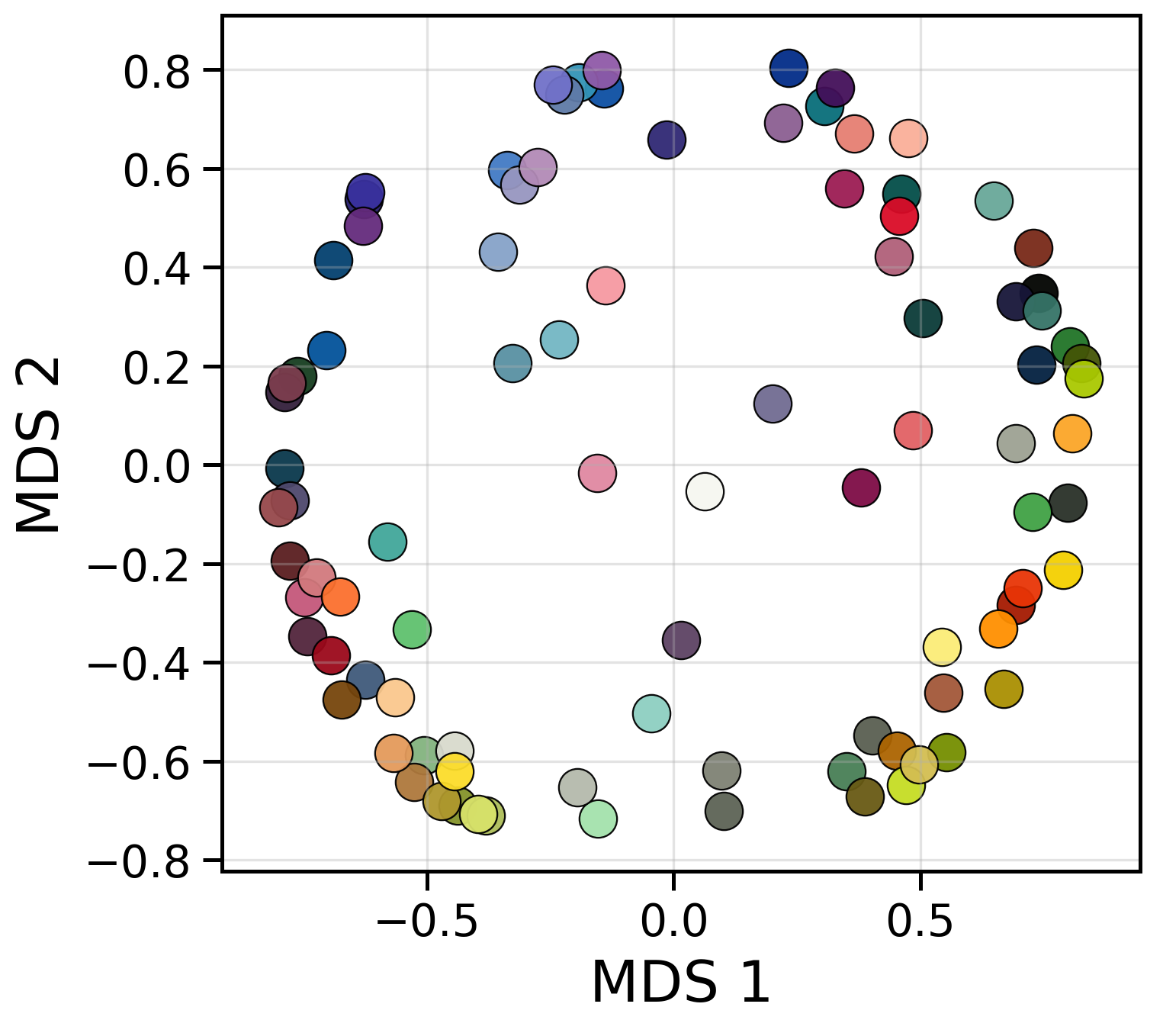}
    \end{tabular}

    \vspace{0.3em}
    \captionof{figure}{Layer-wise emergence of 2D color geometry in LLaMA-3-8B across model depth. Panels show (top-left) human perceptual geometry, (top-right) early-layer representation, (bottom-left) peak-alignment layer, and (bottom-right) final-layer MDS representation.}
    \label{fig:appendix_color_maps_2d}
\end{center}

\begin{center}
    \setlength{\tabcolsep}{2pt}
    \begin{tabular}{cc}
        \includegraphics[width=0.28\textwidth,height=0.28\textwidth]{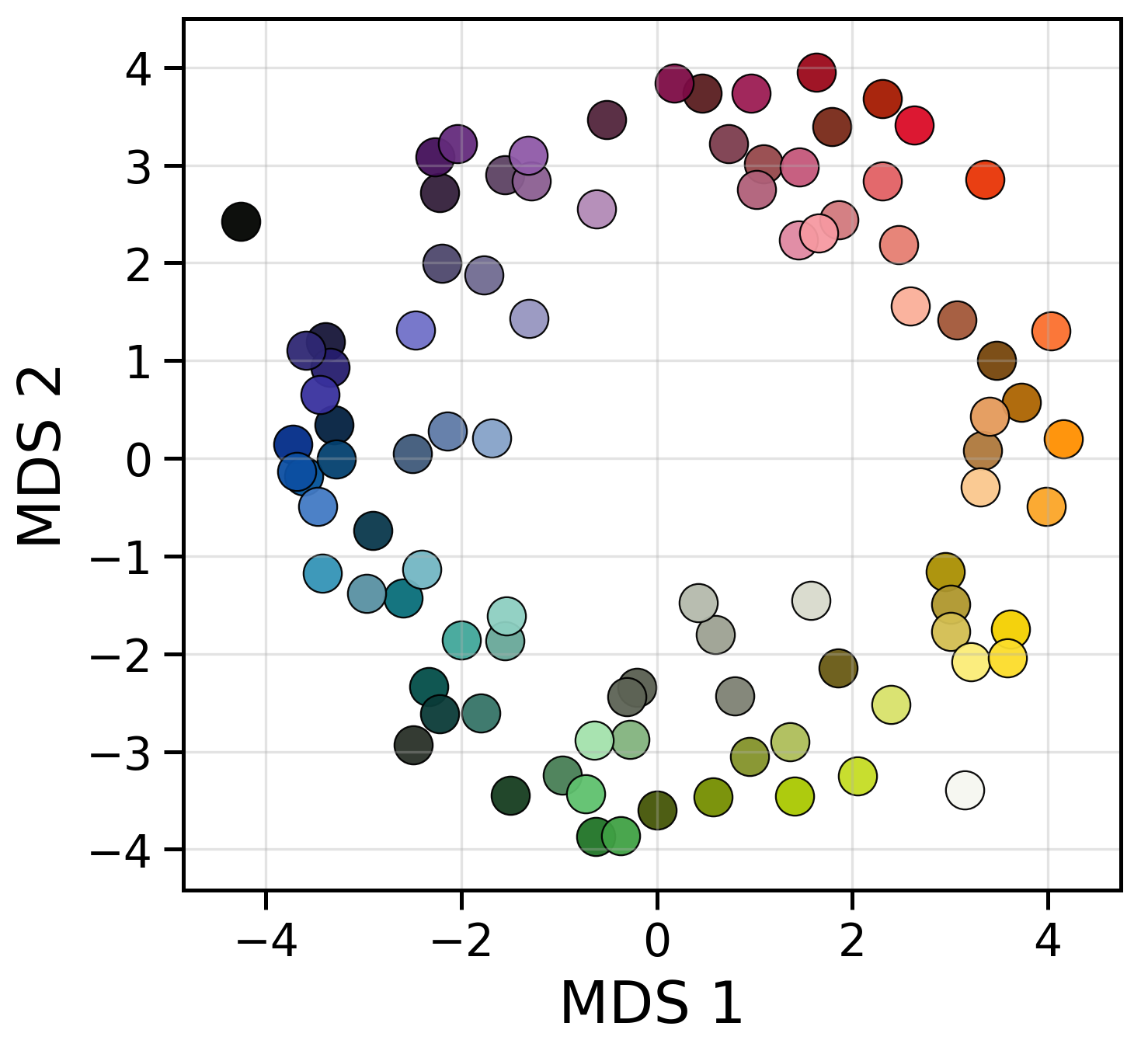} &
        \includegraphics[width=0.28\textwidth,height=0.28\textwidth]{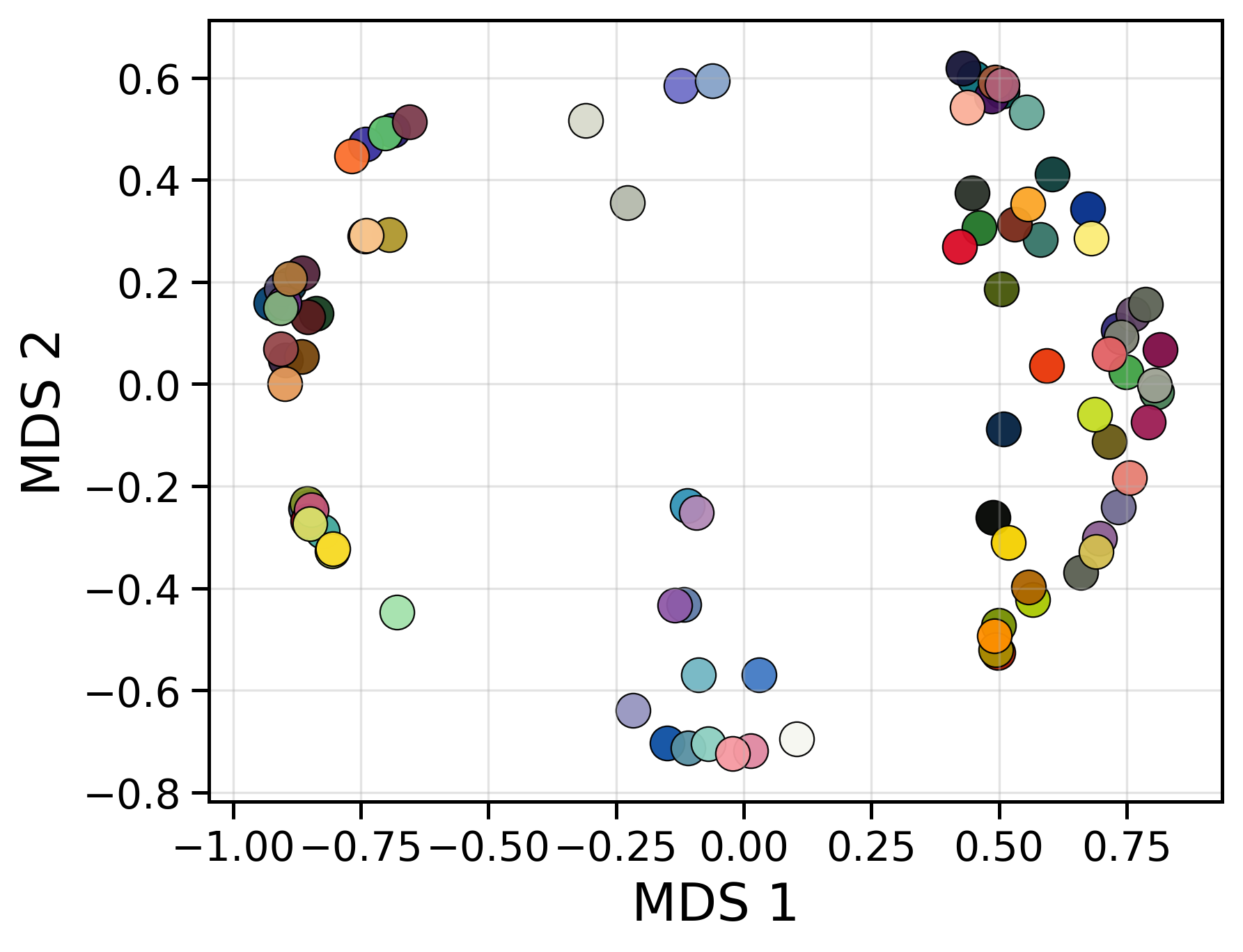} \\

        \includegraphics[width=0.28\textwidth,height=0.28\textwidth]{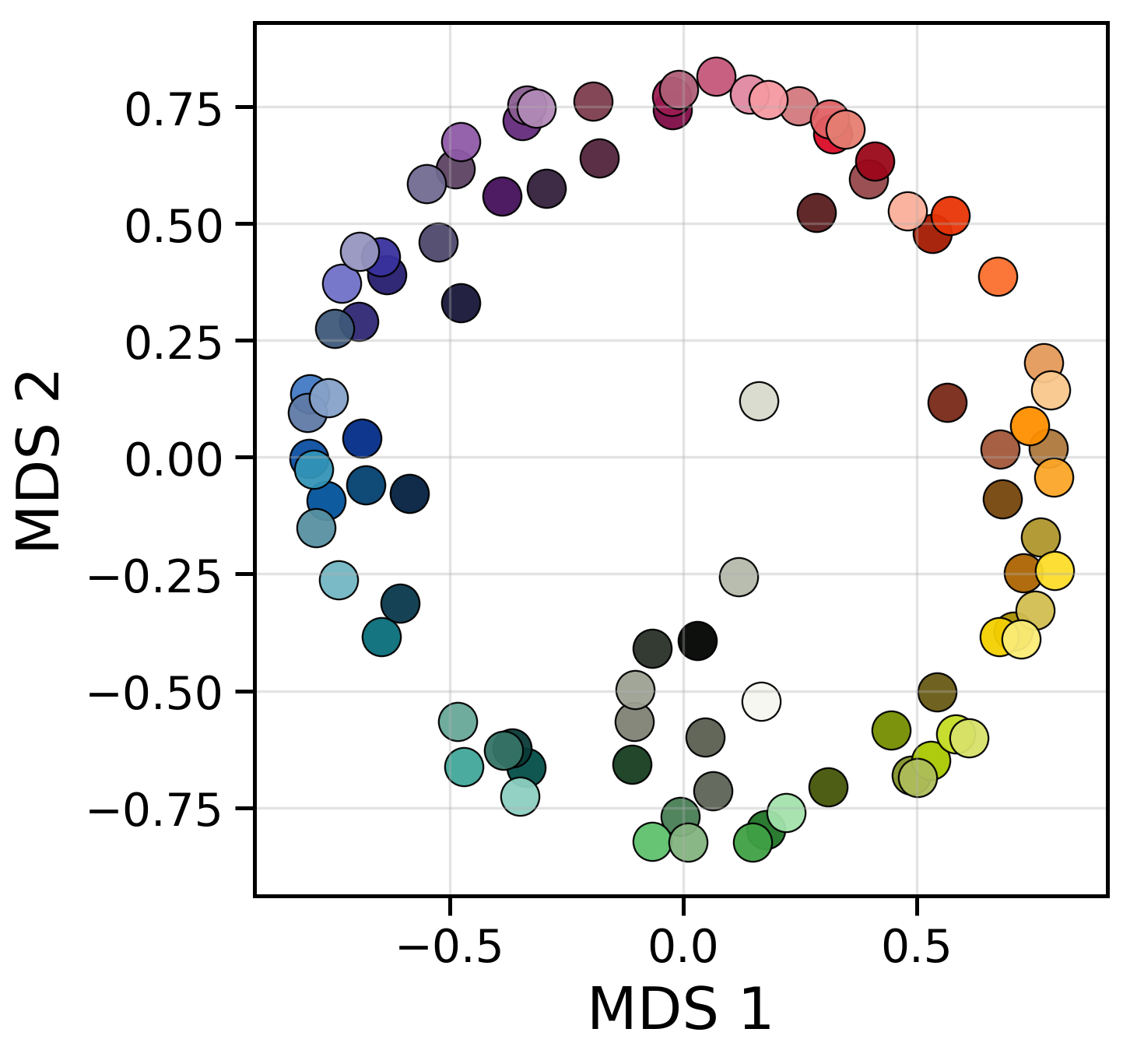} &
        \includegraphics[width=0.28\textwidth,height=0.28\textwidth]{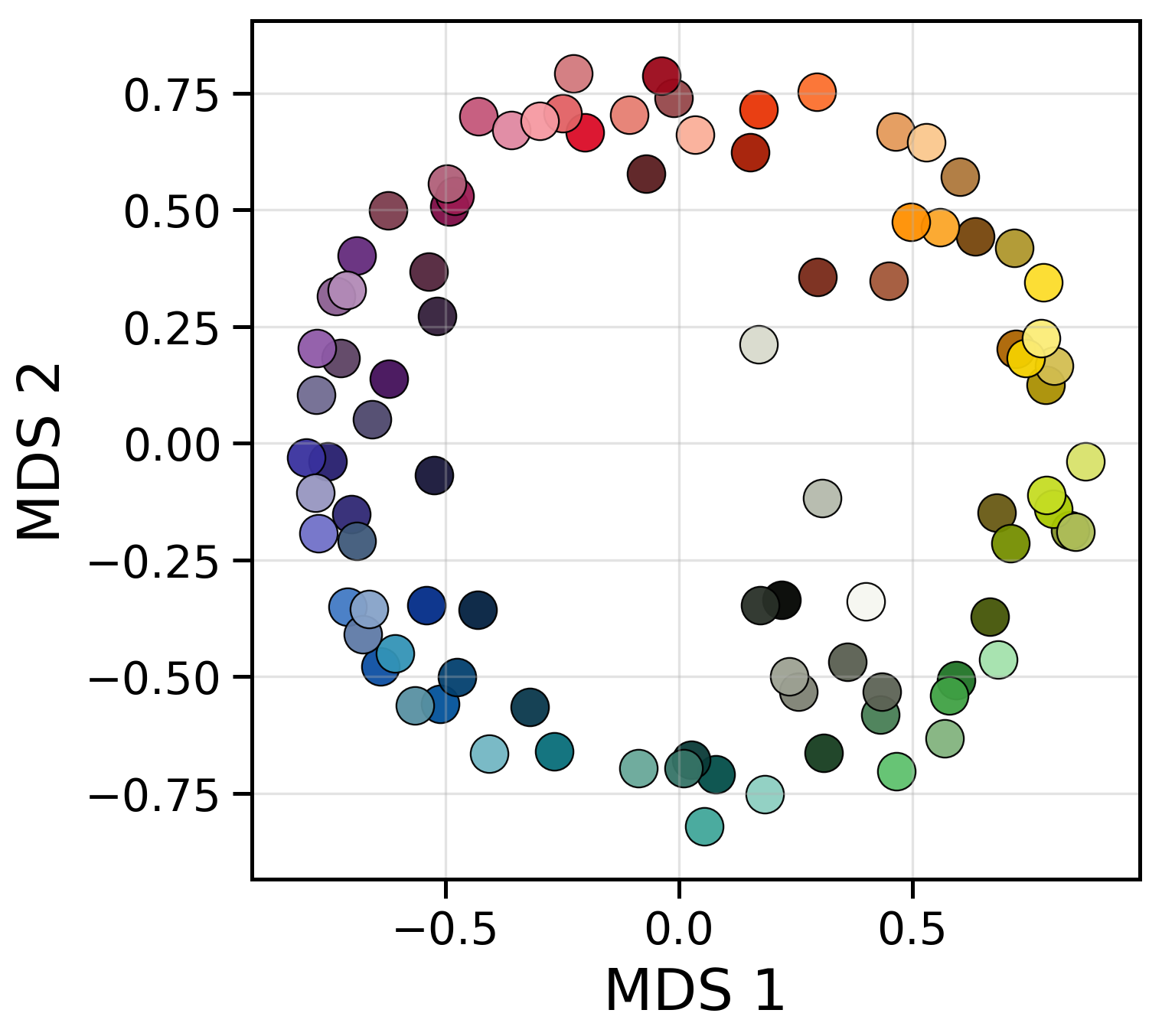}
    \end{tabular}

    \vspace{0.3em}
    \captionof{figure}{Layer-wise emergence of 2D color geometry in Qwen-3-4B across model depth. Panels show (top-left) human perceptual geometry, (top-right) early-layer representation, (bottom-left) peak-alignment layer, and (bottom-right) final-layer MDS representation.}
    \label{fig:appendix_color_maps_2d_qwen}
\end{center}

\begin{center}
    \setlength{\tabcolsep}{2pt}
    \begin{tabular}{cc}
        \includegraphics[width=0.35\textwidth,height=0.28\textwidth]{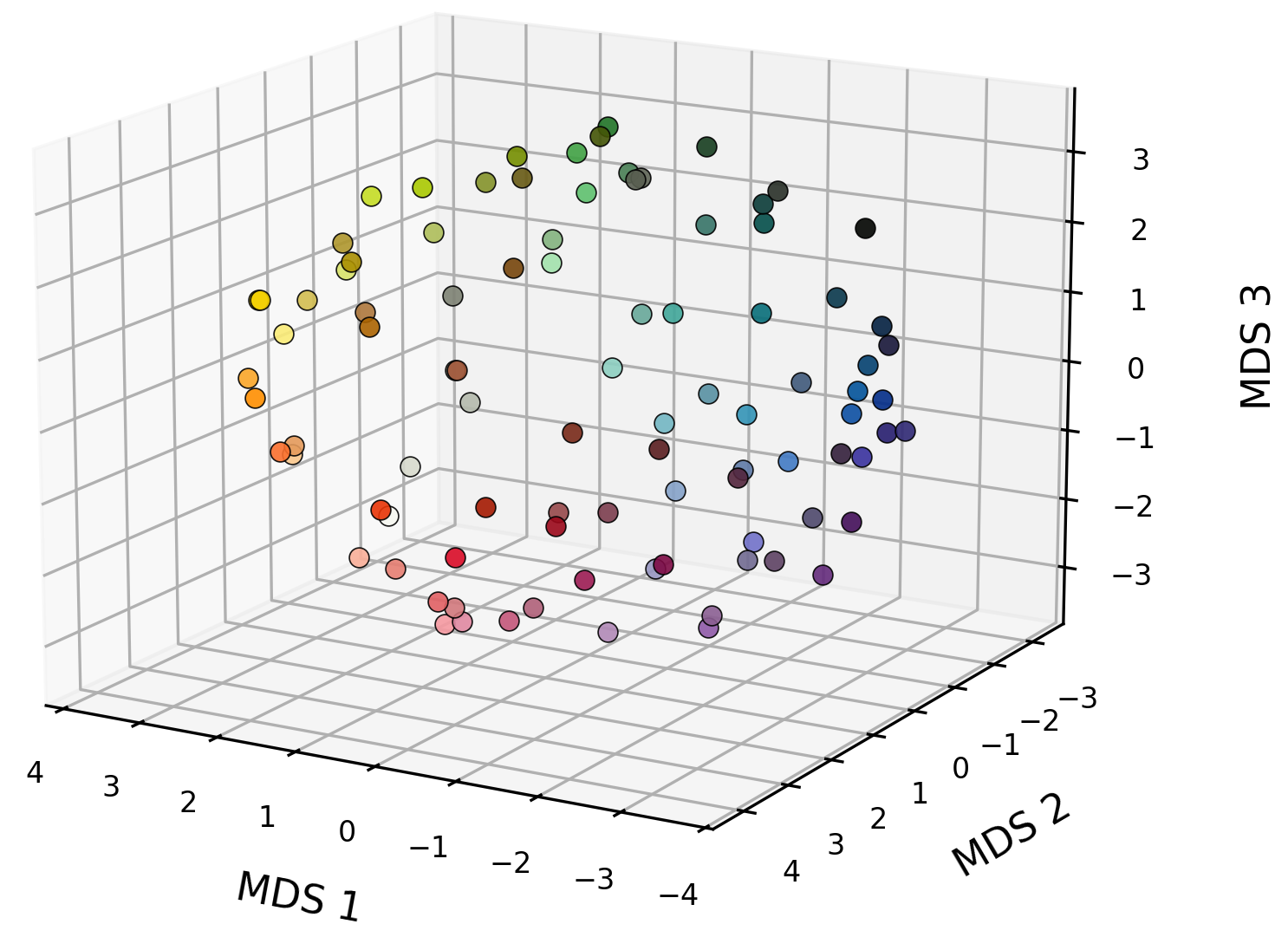} &
        \includegraphics[width=0.35\textwidth,height=0.28\textwidth]{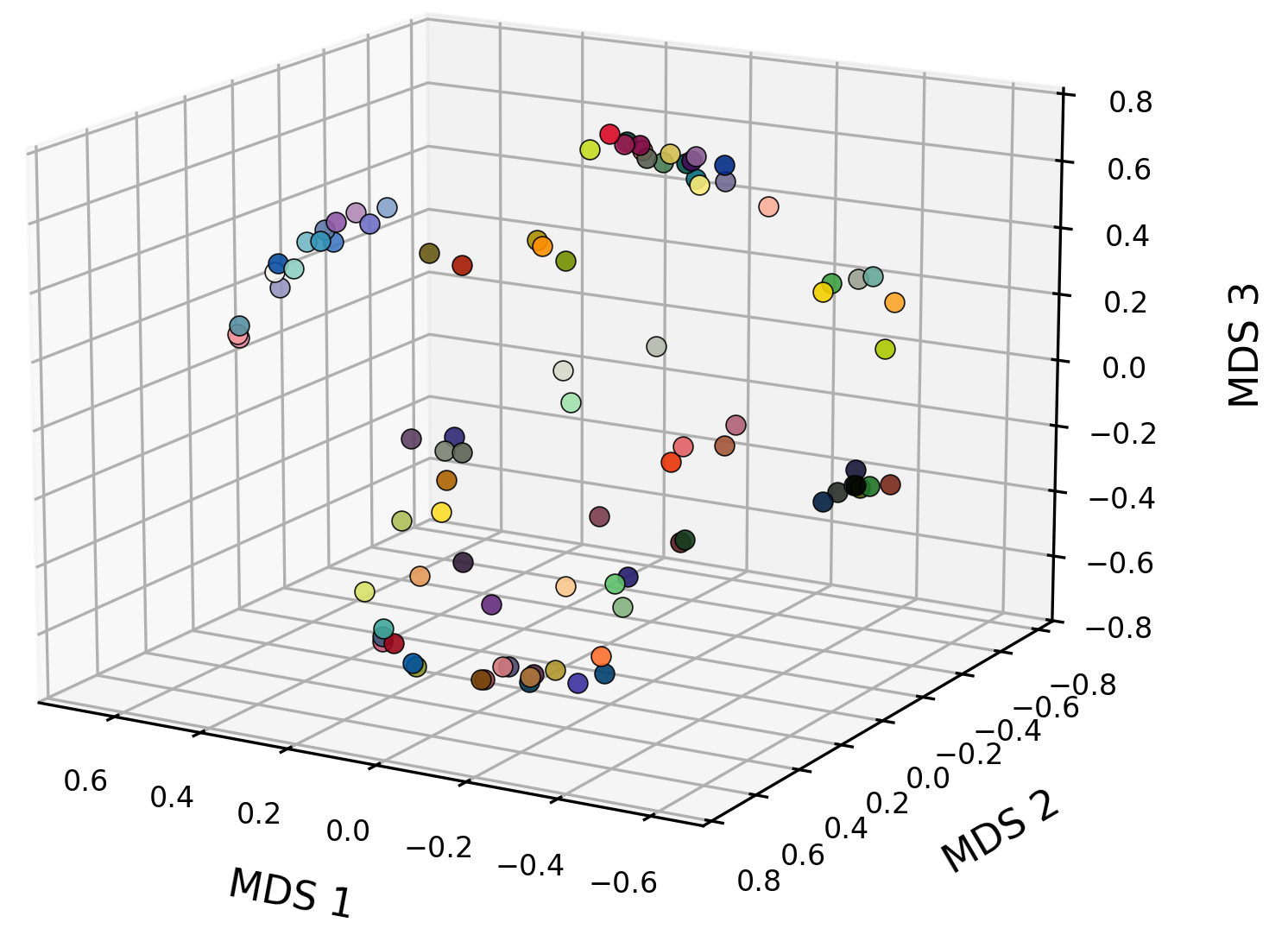} \\

        \includegraphics[width=0.35\textwidth,height=0.28\textwidth]{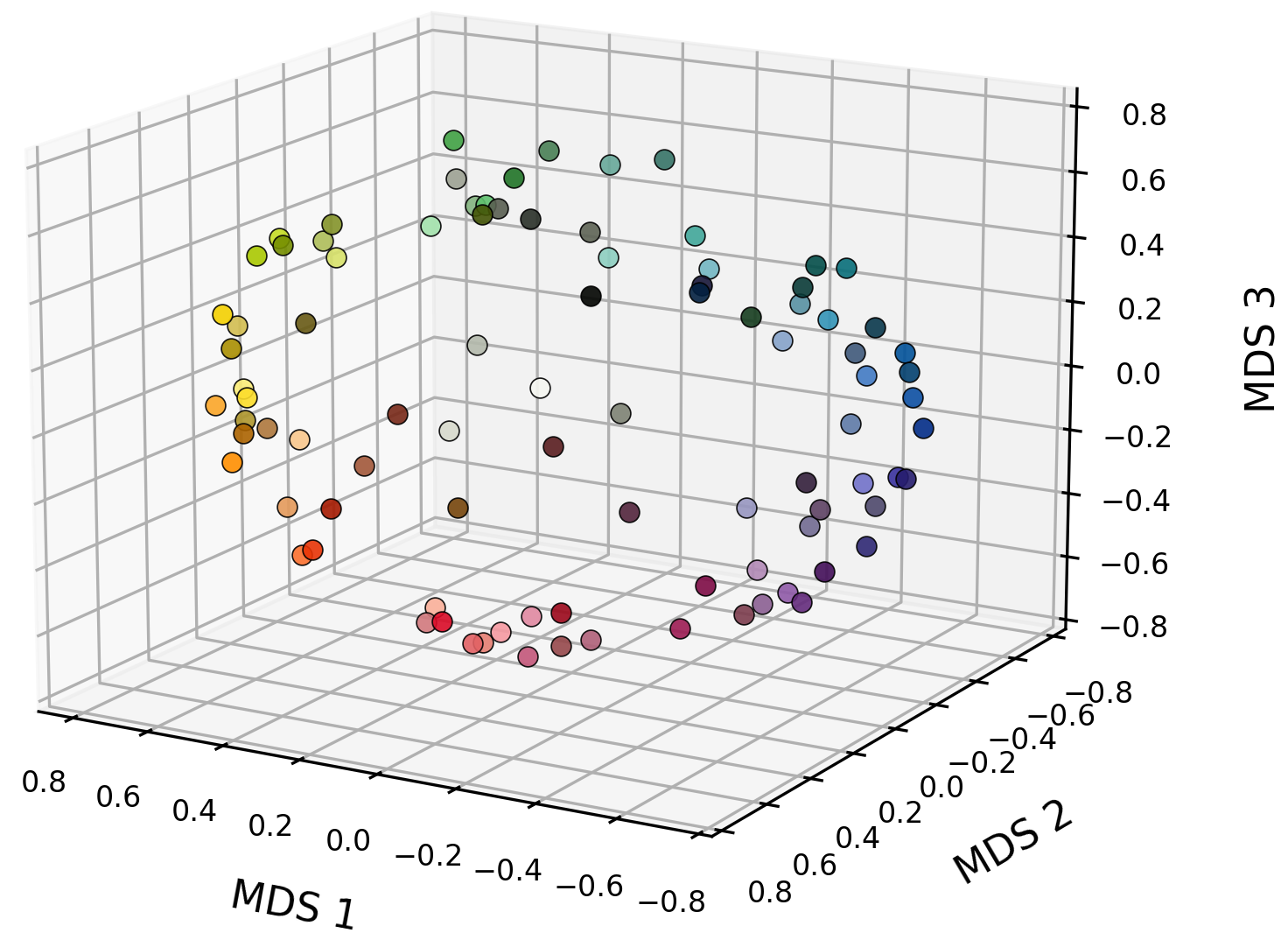} &
        \includegraphics[width=0.35\textwidth,height=0.28\textwidth]{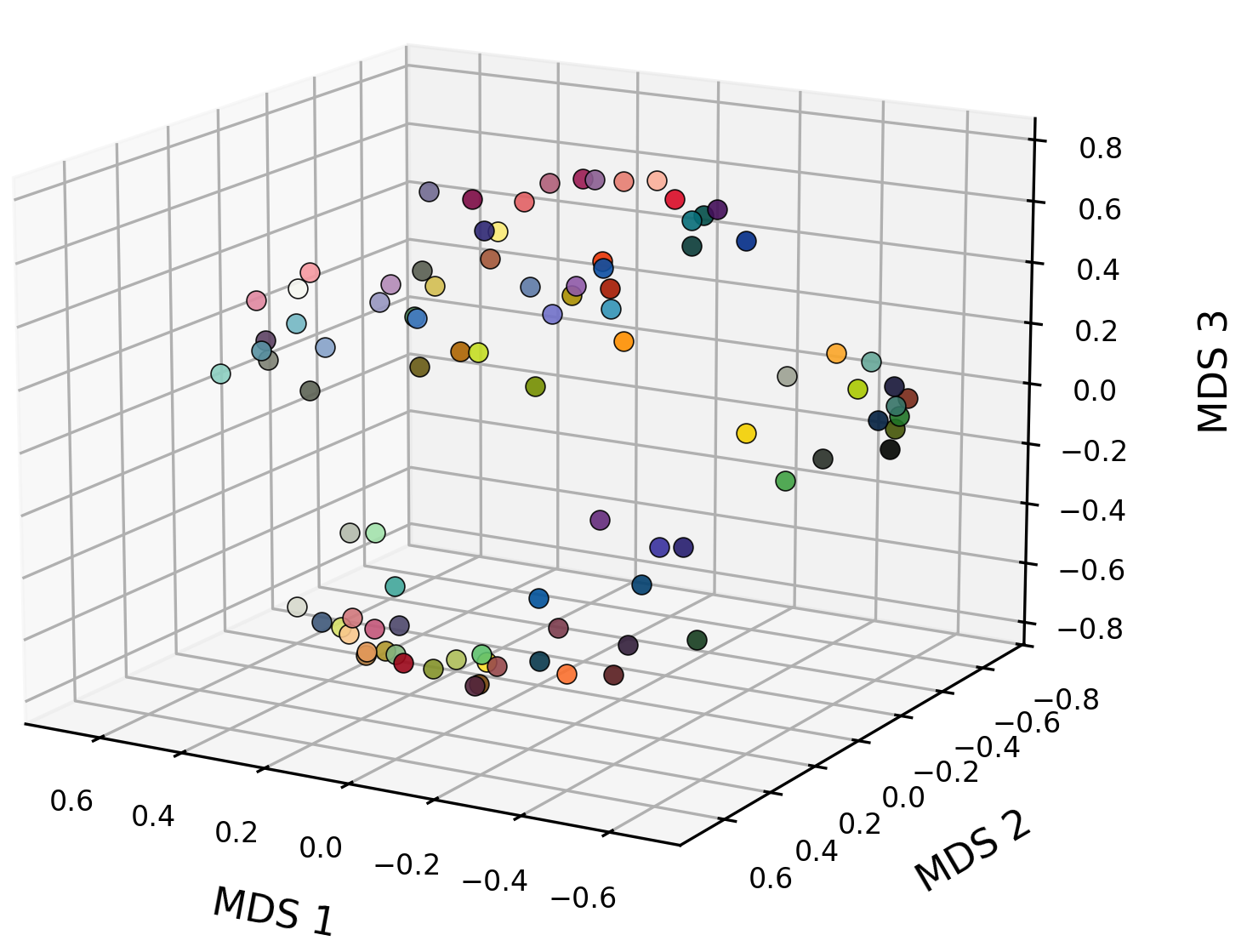}
    \end{tabular}

    \vspace{0.3em}
    \captionof{figure}{Layer-wise emergence of 3D color geometry in LLaMA-3-8B across model depth. Panels show (top-left) human perceptual geometry, (top-right) early-layer representation, (bottom-left) peak-alignment layer, and (bottom-right) final-layer MDS representation.}
    \label{fig:appendix_color_maps_3d}
\end{center}

\begin{center}
    \setlength{\tabcolsep}{2pt}
    \begin{tabular}{cc}
        \includegraphics[width=0.28\textwidth,height=0.28\textwidth]{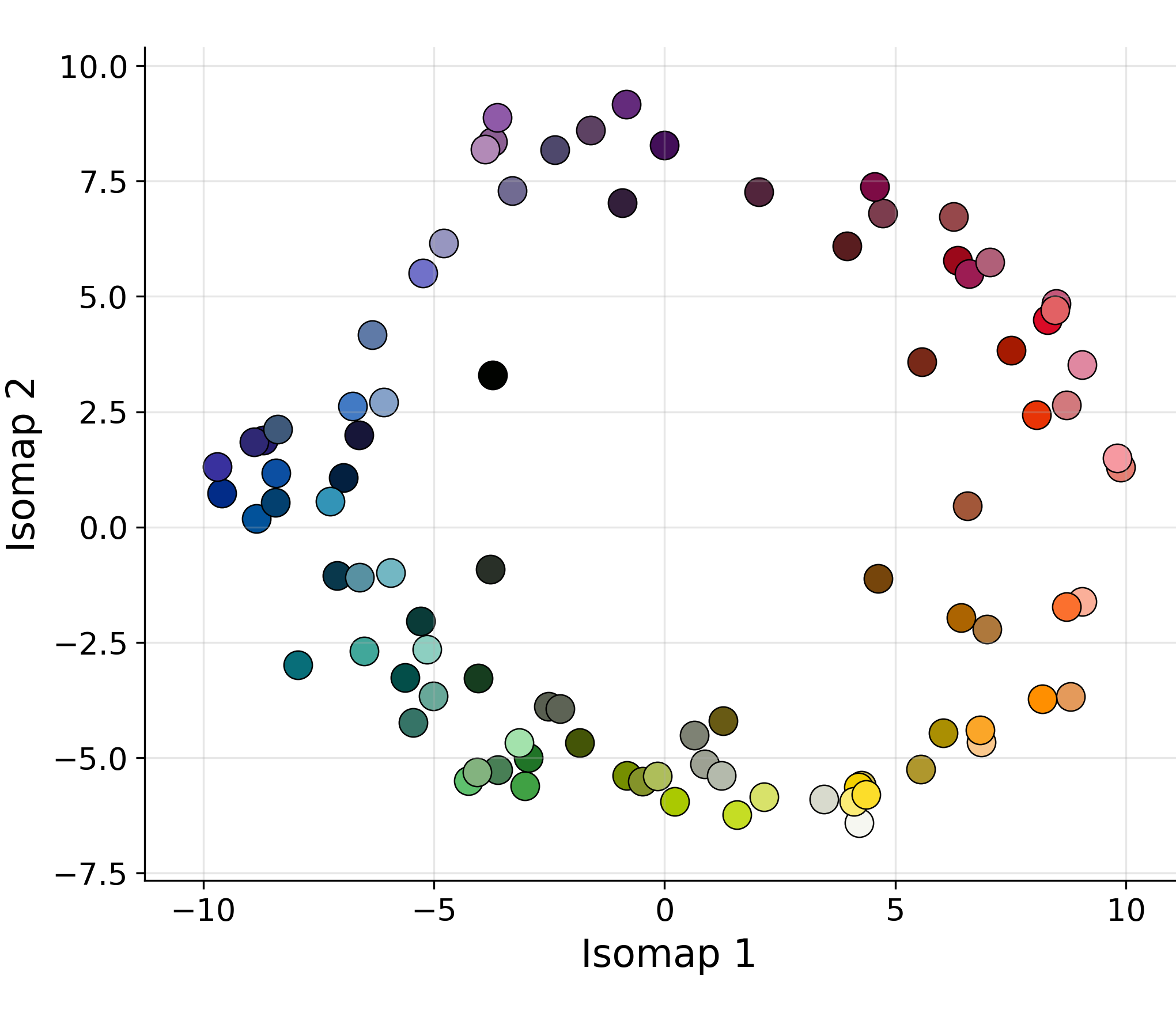} &
        \includegraphics[width=0.28\textwidth,height=0.28\textwidth]{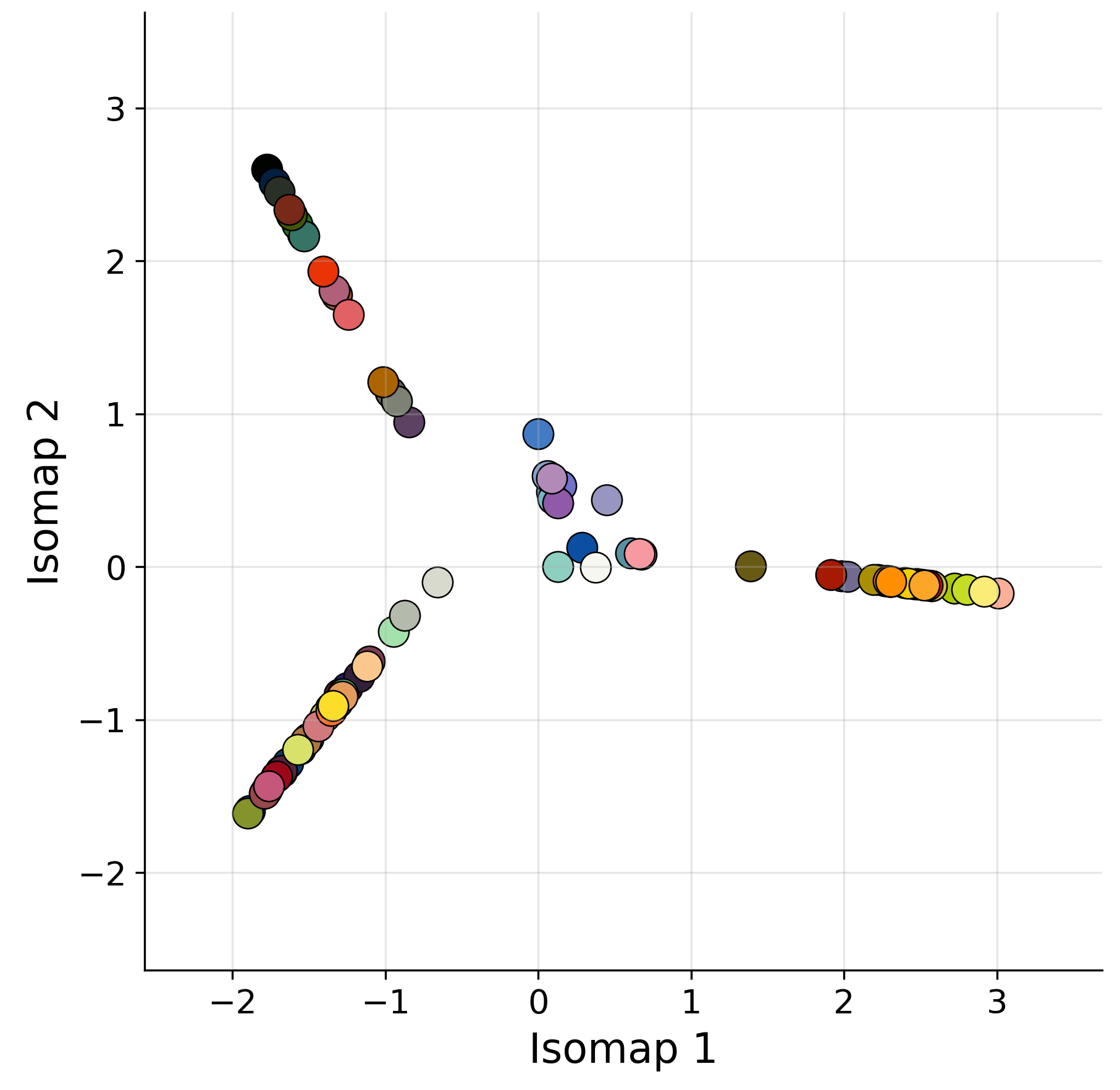} \\

        \includegraphics[width=0.28\textwidth,height=0.28\textwidth]{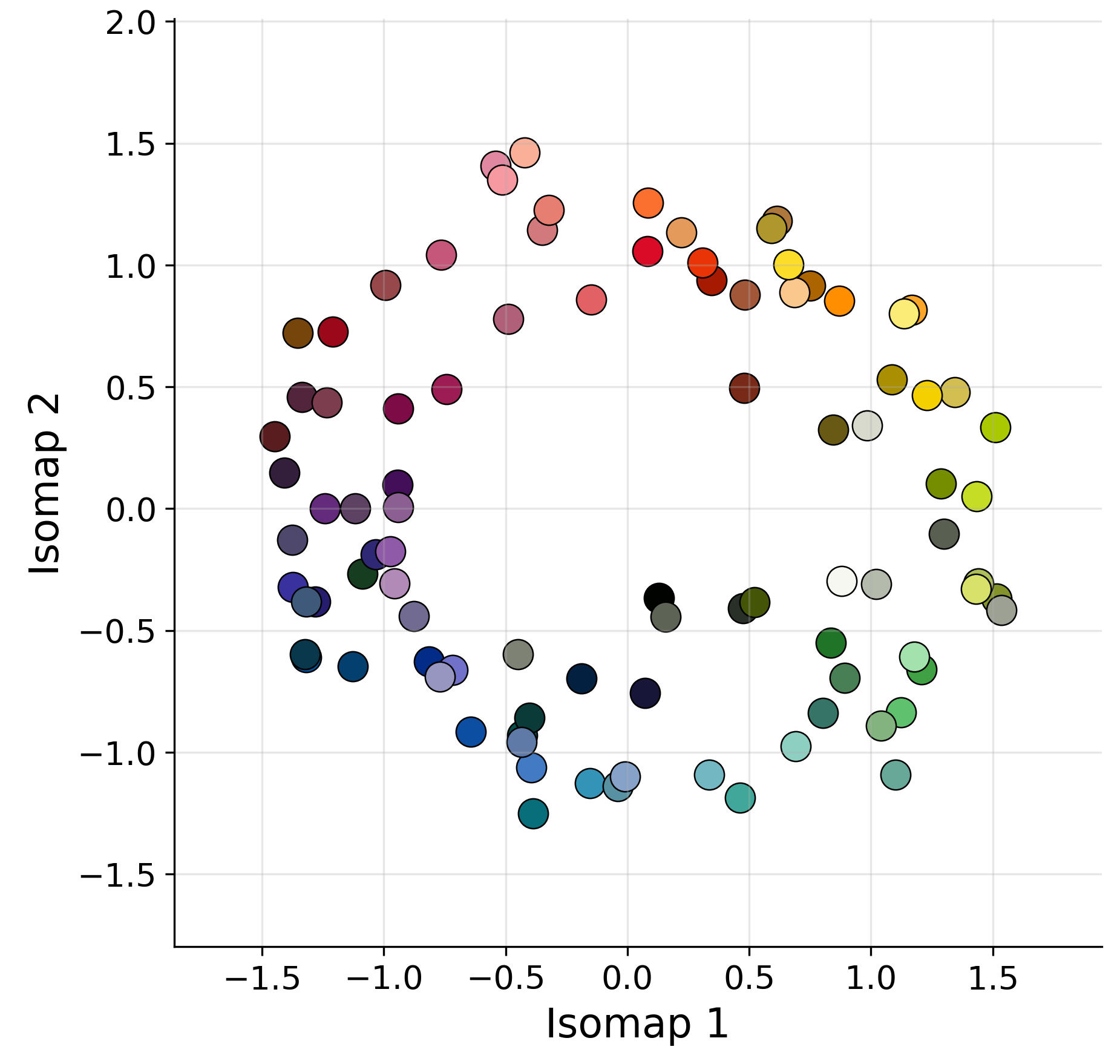} &
        \includegraphics[width=0.28\textwidth,height=0.28\textwidth]{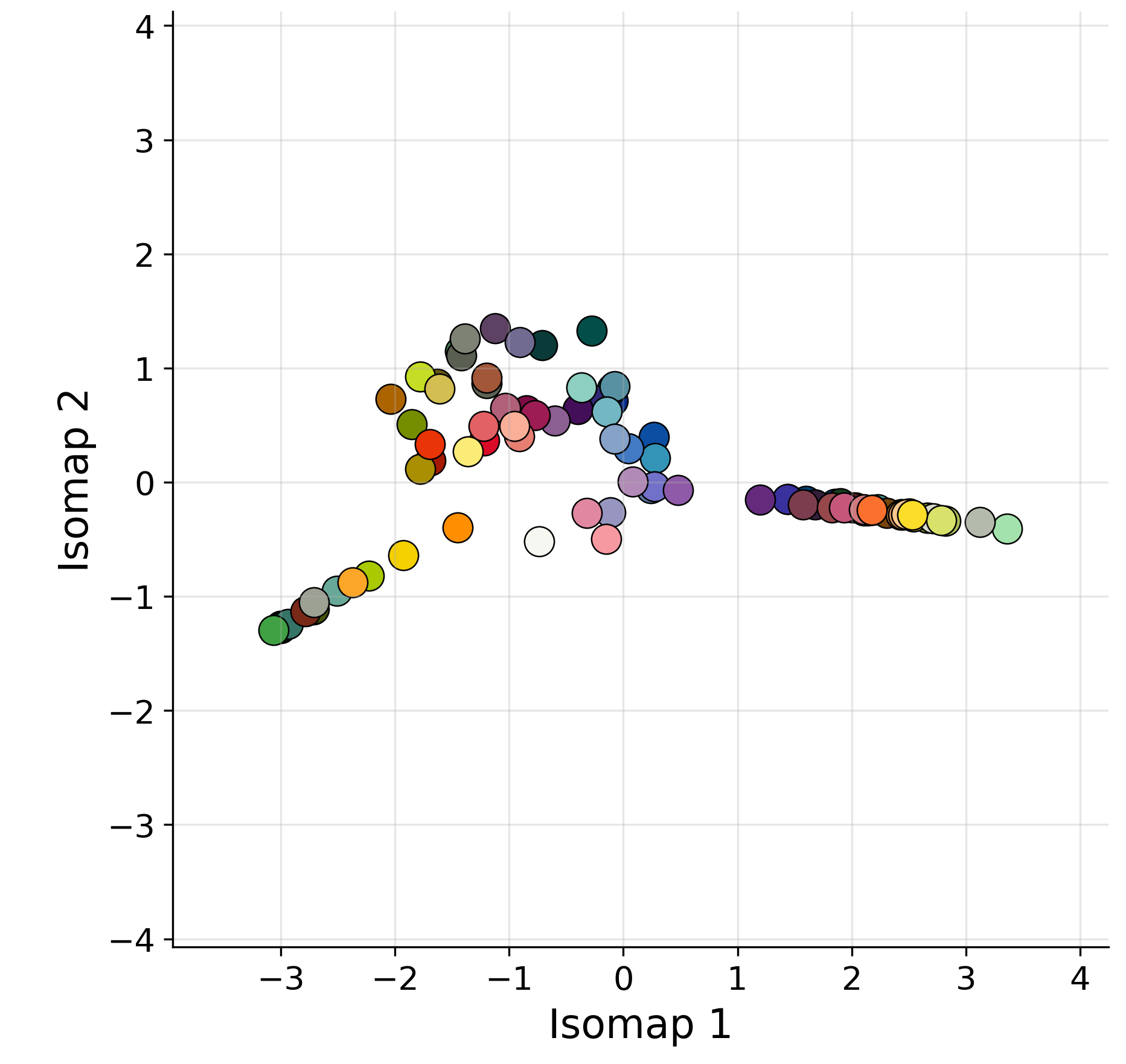}
    \end{tabular}

    \vspace{0.3em}
    \captionof{figure}{Layer-wise emergence of 2D color geometry in LLaMA-3-8B across model depth. Panels show (top-left) human perceptual geometry, (top-right) early-layer representation, (bottom-left) peak-alignment layer, and (bottom-right) final-layer isomap representation.}
    \label{fig:appendix_color_maps_2d_iso}
\end{center}

\begin{center}
    \setlength{\tabcolsep}{2pt}
    \begin{tabular}{cc}
        \includegraphics[width=0.35\textwidth,height=0.28\textwidth]{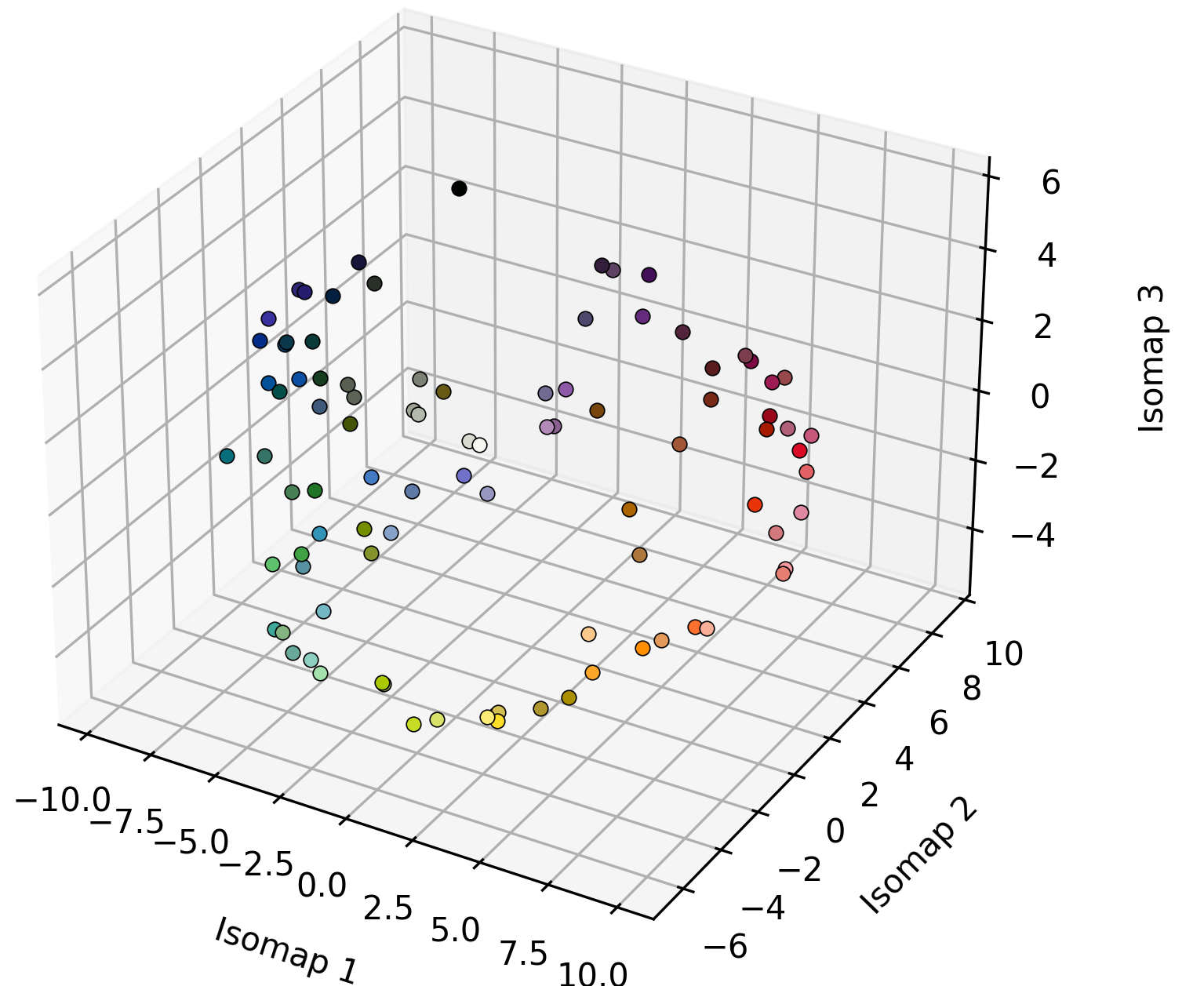} &
        \includegraphics[width=0.35\textwidth,height=0.28\textwidth]{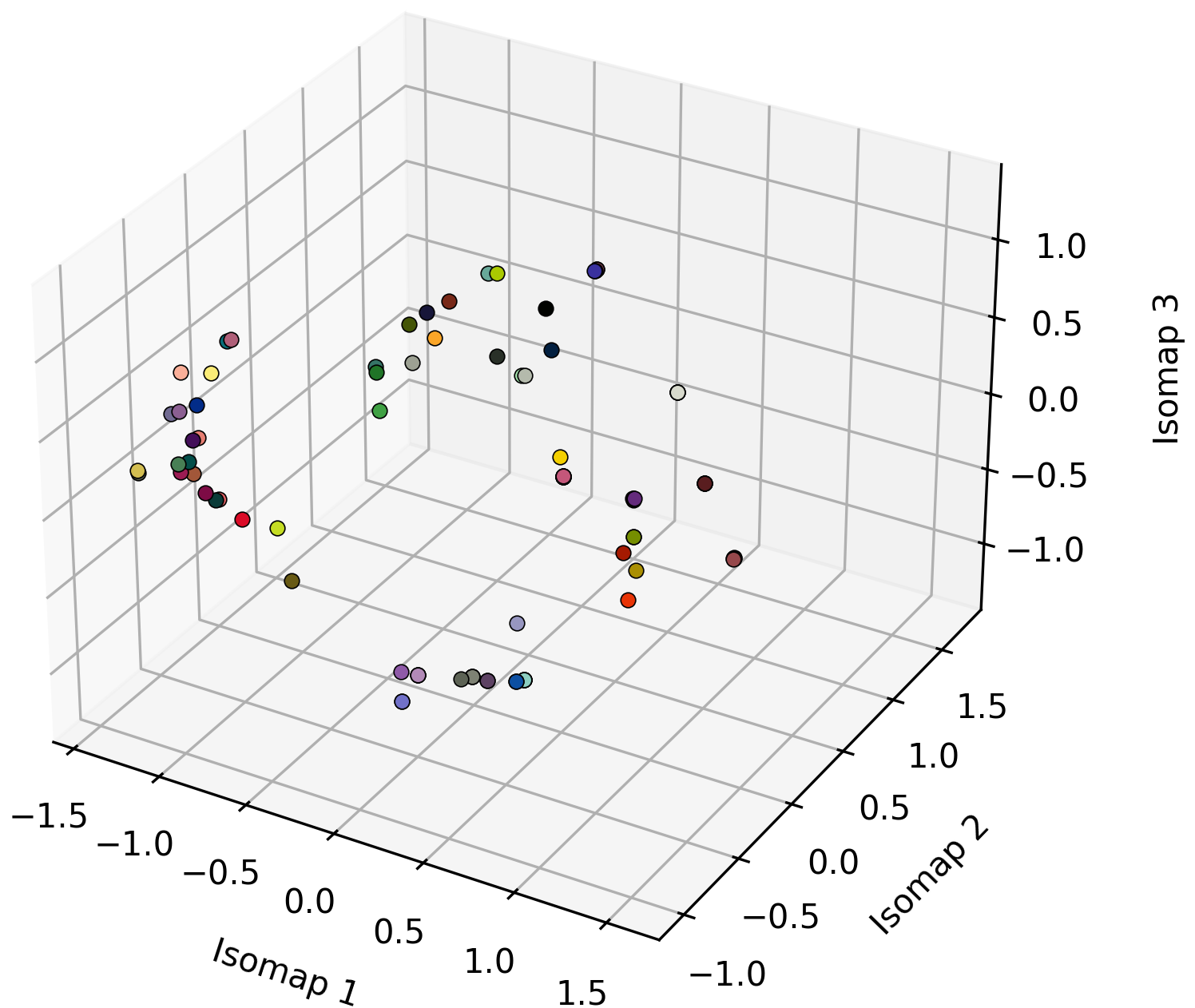} \\

        \includegraphics[width=0.35\textwidth,height=0.28\textwidth]{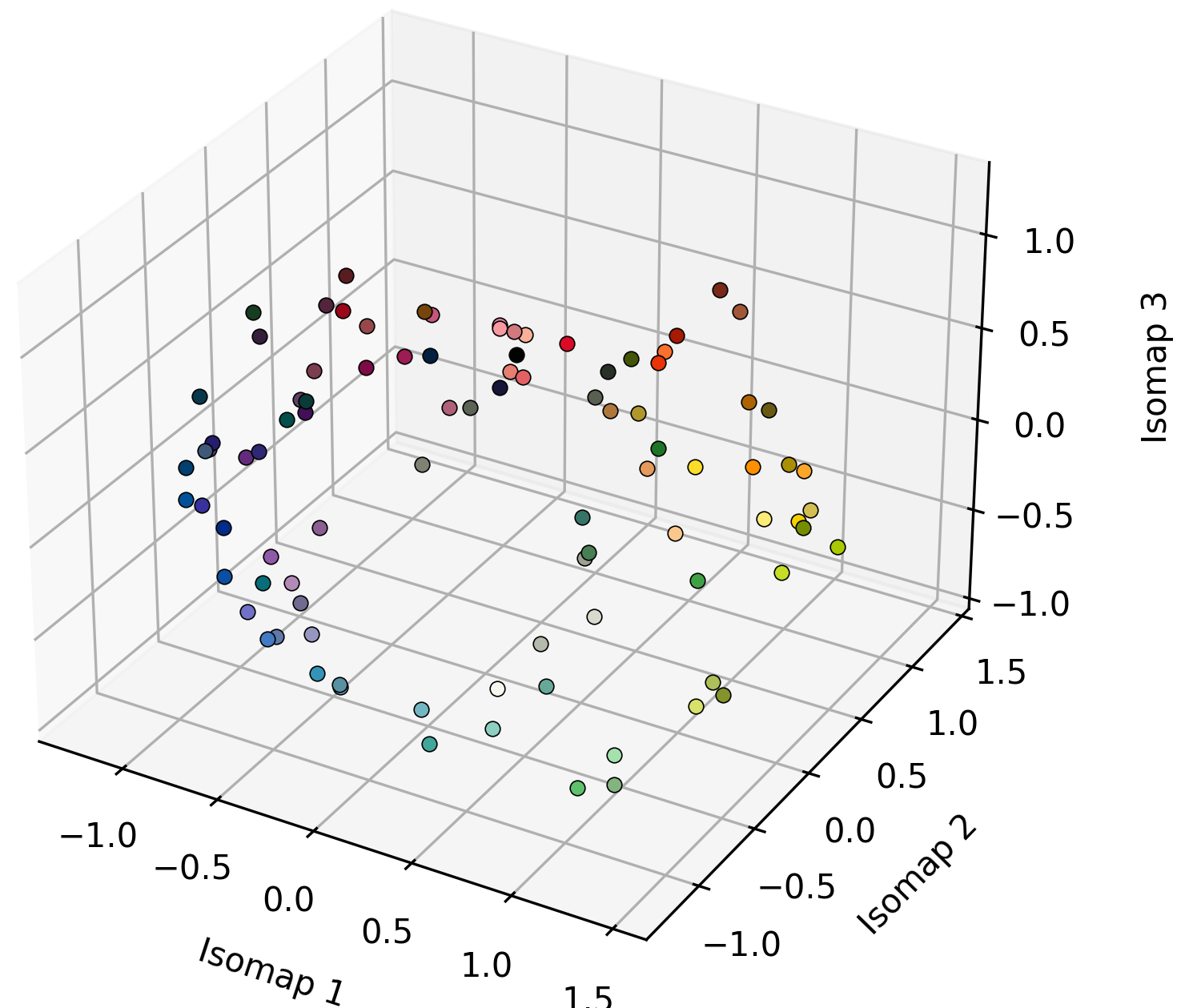} &
        \includegraphics[width=0.35\textwidth,height=0.28\textwidth]{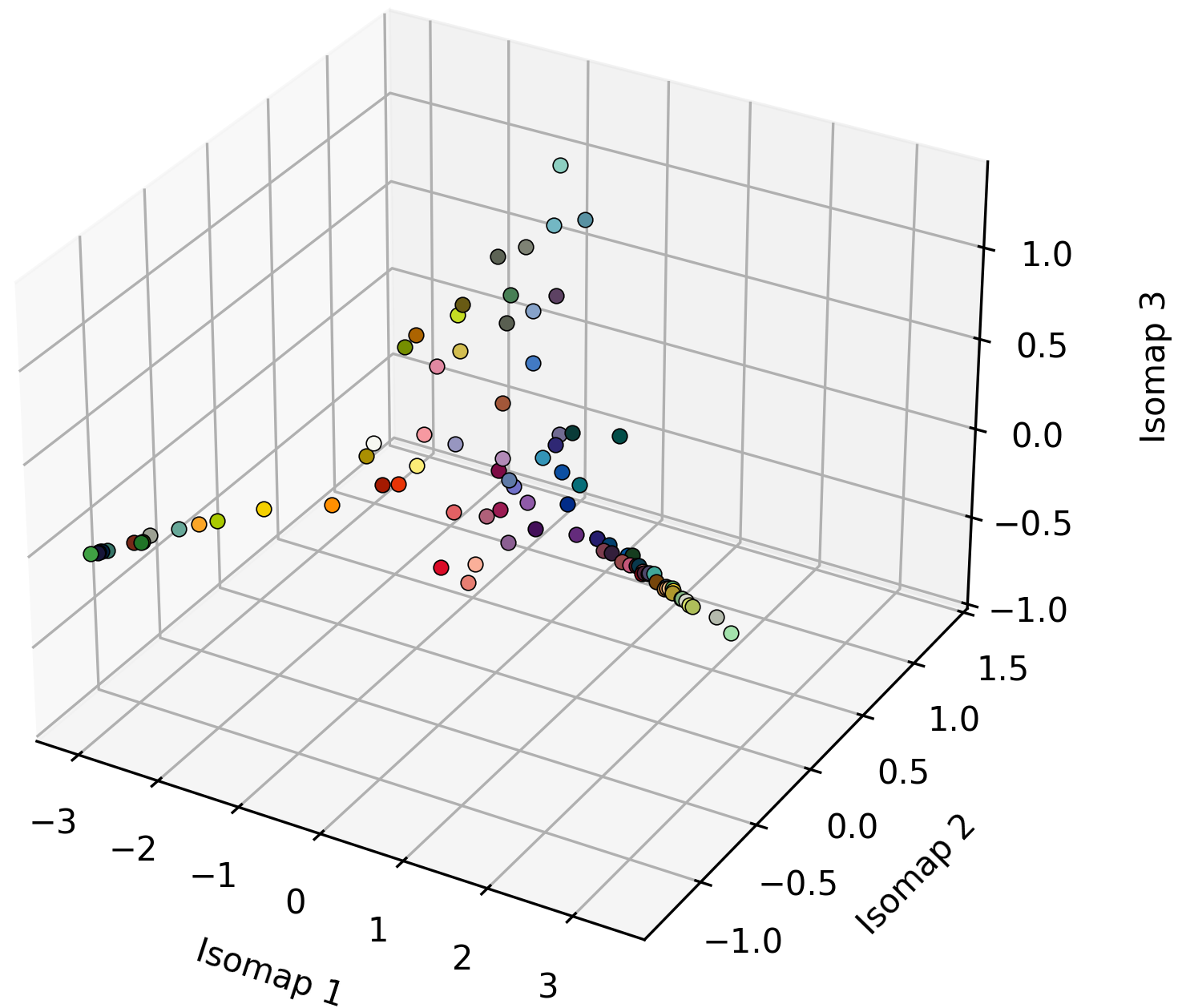}
    \end{tabular}

    \vspace{0.3em}
    \captionof{figure}{Layer-wise emergence of 3D color geometry in LLaMA-3-8B across model depth. Panels show (top-left) human perceptual geometry, (top-right) early-layer representation, (bottom-left) peak-alignment layer, and (bottom-right) final-layer isomap representation.}
    \label{fig:appendix_color_maps_3d_iso}
\end{center}

\begin{center}
    \setlength{\tabcolsep}{2pt}
    \begin{tabular}{cc}
        \includegraphics[width=0.28\textwidth,height=0.28\textwidth]{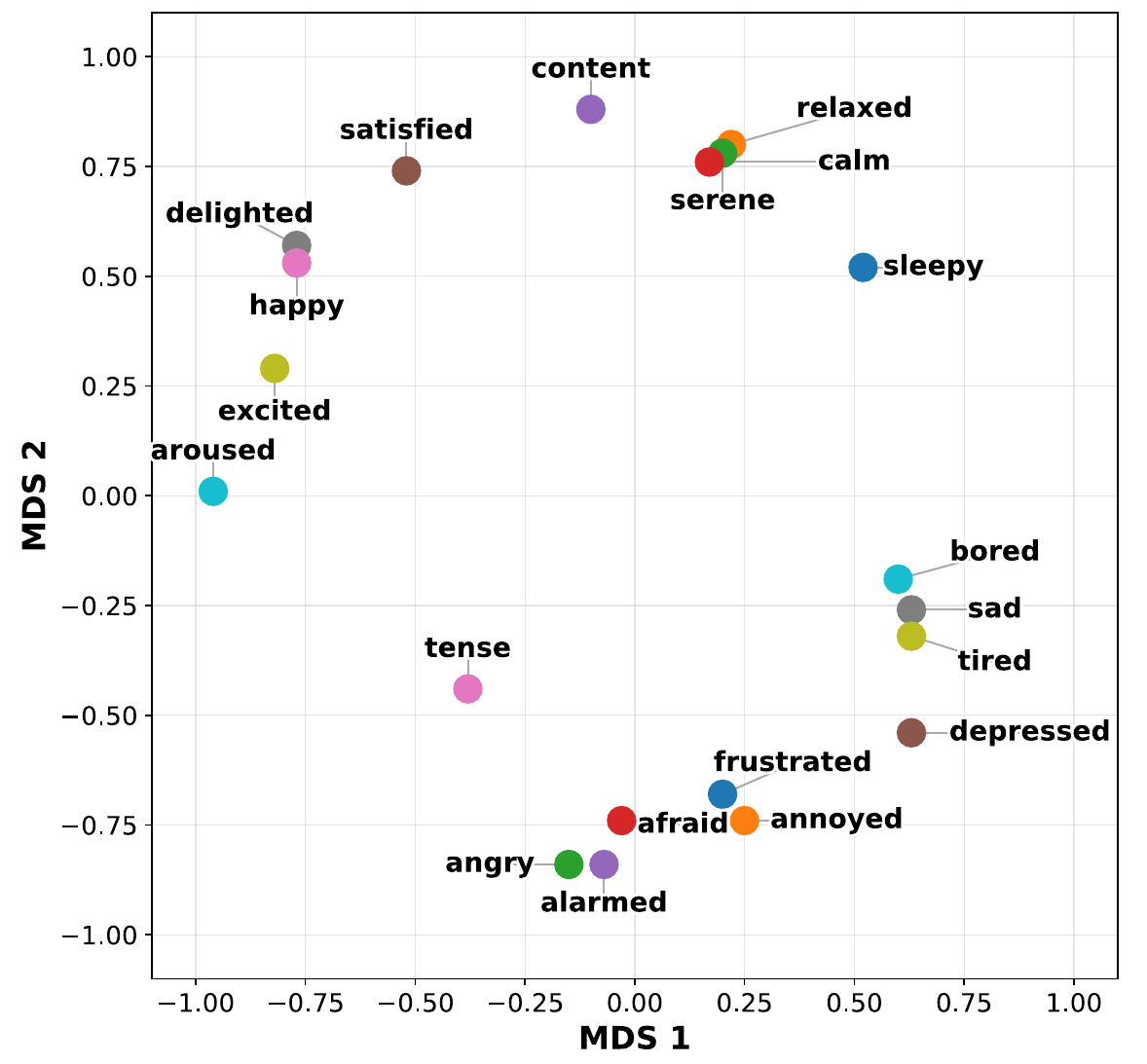} &
        \includegraphics[width=0.28\textwidth,height=0.28\textwidth]{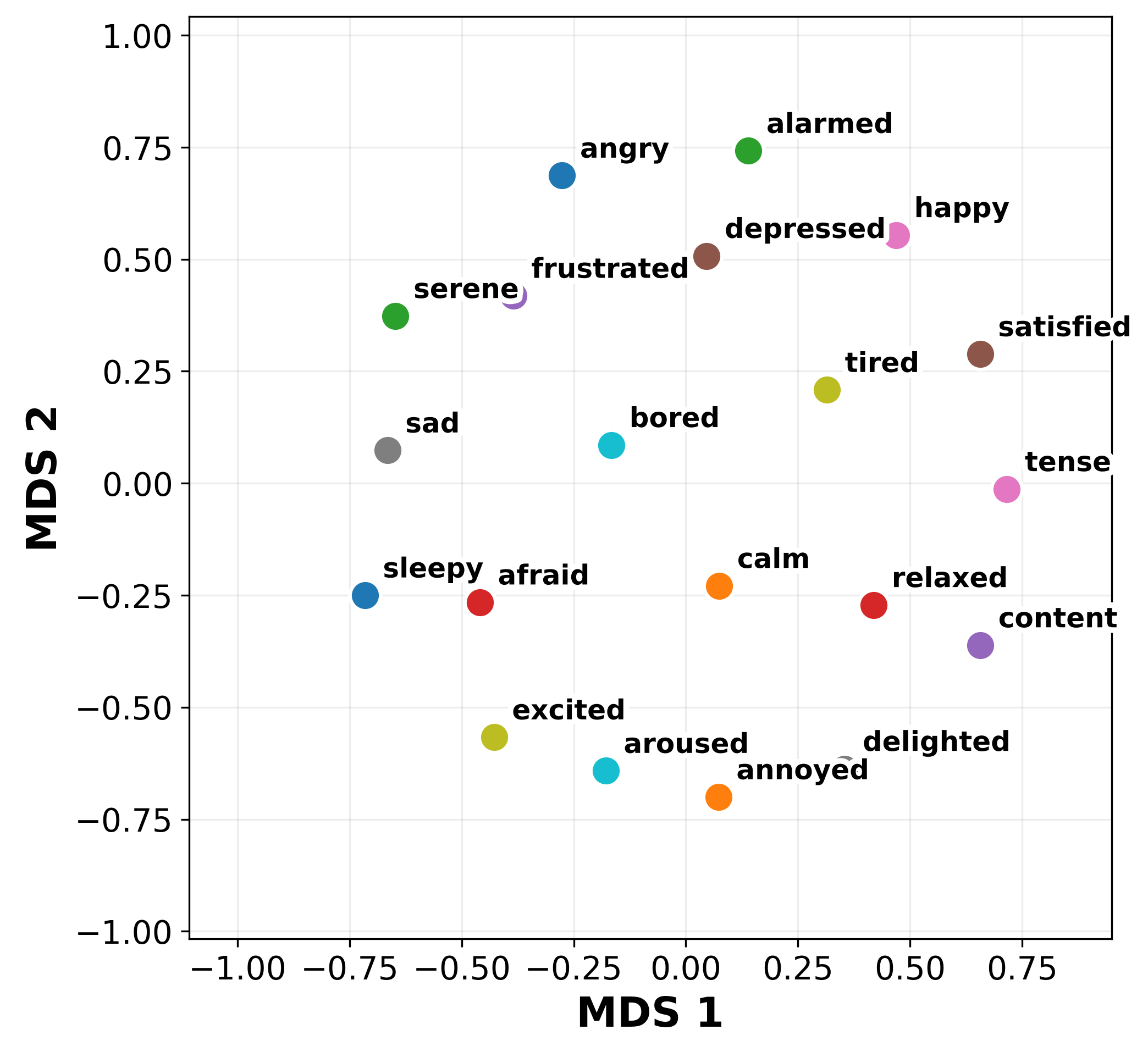} \\

        \includegraphics[width=0.28\textwidth,height=0.28\textwidth]{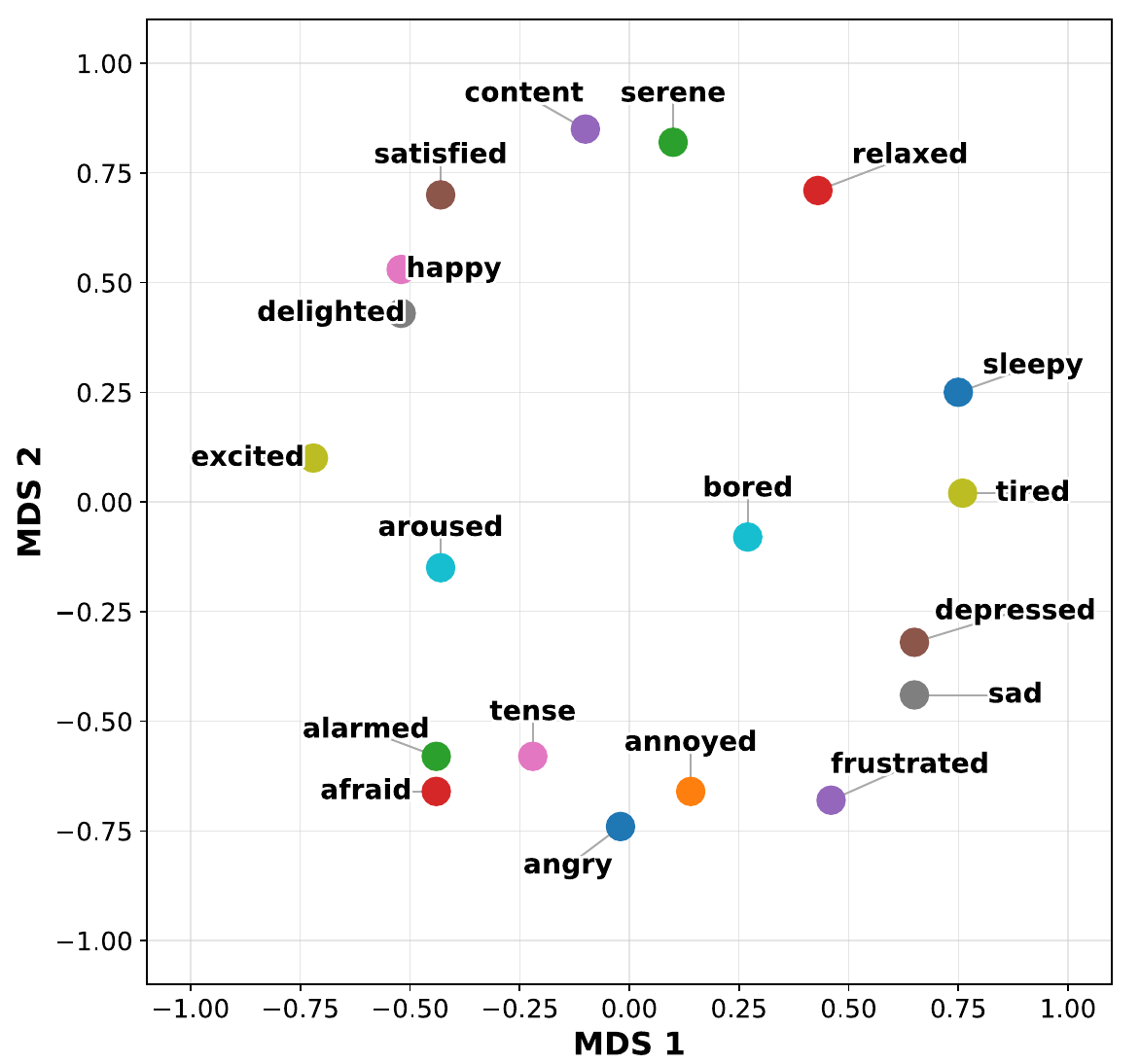} &
        \includegraphics[width=0.28\textwidth,height=0.28\textwidth]{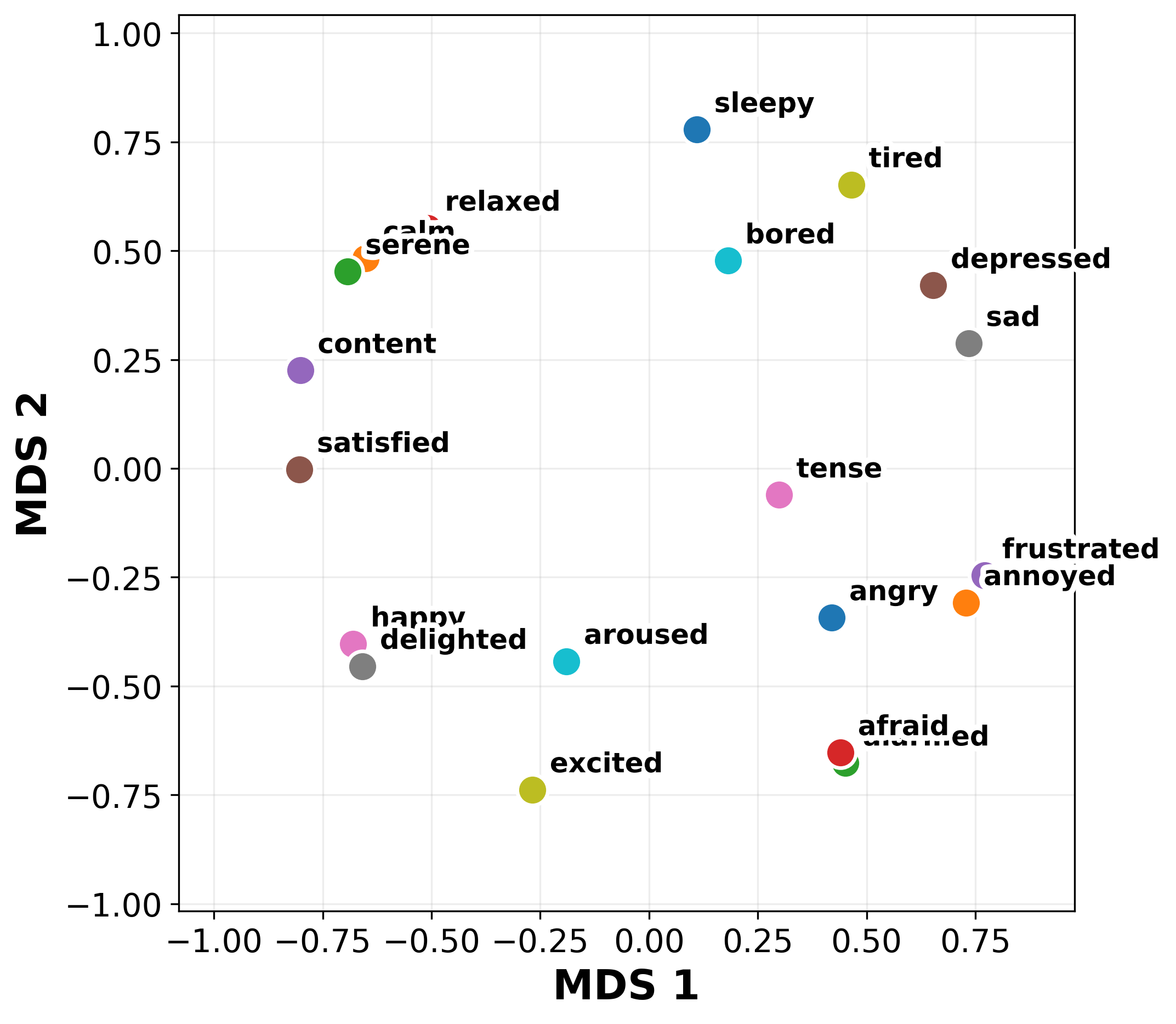}
    \end{tabular}

    \vspace{0.3em}
    \captionof{figure}{Layer-wise emergence of emotion geometry in Gemma-7B across model depth. Panels show (top-left) human perceptual geometry, (top-right) early-layer representation, (bottom-left) peak-alignment layer, and (bottom-right) final-layer MDS representation.}
    \label{fig:appendix_emo_maps_2d_mds}
\end{center}

\begin{center}
    \setlength{\tabcolsep}{2pt}
    \begin{tabular}{cc}
        \includegraphics[width=0.28\textwidth,height=0.28\textwidth]{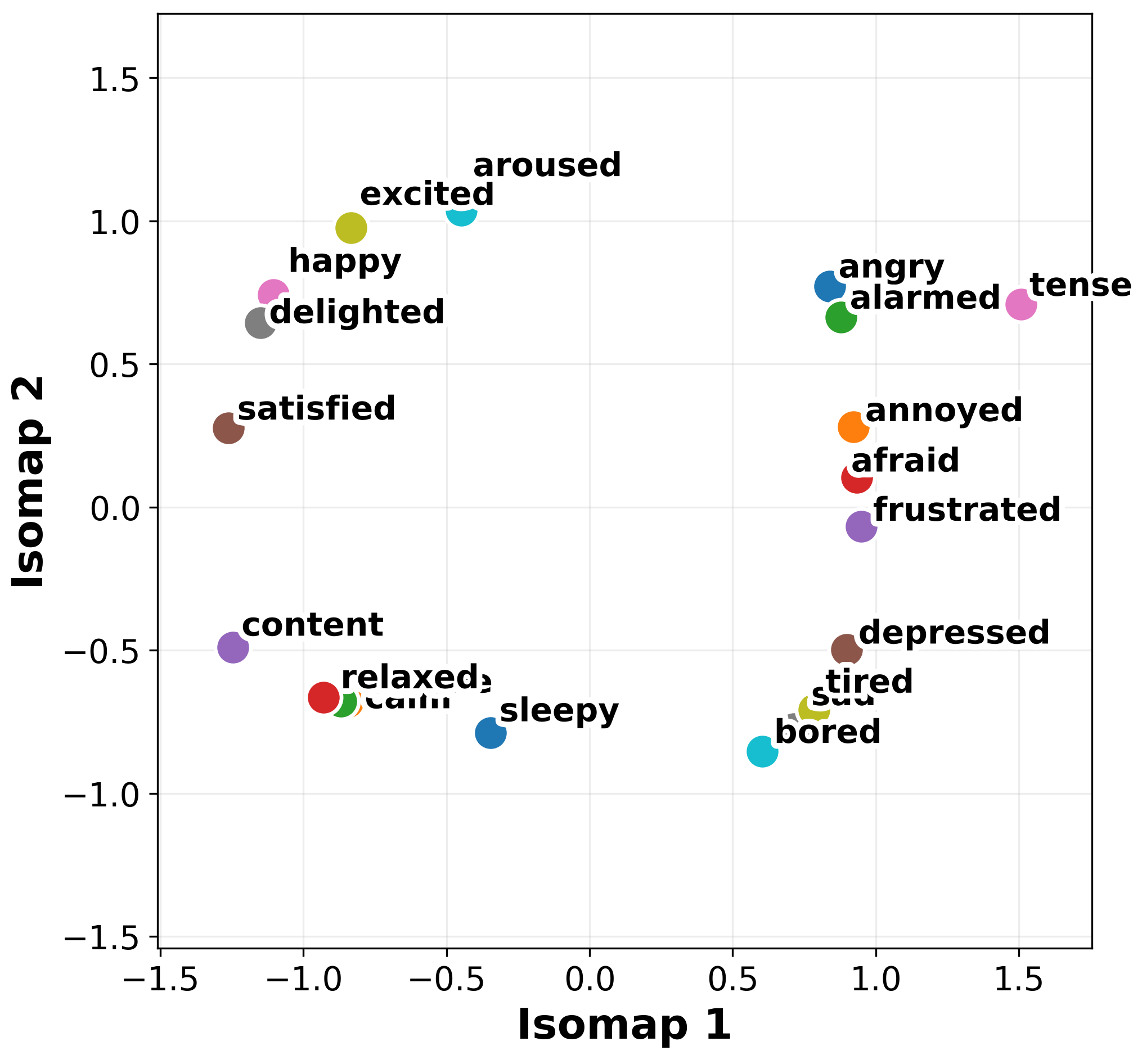} &
        \includegraphics[width=0.28\textwidth,height=0.28\textwidth]{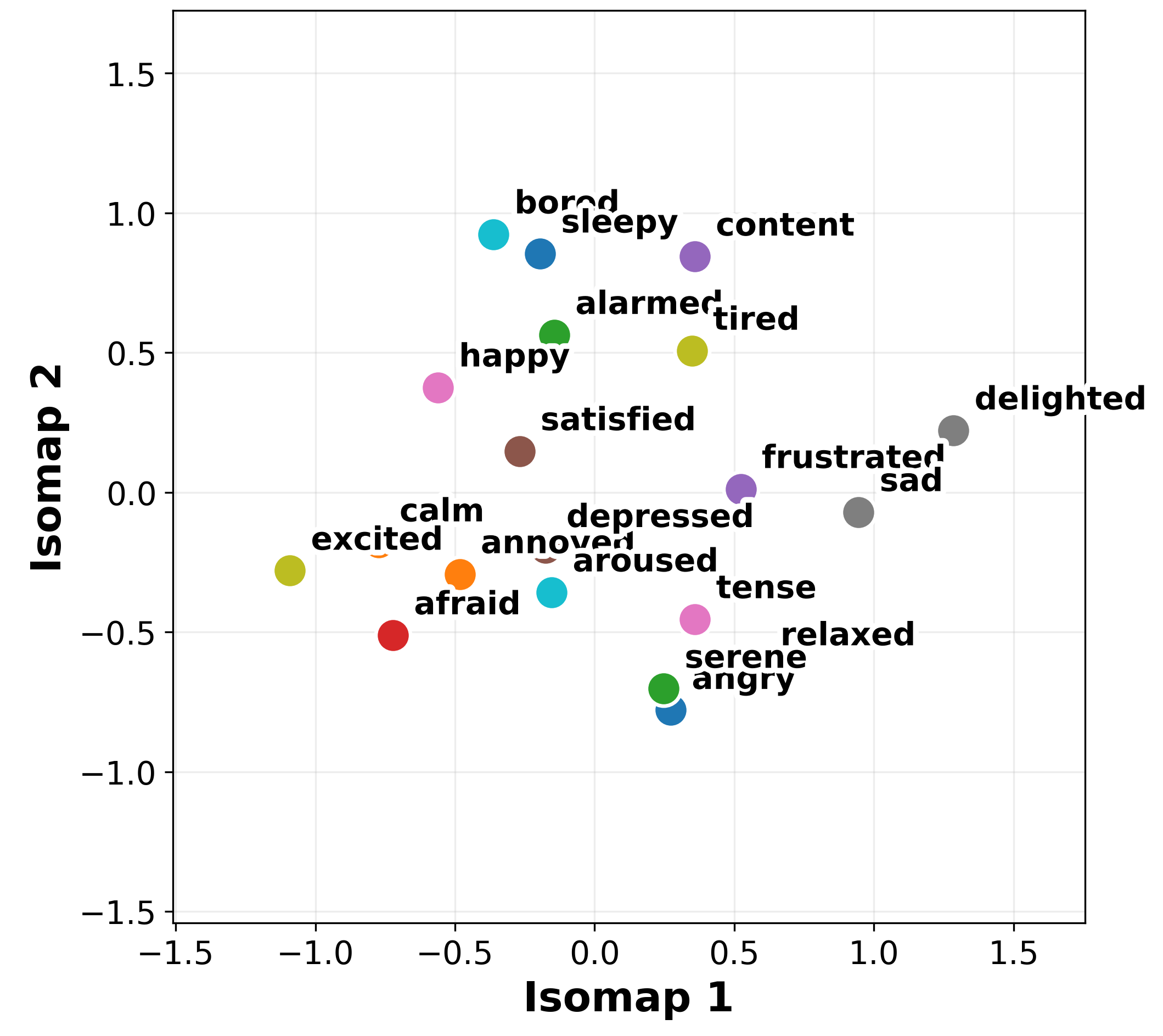} \\

        \includegraphics[width=0.28\textwidth,height=0.28\textwidth]{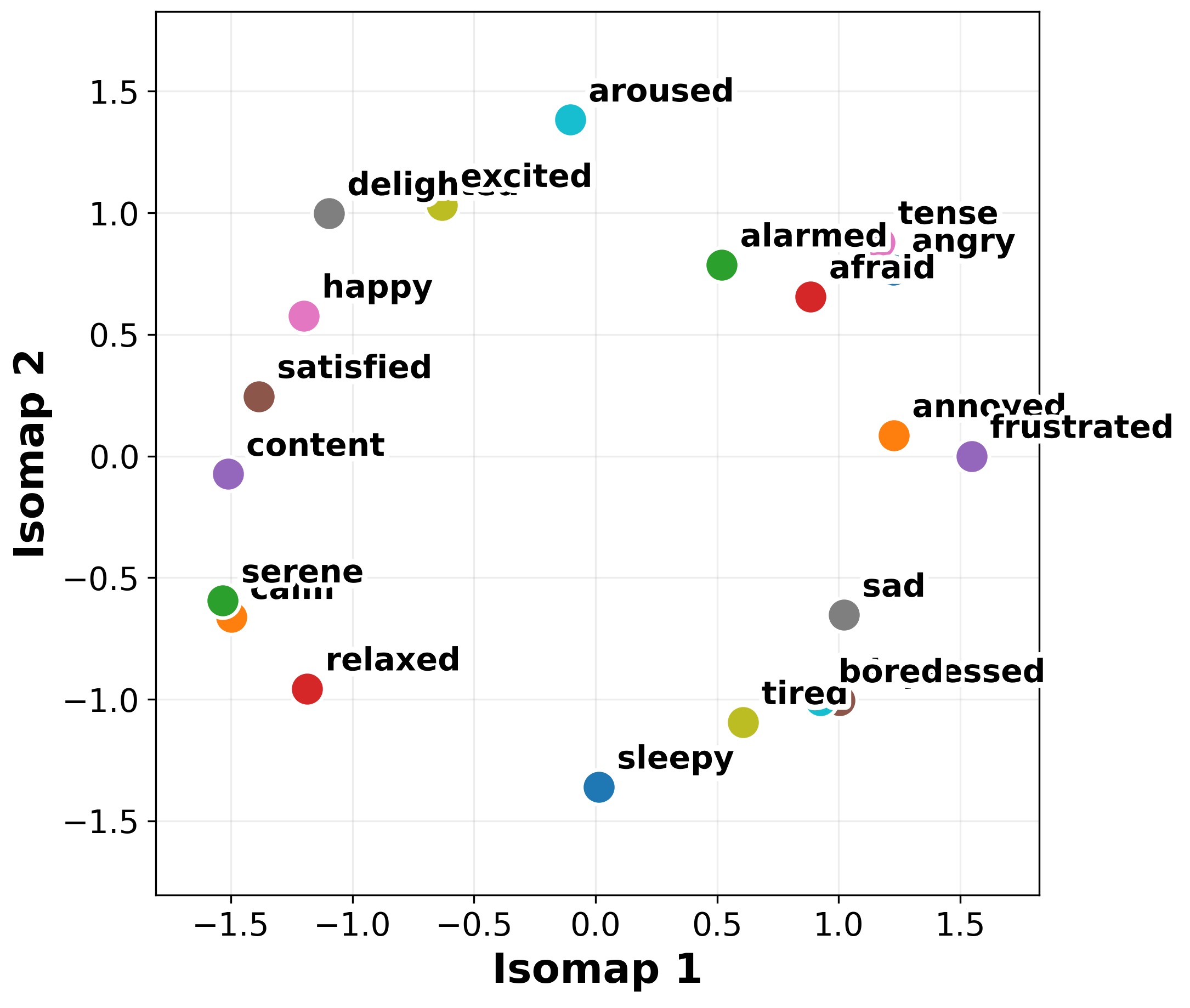} &
        \includegraphics[width=0.28\textwidth,height=0.28\textwidth]{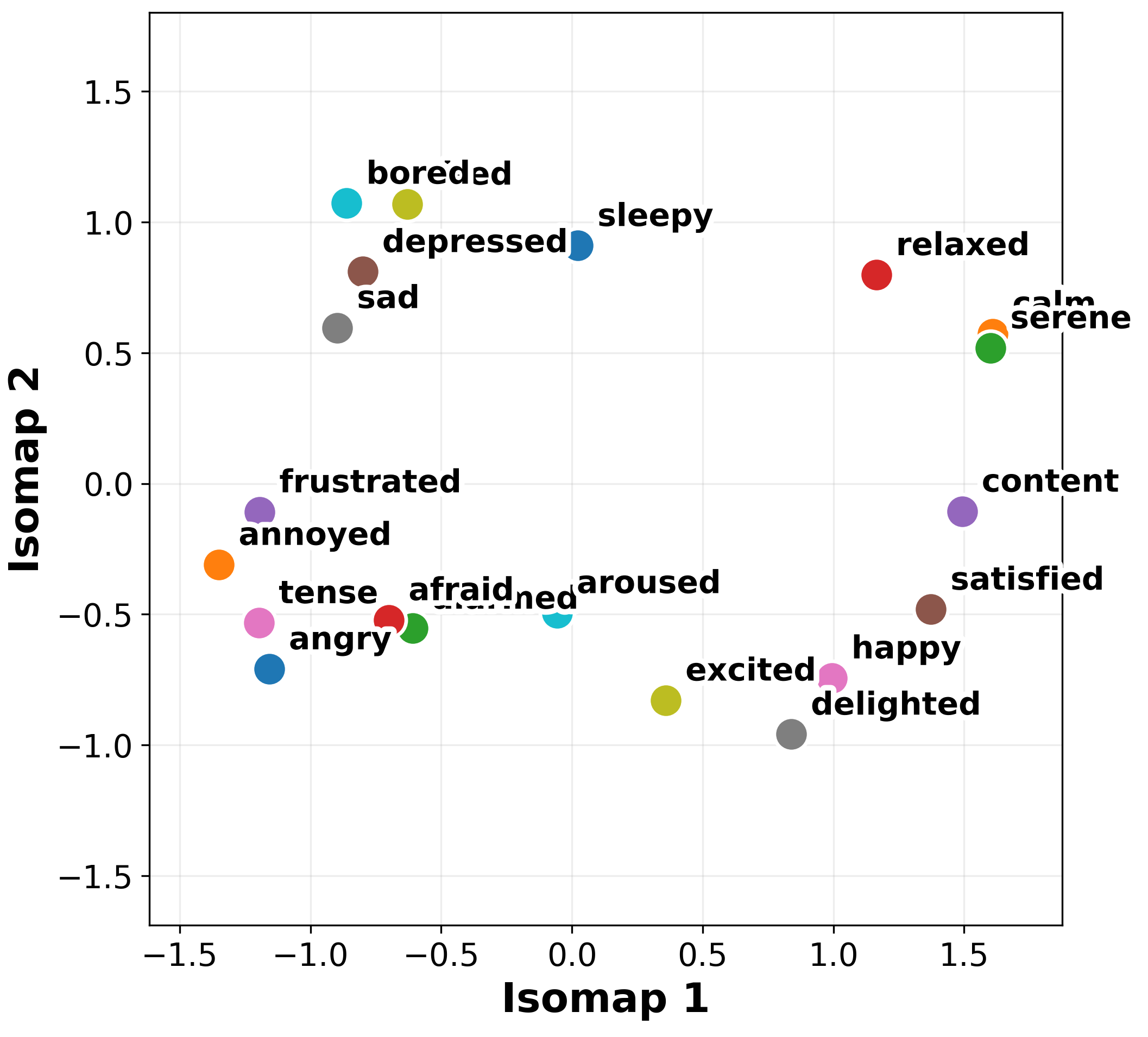}
    \end{tabular}

    \vspace{0.3em}
    \captionof{figure}{Layer-wise emergence of emotion geometry in Gemma-7B across model depth. Panels show (top-left) human perceptual geometry, (top-right) early-layer representation, (bottom-left) peak-alignment layer, and (bottom-right) final-layer isomap representation.}
    \label{fig:appendix_emo_maps_2d_iso}
\end{center}

\begin{center}
    \setlength{\tabcolsep}{2pt}
    \begin{tabular}{cc}
        \includegraphics[width=0.28\textwidth,height=0.28\textwidth]{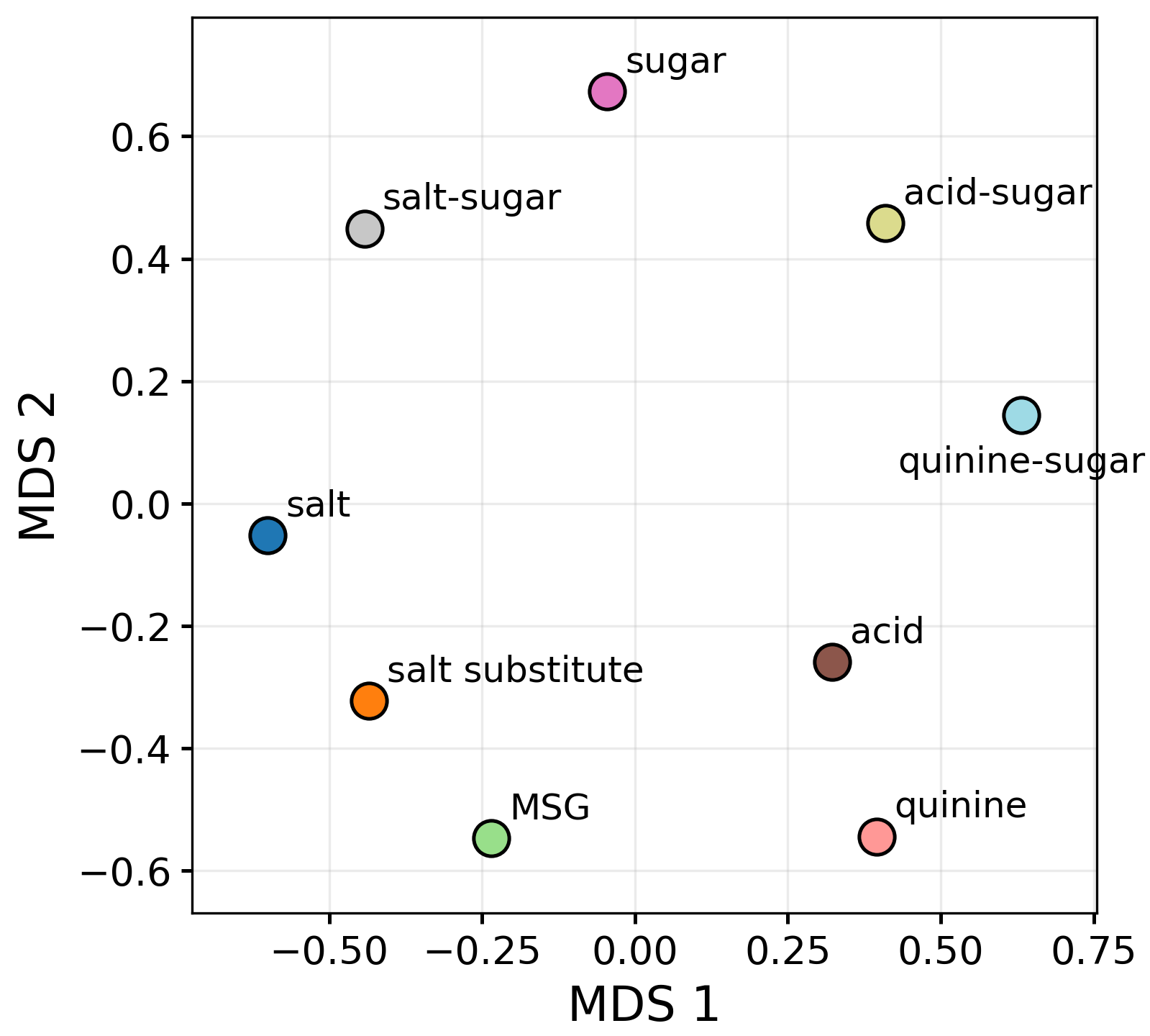} &
        \includegraphics[width=0.28\textwidth,height=0.28\textwidth]{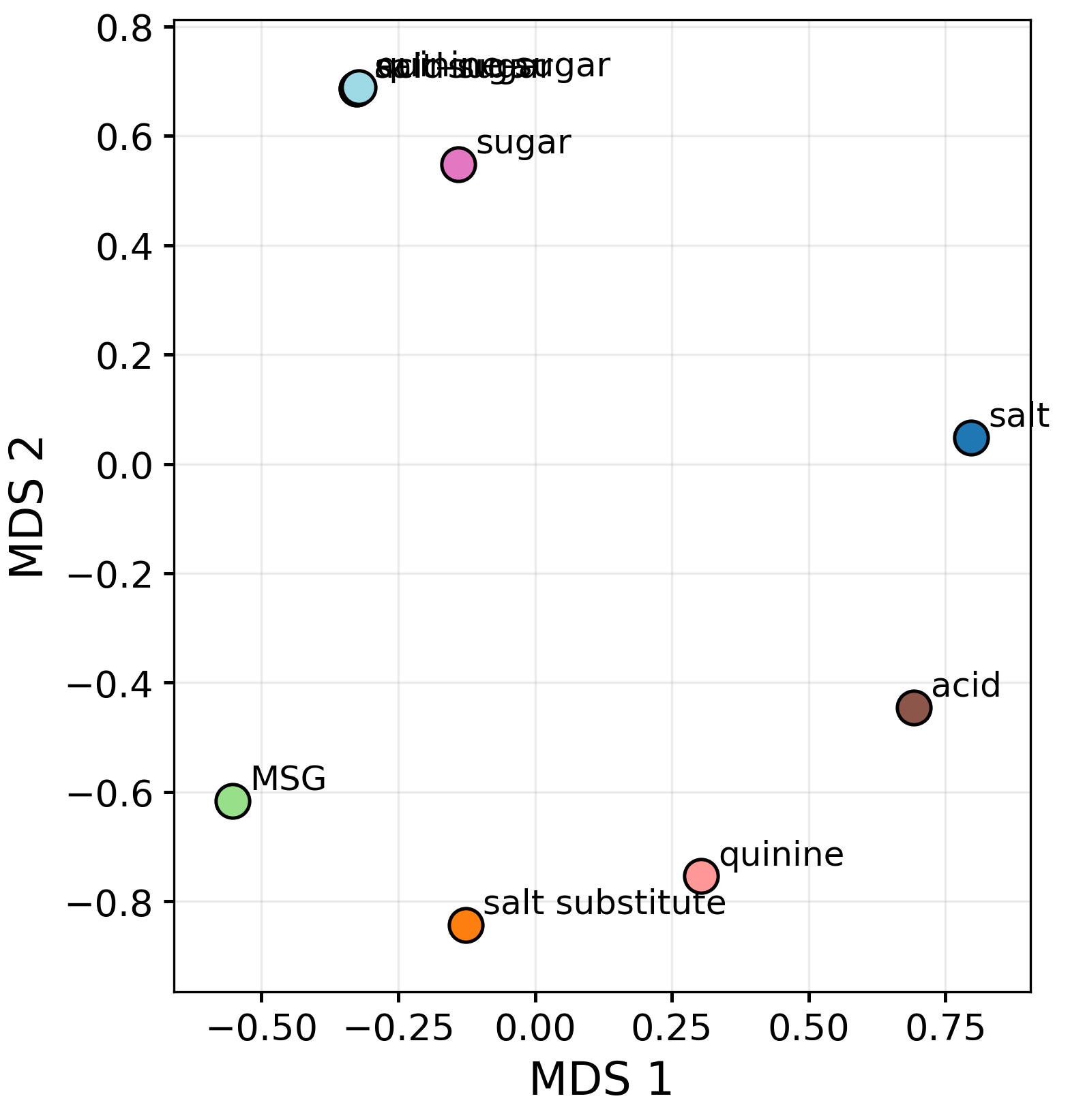} \\

        \includegraphics[width=0.28\textwidth,height=0.28\textwidth]{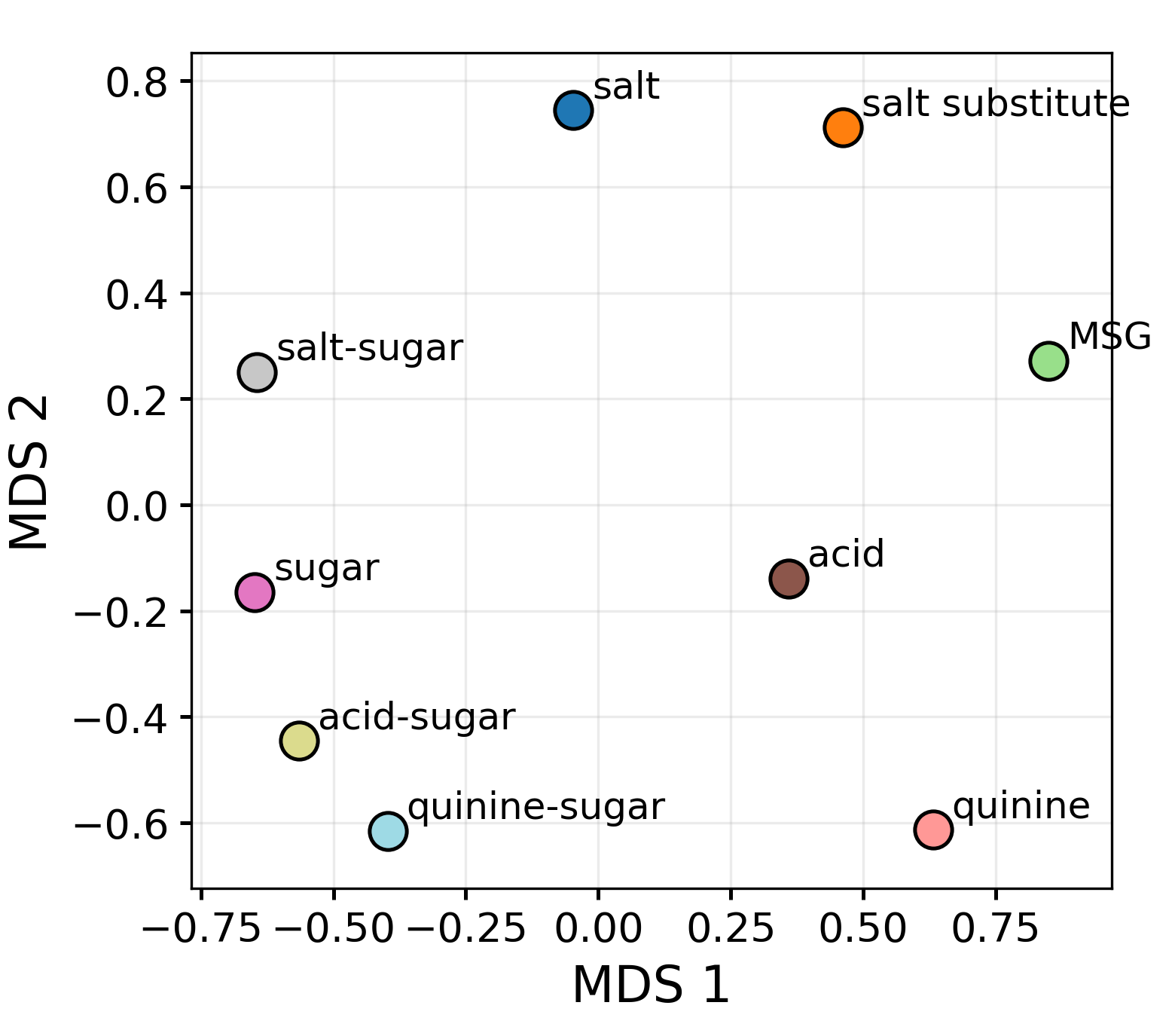} &
        \includegraphics[width=0.35\textwidth,height=0.20\textwidth]{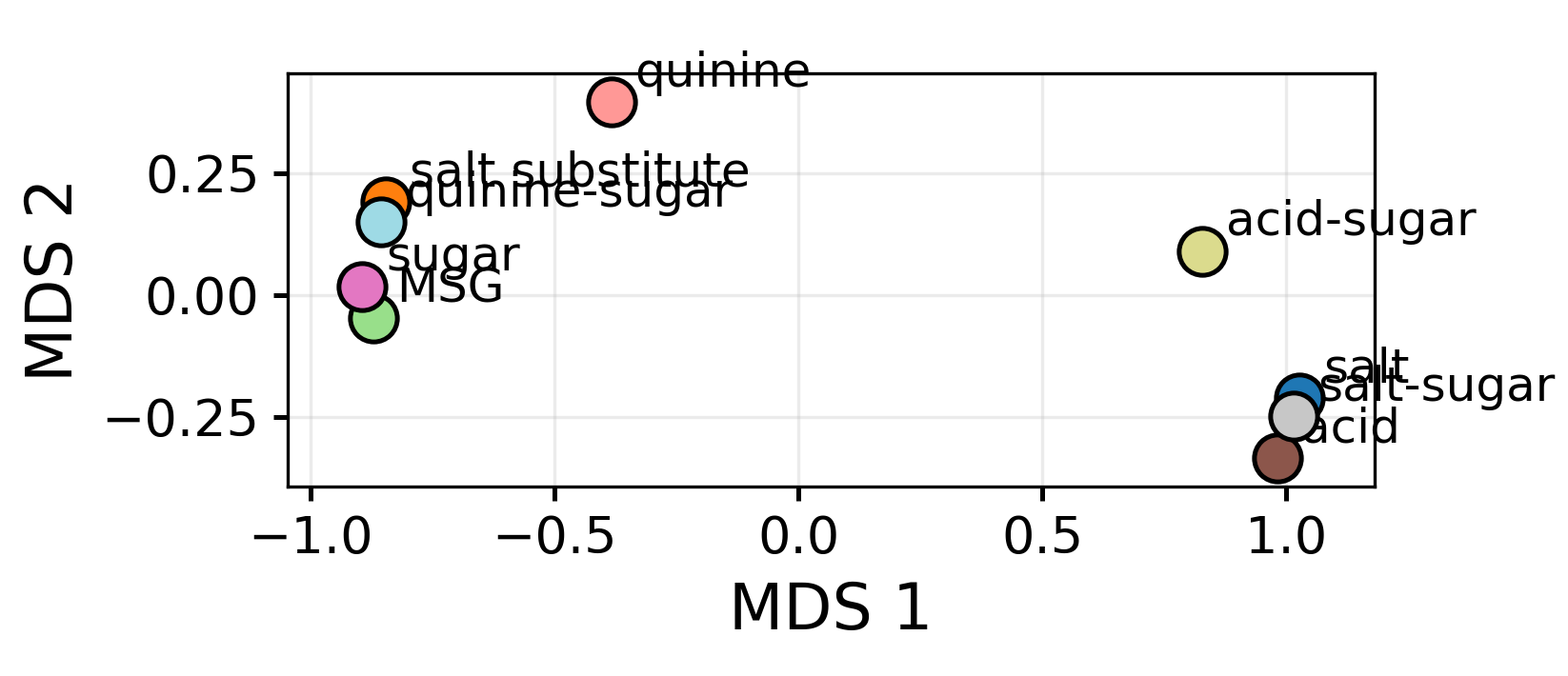}
    \end{tabular}

    \vspace{0.3em}
    \captionof{figure}{Layer-wise emergence of Taste geometry in Gemma-7B across model depth. Panels show (top-left) human perceptual geometry, (top-right) early-layer representation, (bottom-left) peak-alignment layer, and (bottom-right) final-layer MDS representation.}
    \label{fig:appendix_taste_maps_2d_mds}
\end{center}

\begin{center}
    \setlength{\tabcolsep}{2pt}
    \begin{tabular}{cc}
        \includegraphics[width=0.30\textwidth,height=0.28\textwidth]{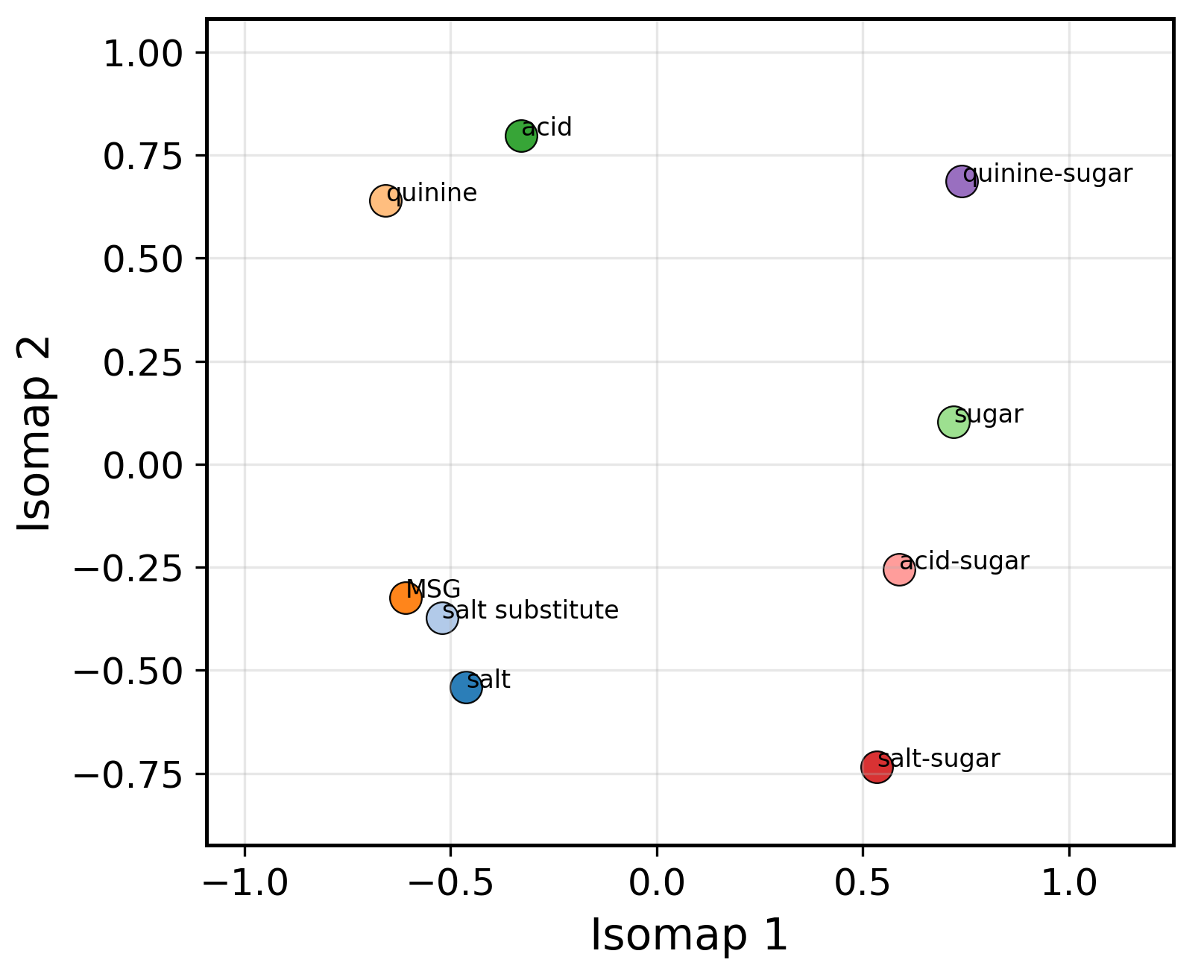} &
        \includegraphics[width=0.30\textwidth,height=0.28\textwidth]{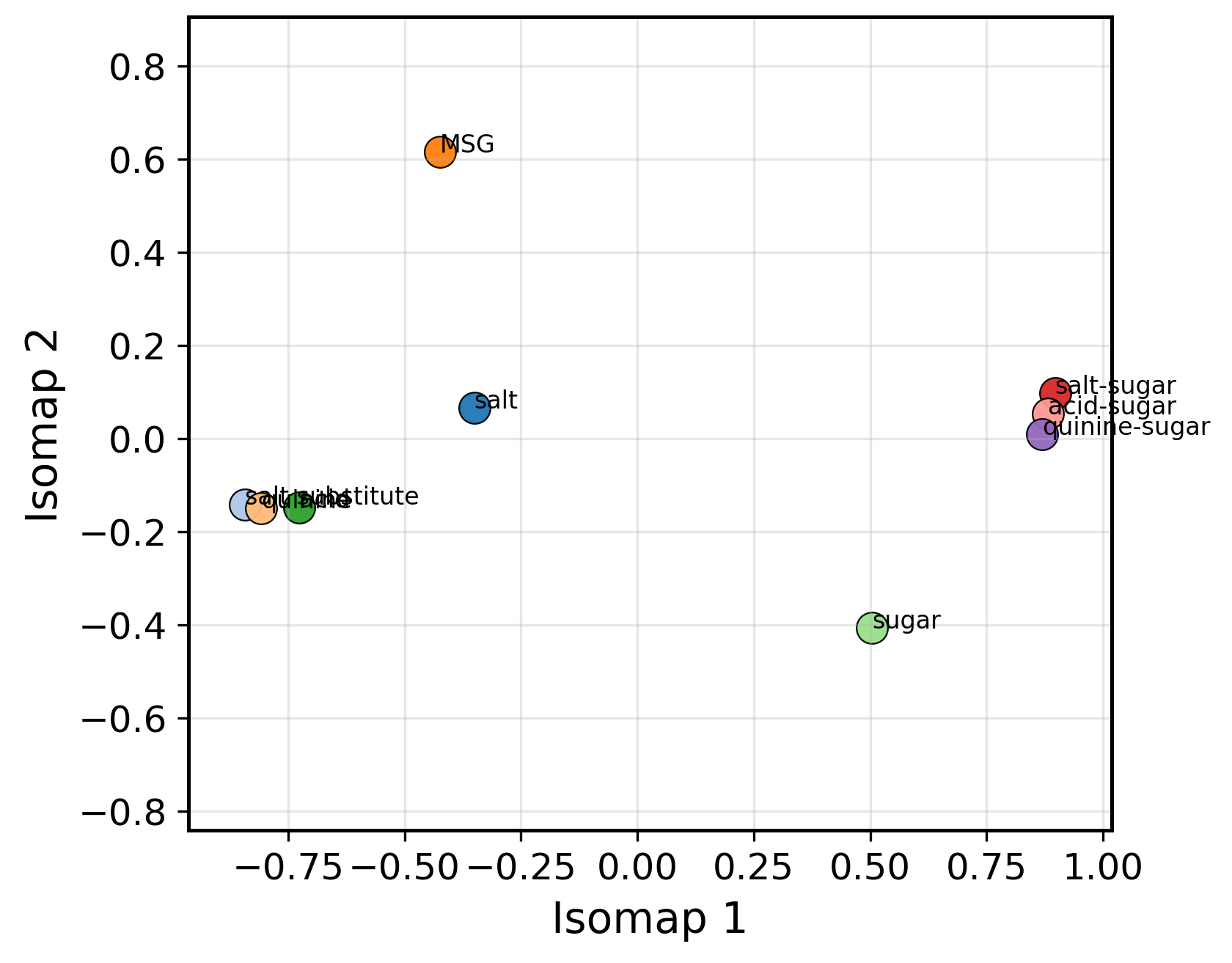} \\

        \includegraphics[width=0.30\textwidth,height=0.28\textwidth]{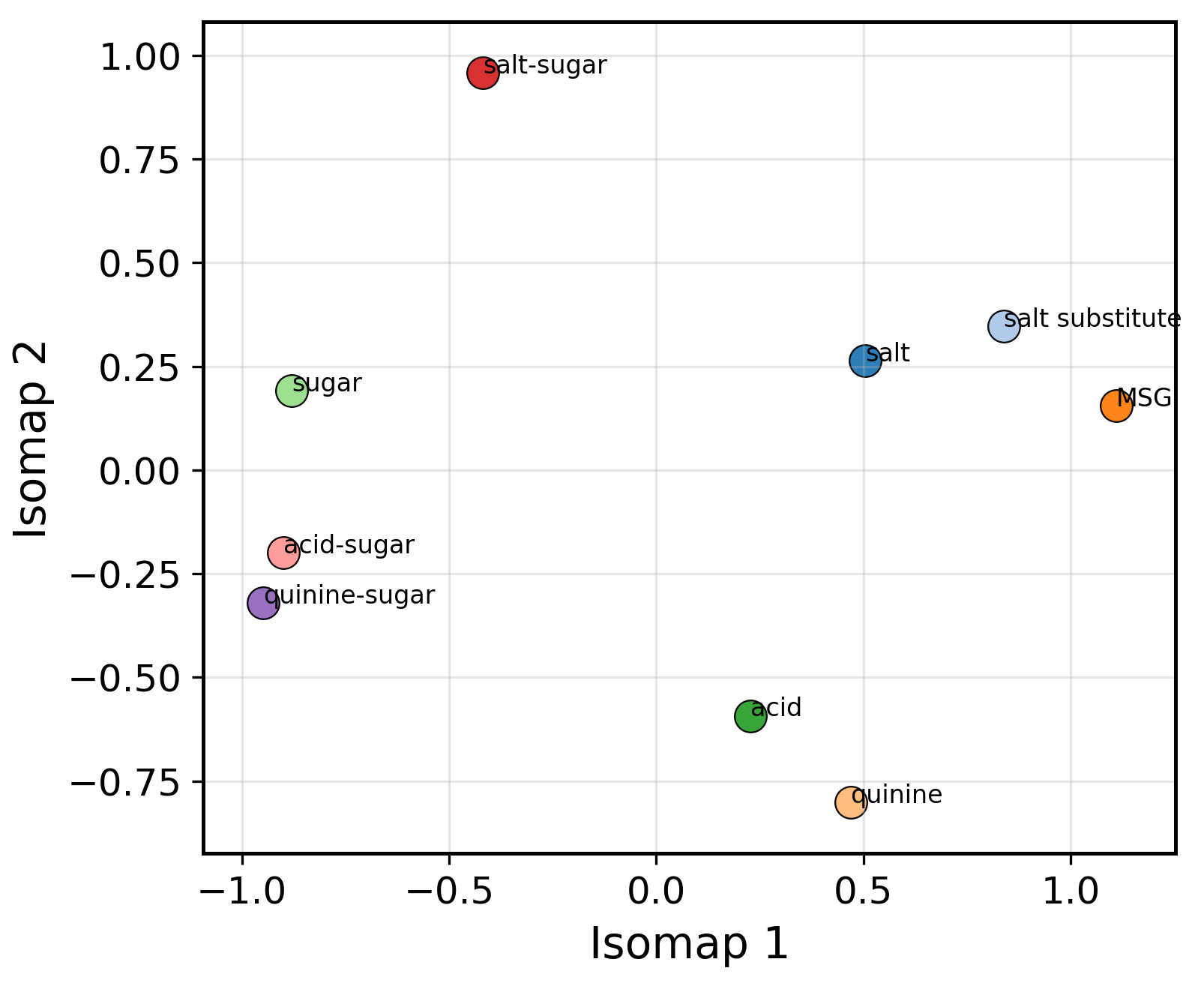} &
        \includegraphics[width=0.30\textwidth,height=0.25\textwidth]{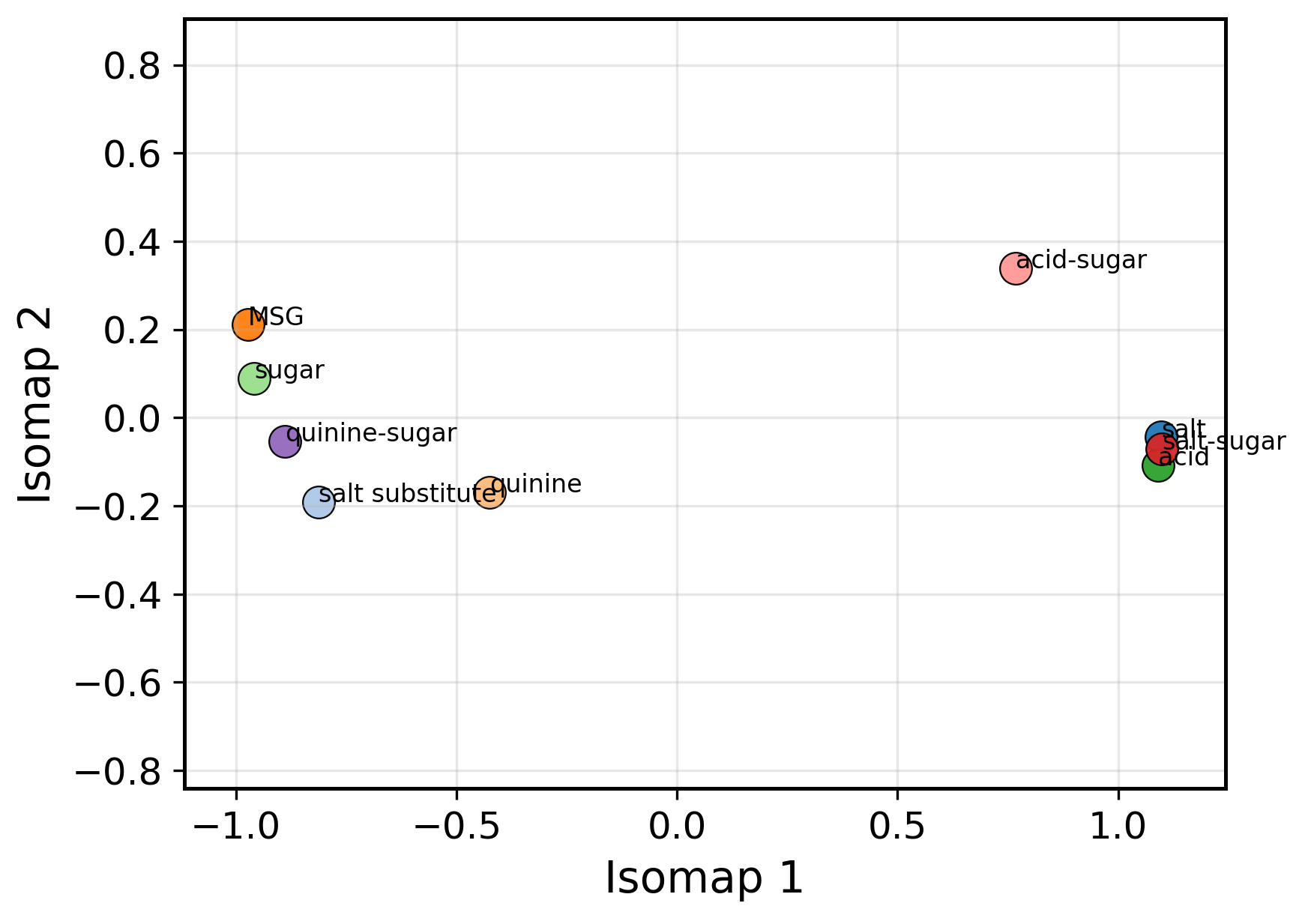}
    \end{tabular}

    \vspace{0.3em}
    \captionof{figure}{Layer-wise emergence of Taste geometry in Gemma-7B across model depth. Panels show (top-left) human perceptual geometry, (top-right) early-layer representation, (bottom-left) peak-alignment layer, and (bottom-right) final-layer isomap representation.}
    \label{fig:appendix_taste_maps_2d_isomap}
\end{center}

\begin{center}
    \setlength{\tabcolsep}{2pt}
    \begin{tabular}{cc}
        \includegraphics[width=0.28\textwidth,height=0.28\textwidth]{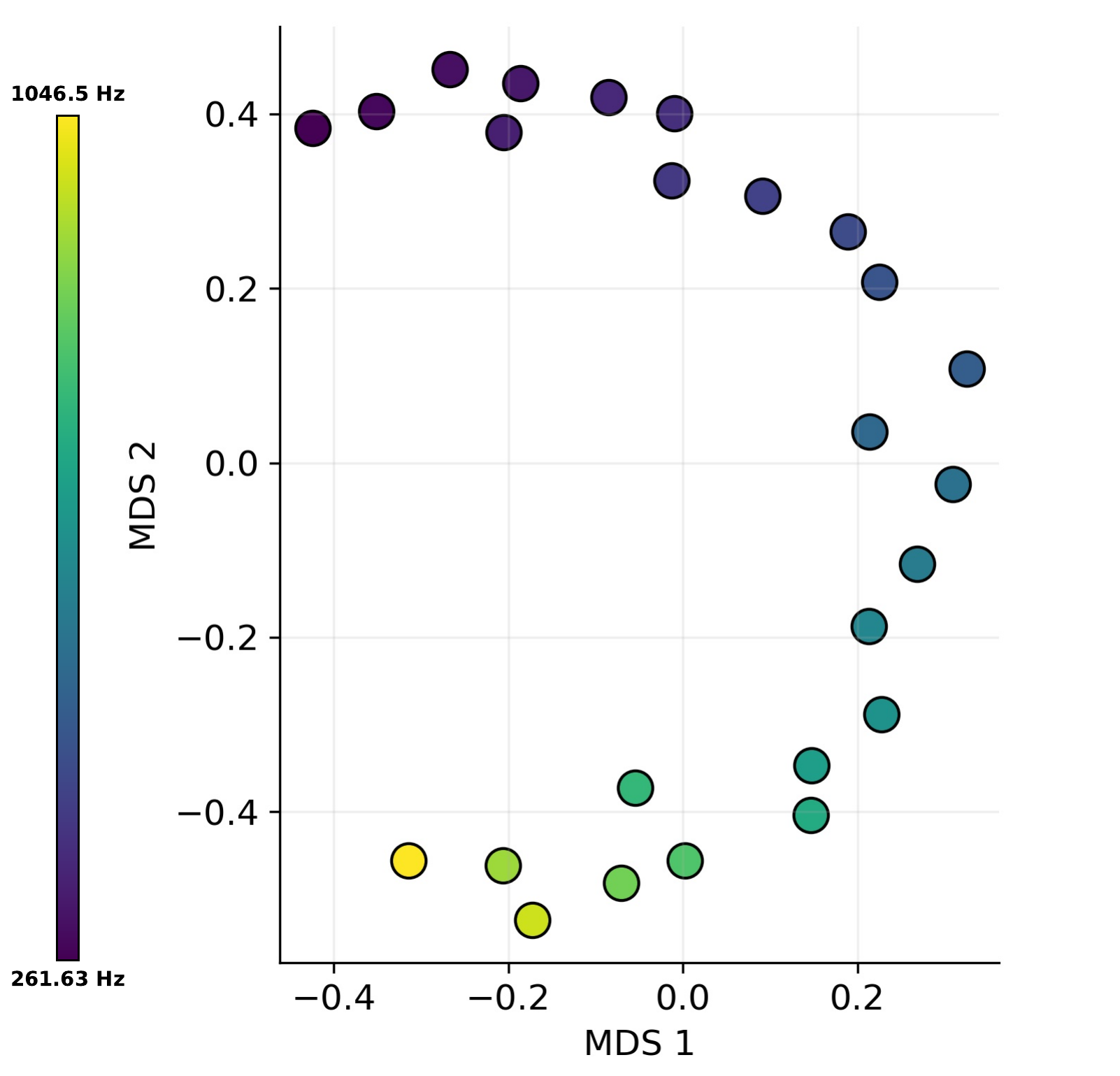} &
        \includegraphics[width=0.28\textwidth,height=0.28\textwidth]{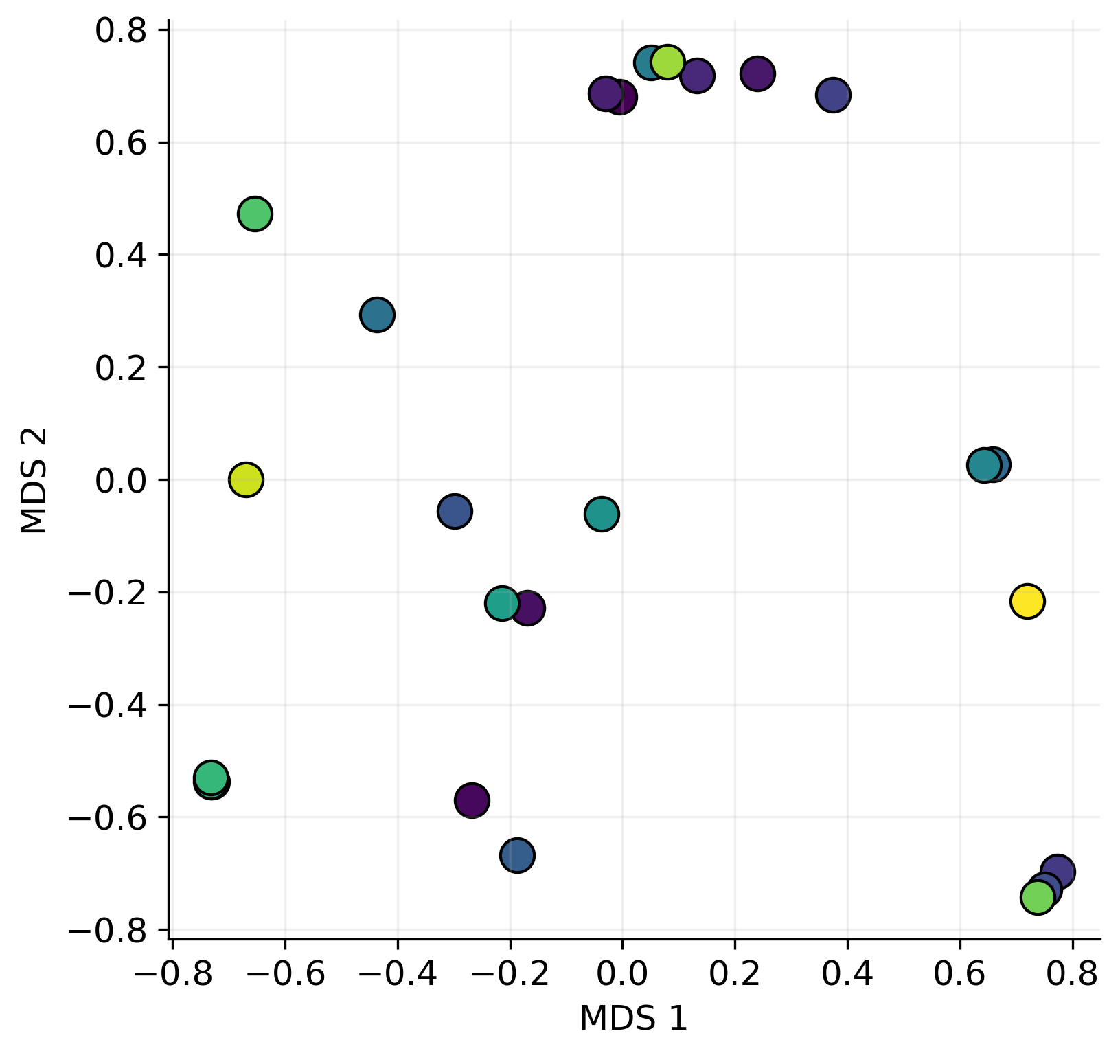} \\

        \includegraphics[width=0.28\textwidth,height=0.28\textwidth]{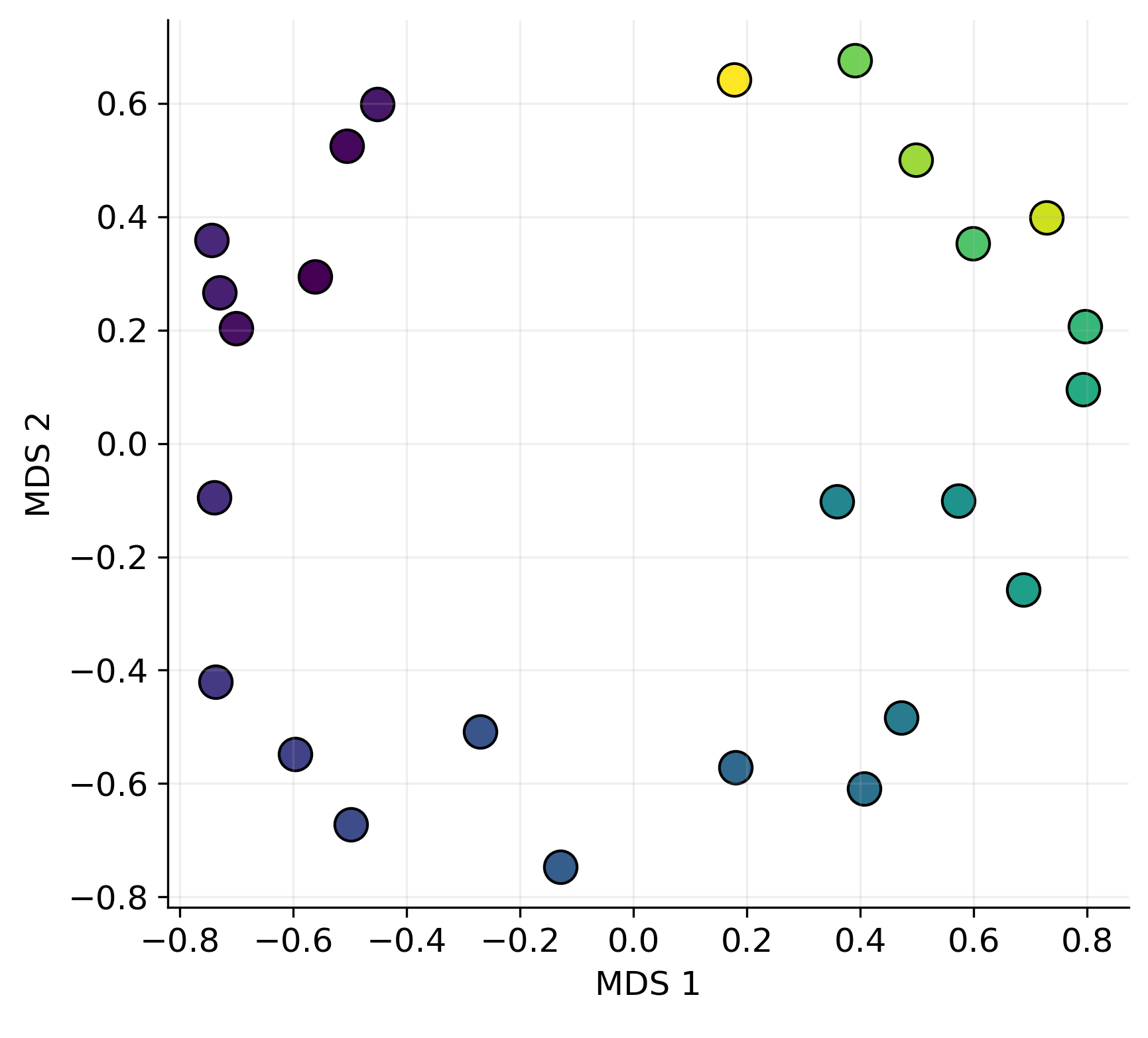} &
        \includegraphics[width=0.28\textwidth,height=0.28\textwidth]{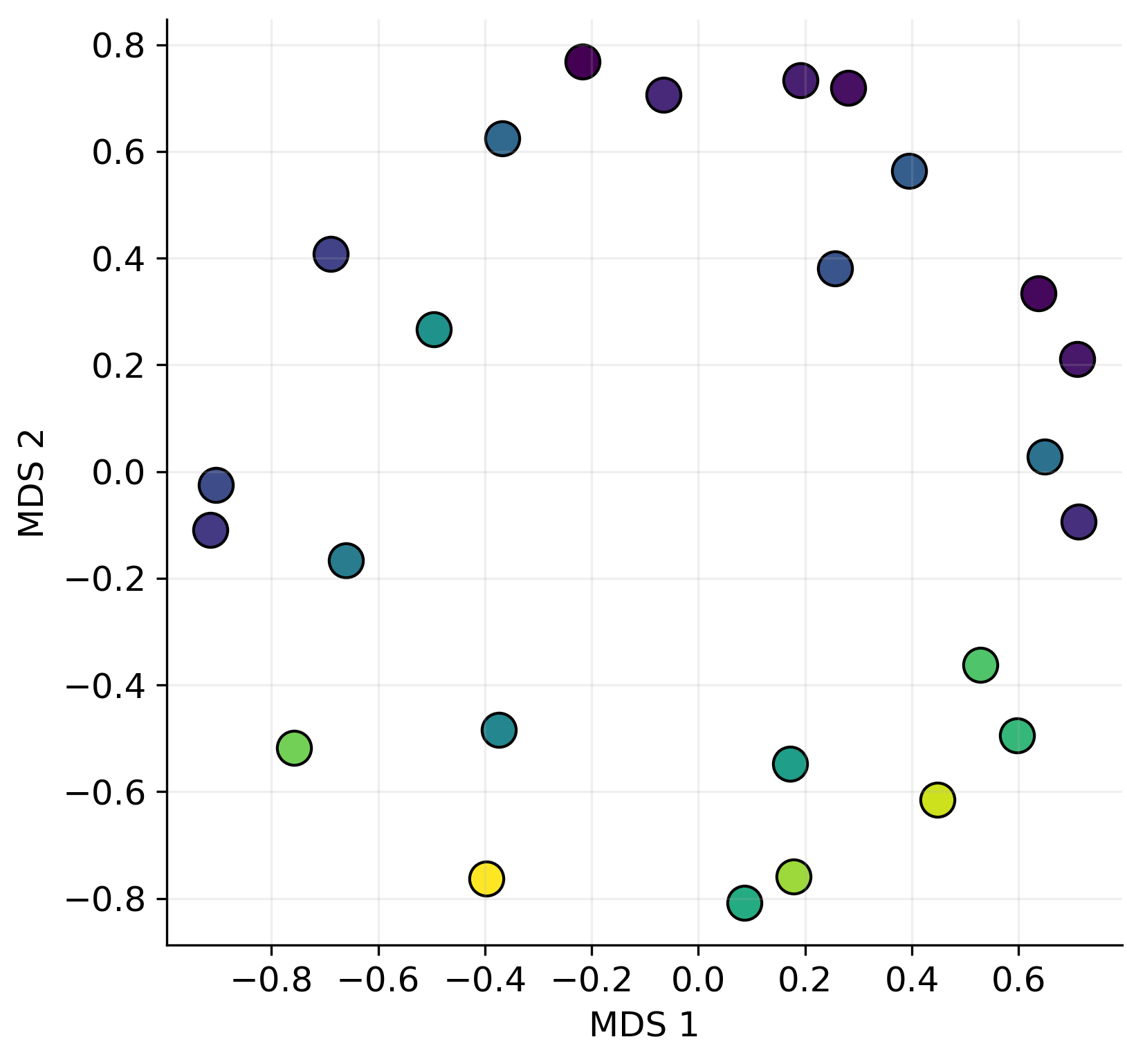}
    \end{tabular}

    \vspace{0.3em}
    \captionof{figure}{Layer-wise emergence of 2D Pitch geometry in Qwen-3-4B across model depth. Panels show (top-left) human perceptual geometry, (top-right) early-layer representation, (bottom-left) peak-alignment layer, and (bottom-right) final-layer MDS representation.}
    \label{fig:appendix_pitch_maps_2d}
\end{center}

\begin{center}
    \setlength{\tabcolsep}{2pt}
    \begin{tabular}{cc}
        \includegraphics[width=0.28\textwidth,height=0.28\textwidth]{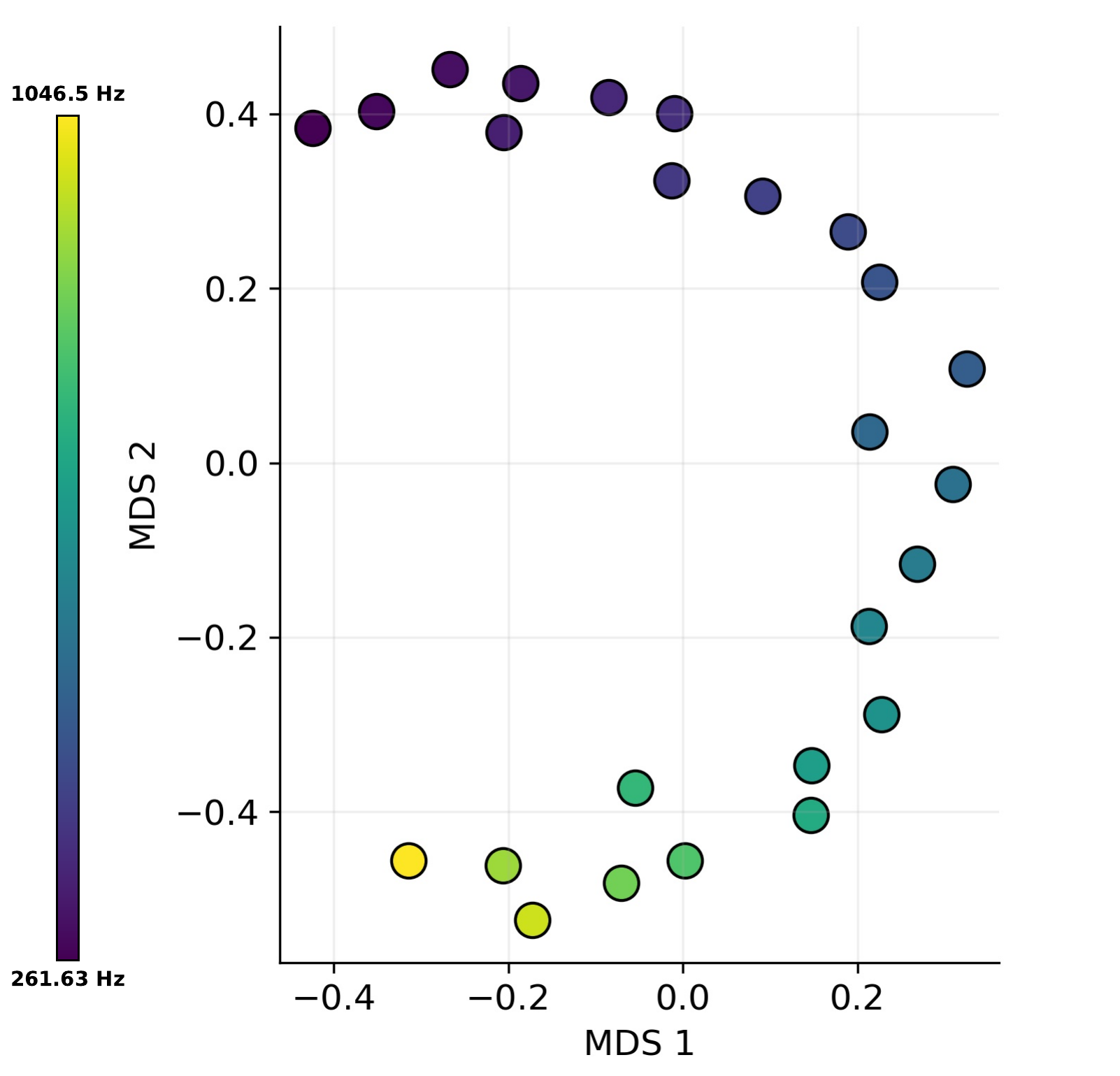} &
        \includegraphics[width=0.28\textwidth,height=0.28\textwidth]{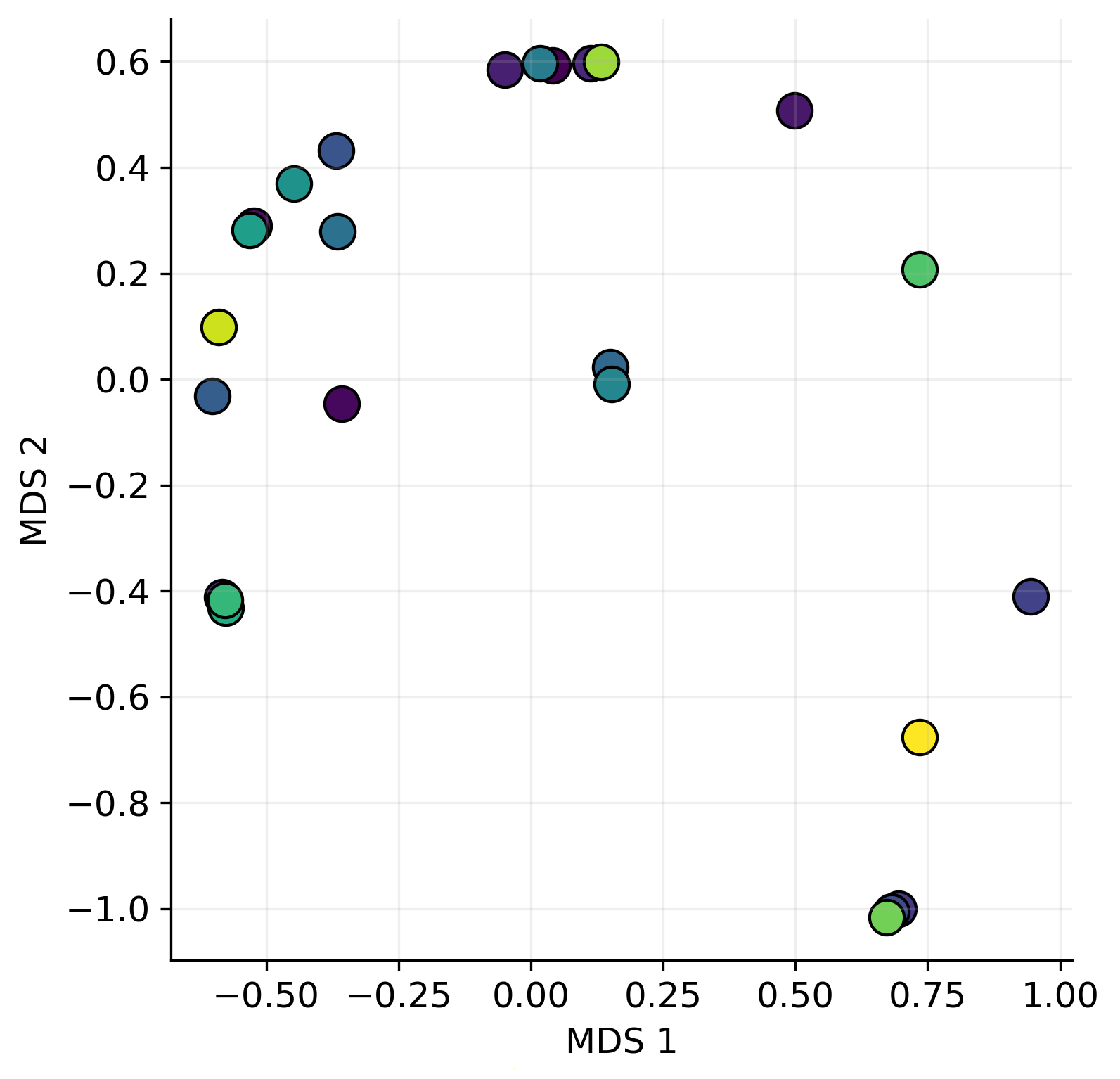} \\

        \includegraphics[width=0.28\textwidth,height=0.28\textwidth]{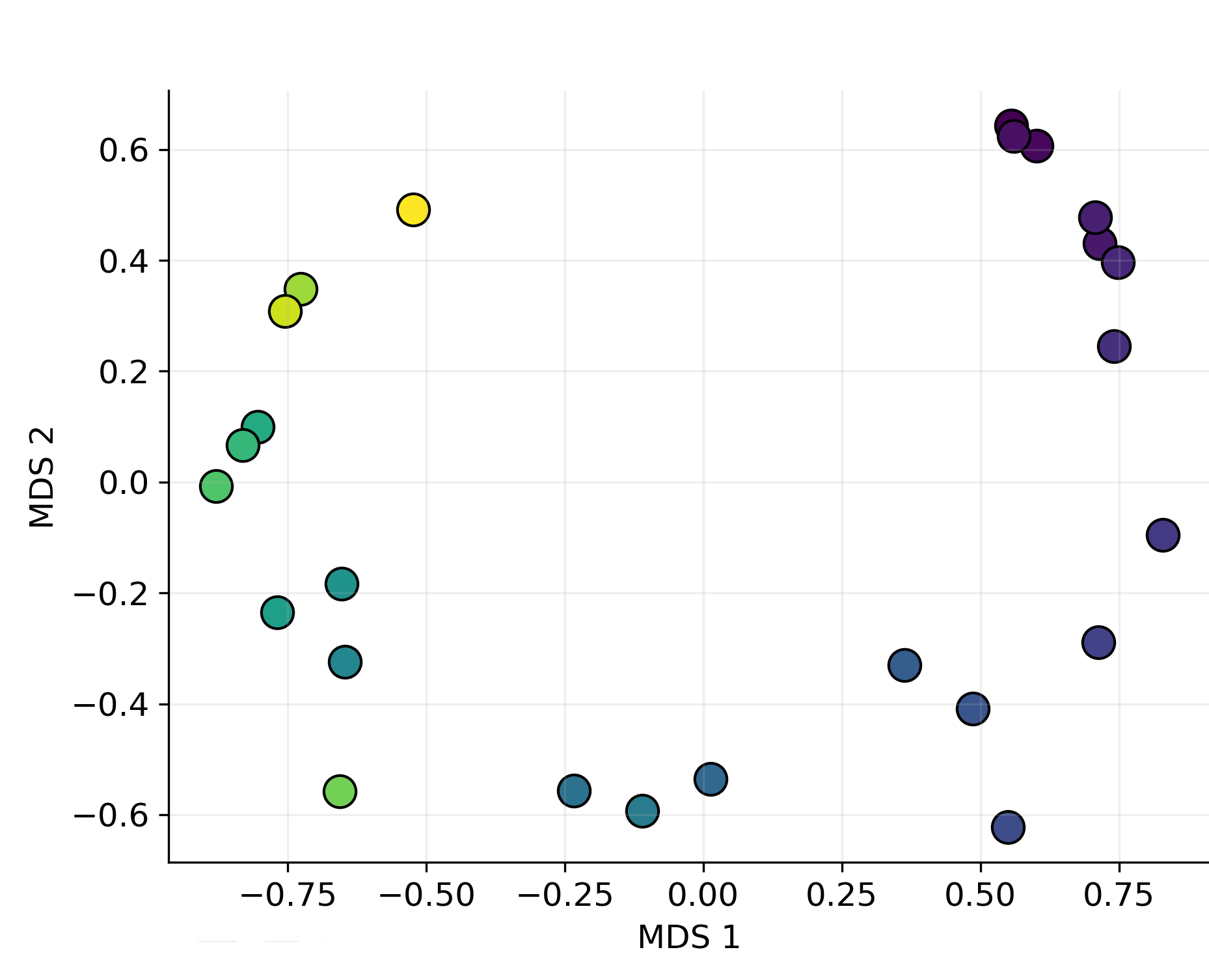} &
        \includegraphics[width=0.28\textwidth,height=0.28\textwidth]{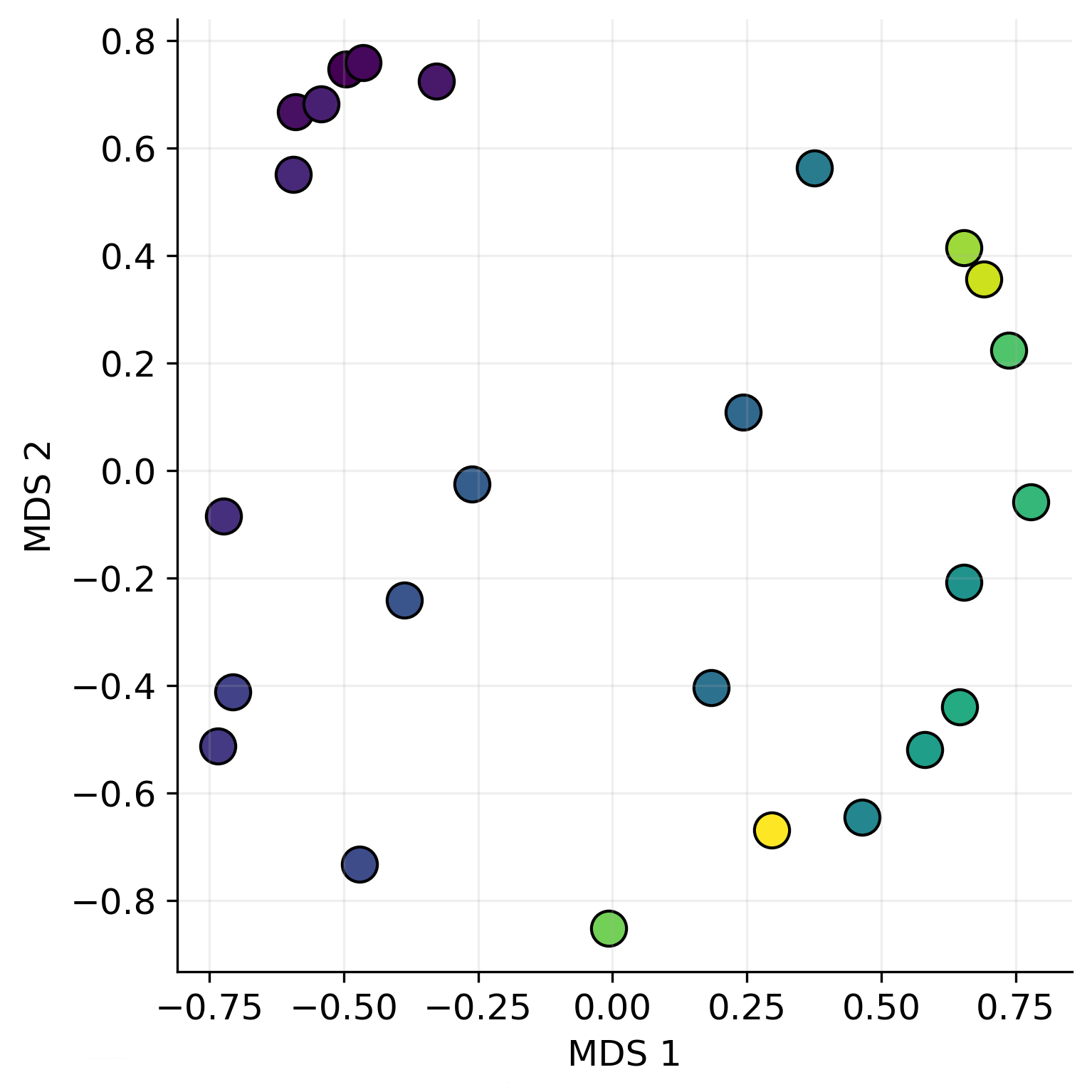}
    \end{tabular}

    \vspace{0.3em}
    \captionof{figure}{Layer-wise emergence of 2D Pitch geometry in Gemma-7B across model depth. Panels show (top-left) human perceptual geometry, (top-right) early-layer representation, (bottom-left) peak-alignment layer, and (bottom-right) final-layer MDS representation.}
    \label{fig:appendix_pitch_maps_2d_gemma}
\end{center}

\begin{center}
    \setlength{\tabcolsep}{2pt}
    \begin{tabular}{cc}
        \includegraphics[width=0.35\textwidth,height=0.30\textwidth]{Figures/Pitch/Pitch-main/mds/gemma7bit/3d/human-3d-g_colorbar_v4.pdf} &
        \includegraphics[width=0.30\textwidth,height=0.30\textwidth]{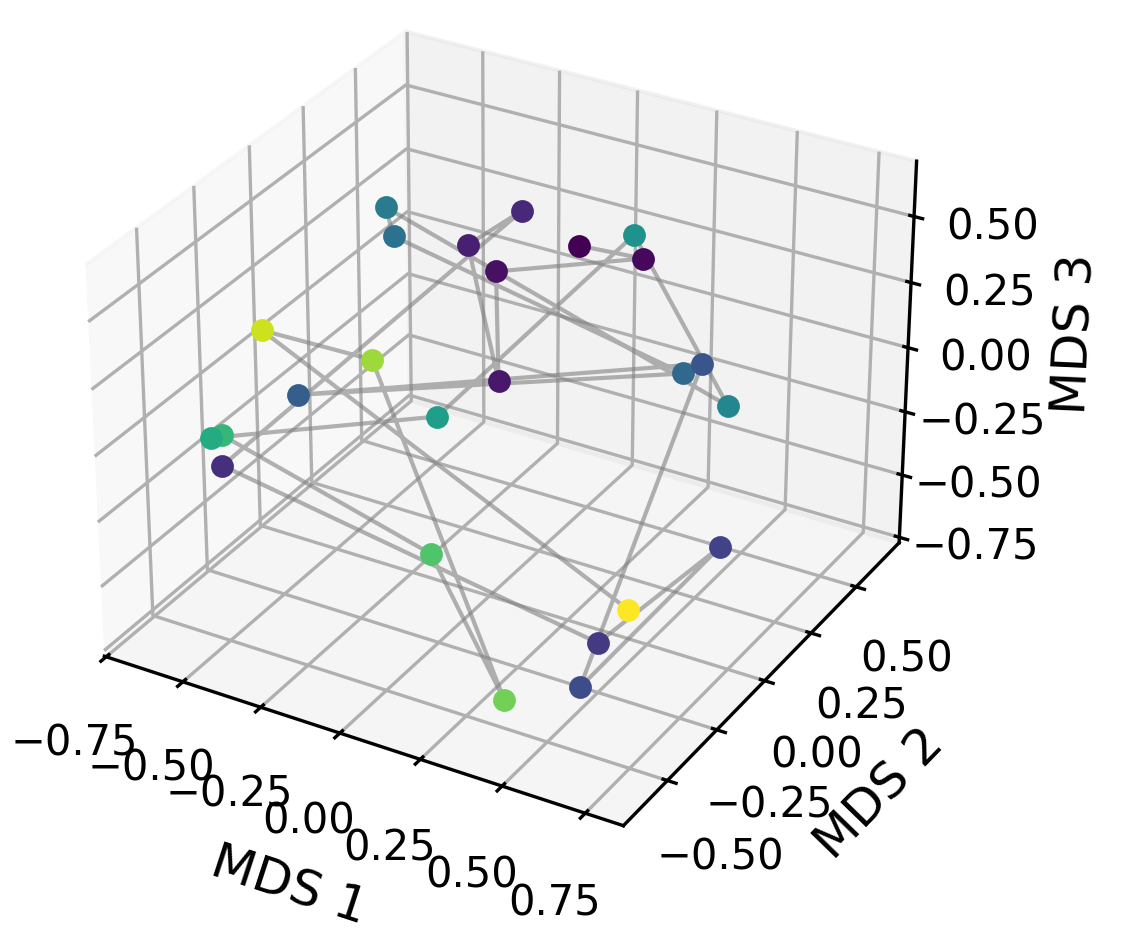} \\

        \includegraphics[width=0.30\textwidth,height=0.30\textwidth]{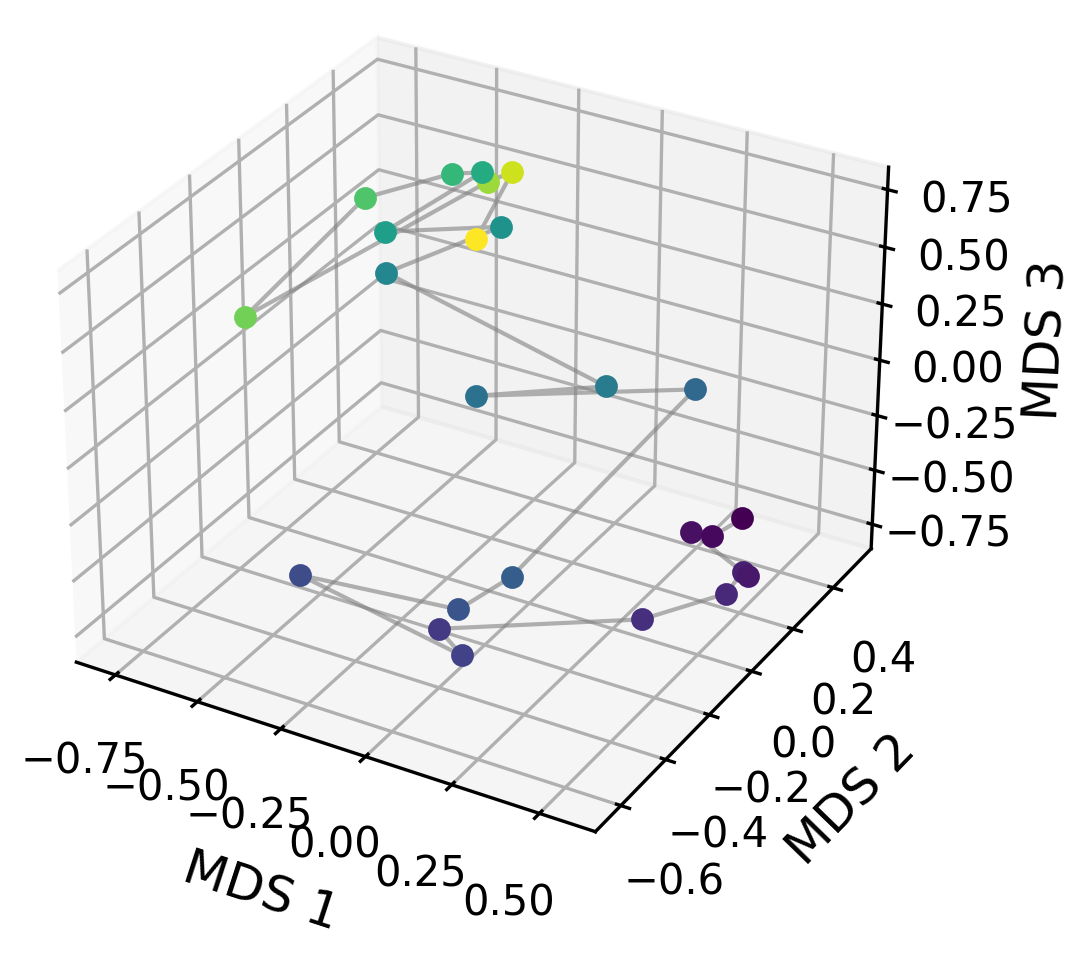} &
        \includegraphics[width=0.30\textwidth,height=0.30\textwidth]{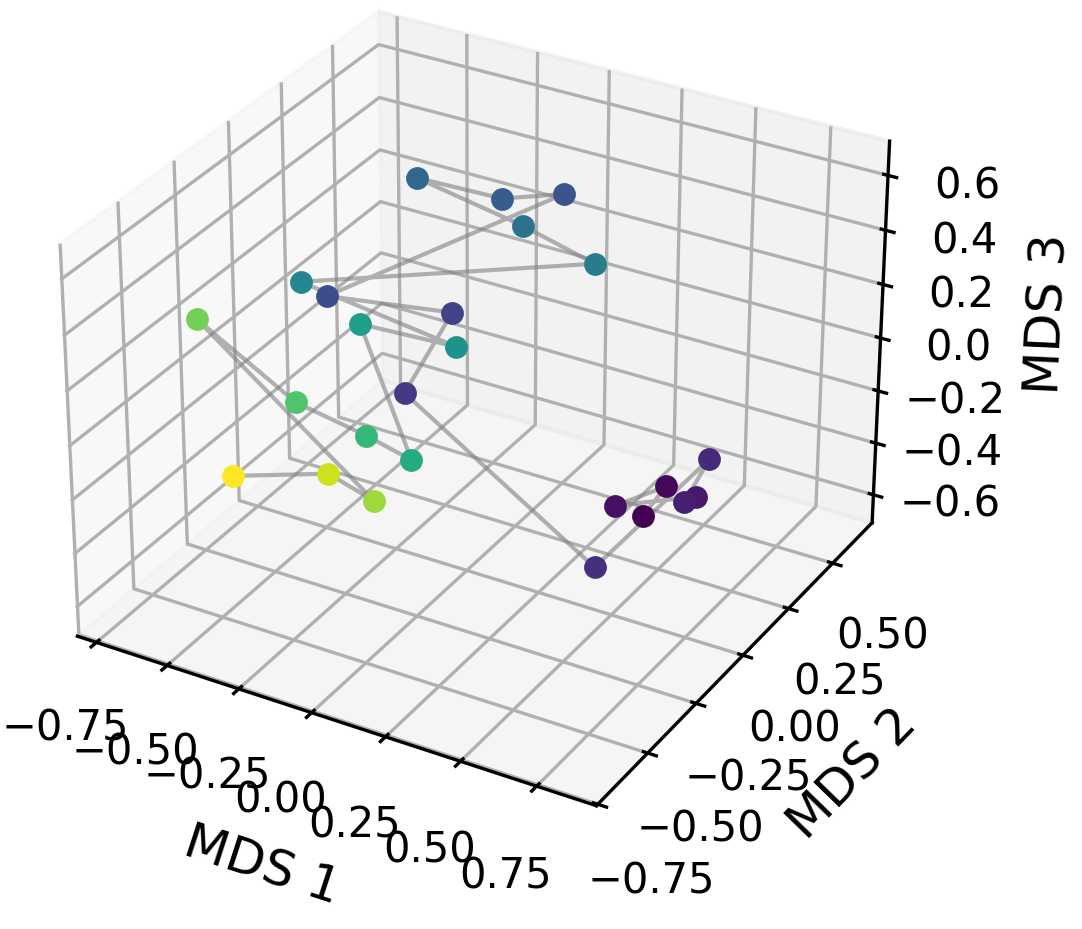}
    \end{tabular}

    \vspace{0.3em}
    \captionof{figure}{Layer-wise emergence of 3D Pitch geometry in Qwen-3-4B across model depth. Panels show (top-left) human perceptual geometry, (top-right) early-layer representation, (bottom-left) peak-alignment layer, and (bottom-right) final-layer MDS representation.}
    \label{fig:appendix_pitch_maps_3d_qwen}
\end{center}

\begin{center}
    \setlength{\tabcolsep}{2pt}
    \begin{tabular}{cc}
        \includegraphics[width=0.35\textwidth,height=0.30\textwidth]{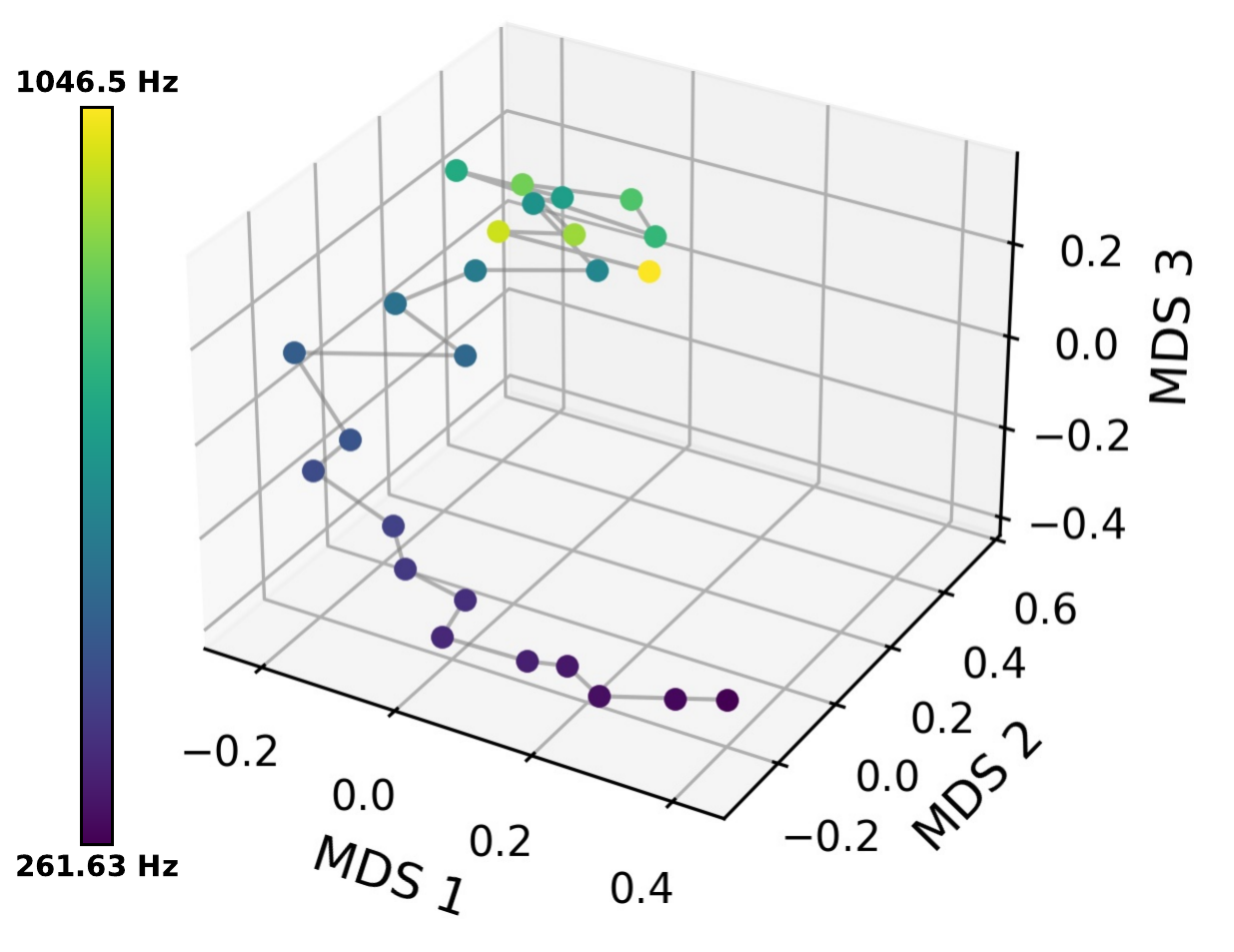} &
        \includegraphics[width=0.30\textwidth,height=0.30\textwidth]{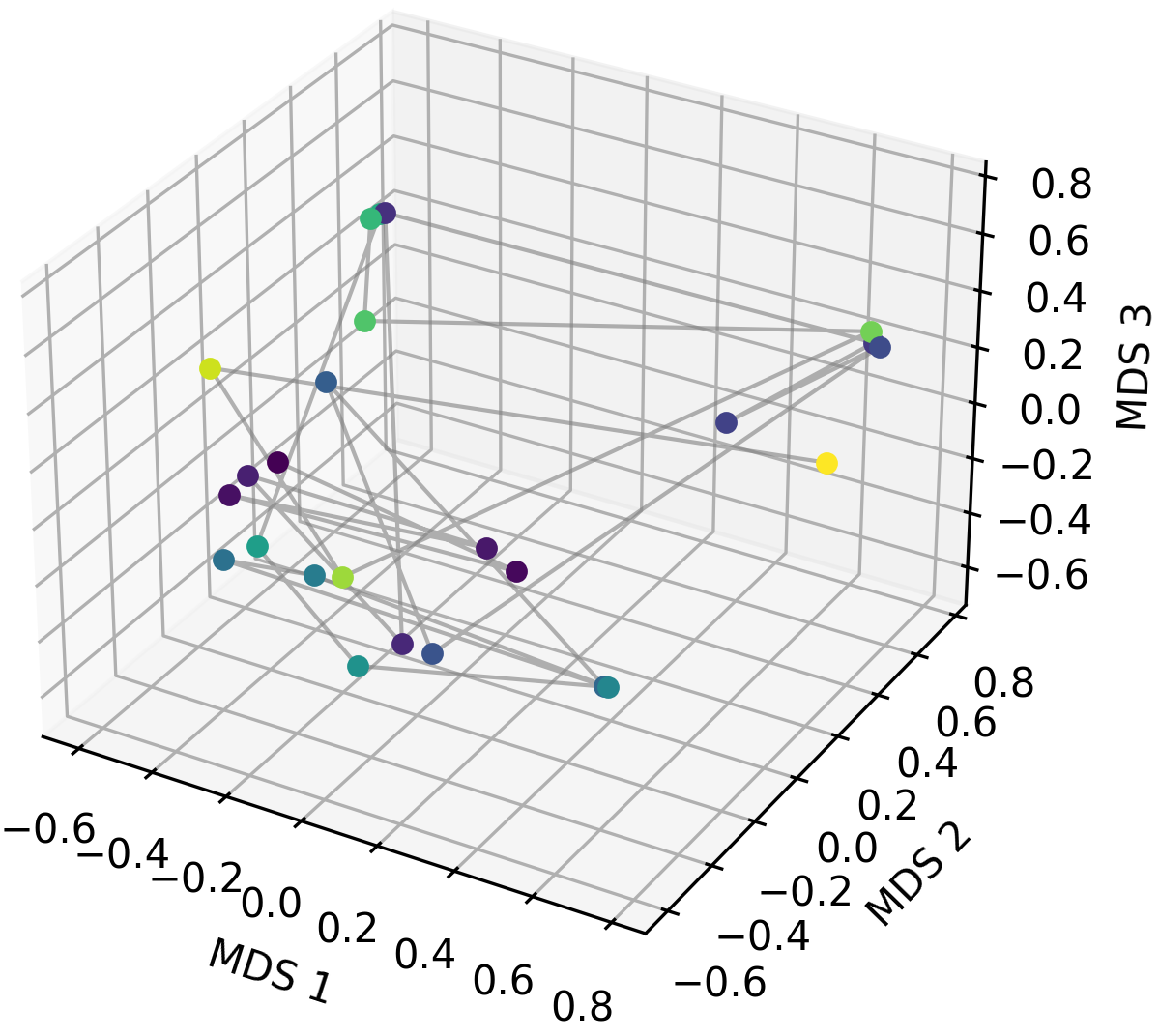} \\

        \includegraphics[width=0.30\textwidth,height=0.30\textwidth]{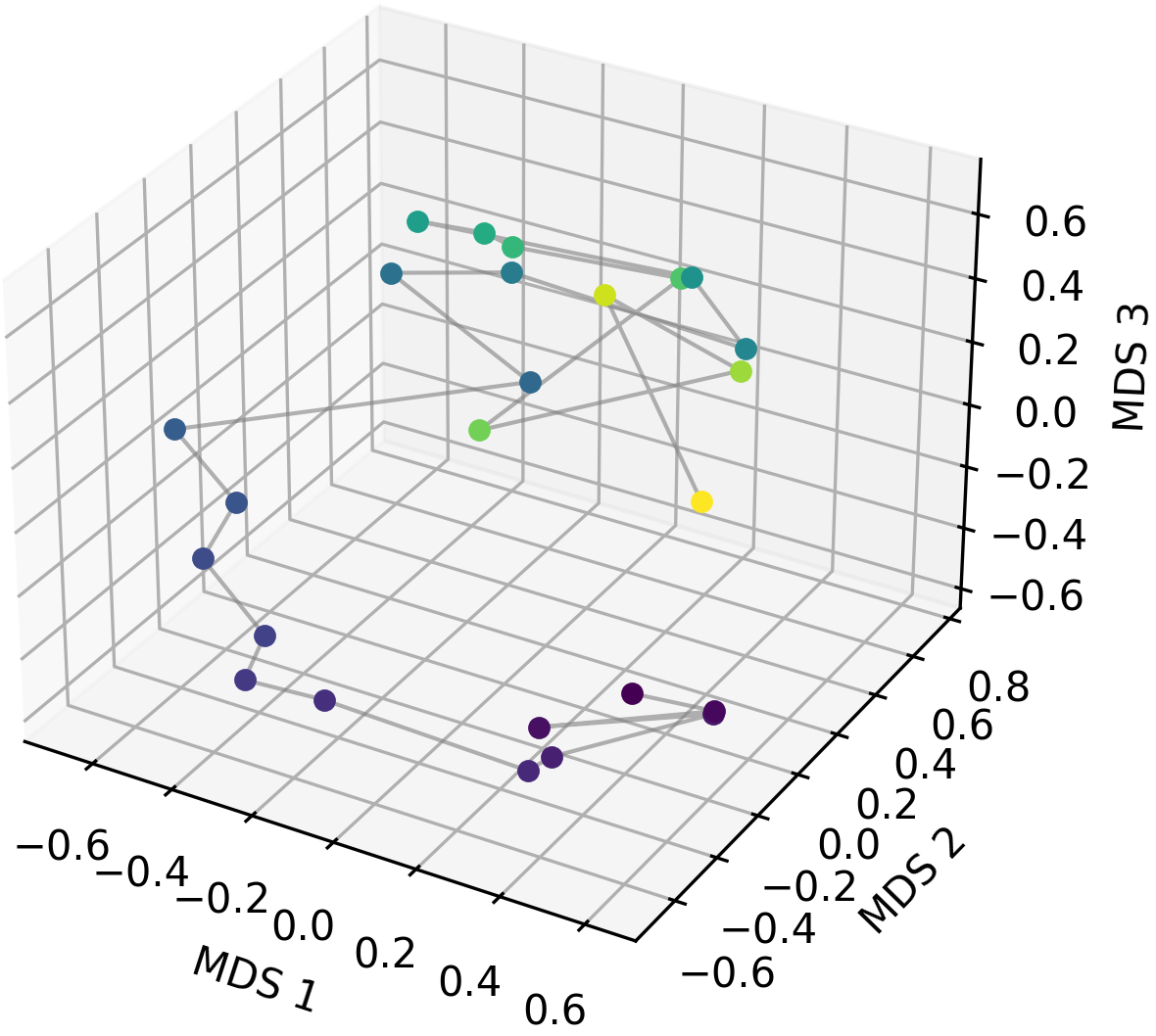} &
        \includegraphics[width=0.30\textwidth,height=0.30\textwidth]{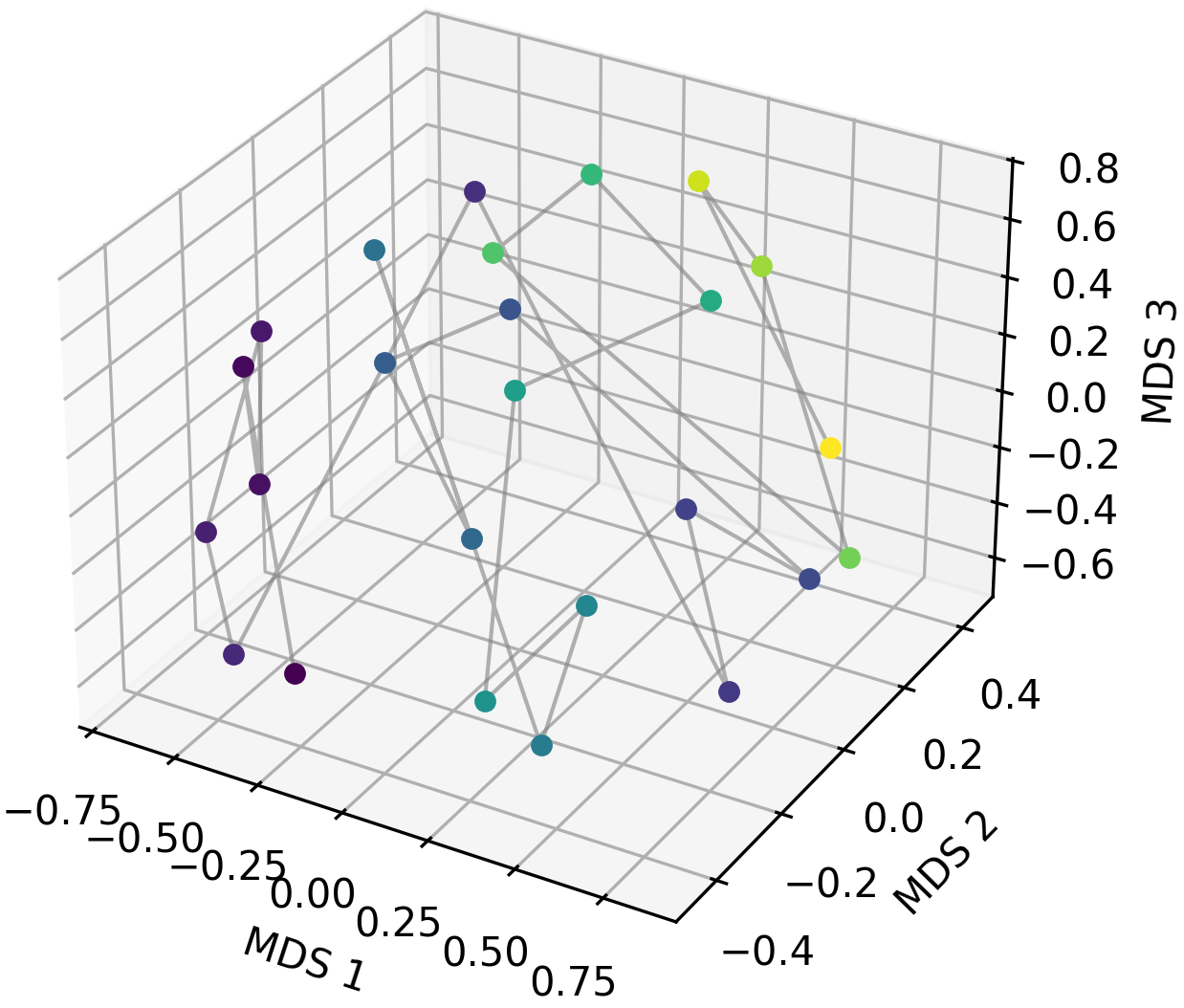}
    \end{tabular}

    \vspace{0.3em}
    \captionof{figure}{Layer-wise emergence of 3D Pitch geometry in Gemma-7B across model depth. Panels show (top-left) human perceptual geometry, (top-right) early-layer representation, (bottom-left) peak-alignment layer, and (bottom-right) final-layer MDS representation.}
    \label{fig:appendix_pitch_maps_3d_gemma}
\end{center}

\begin{center}
    \setlength{\tabcolsep}{2pt}
    \begin{tabular}{cc}
        \includegraphics[width=0.35\textwidth,height=0.25\textwidth]{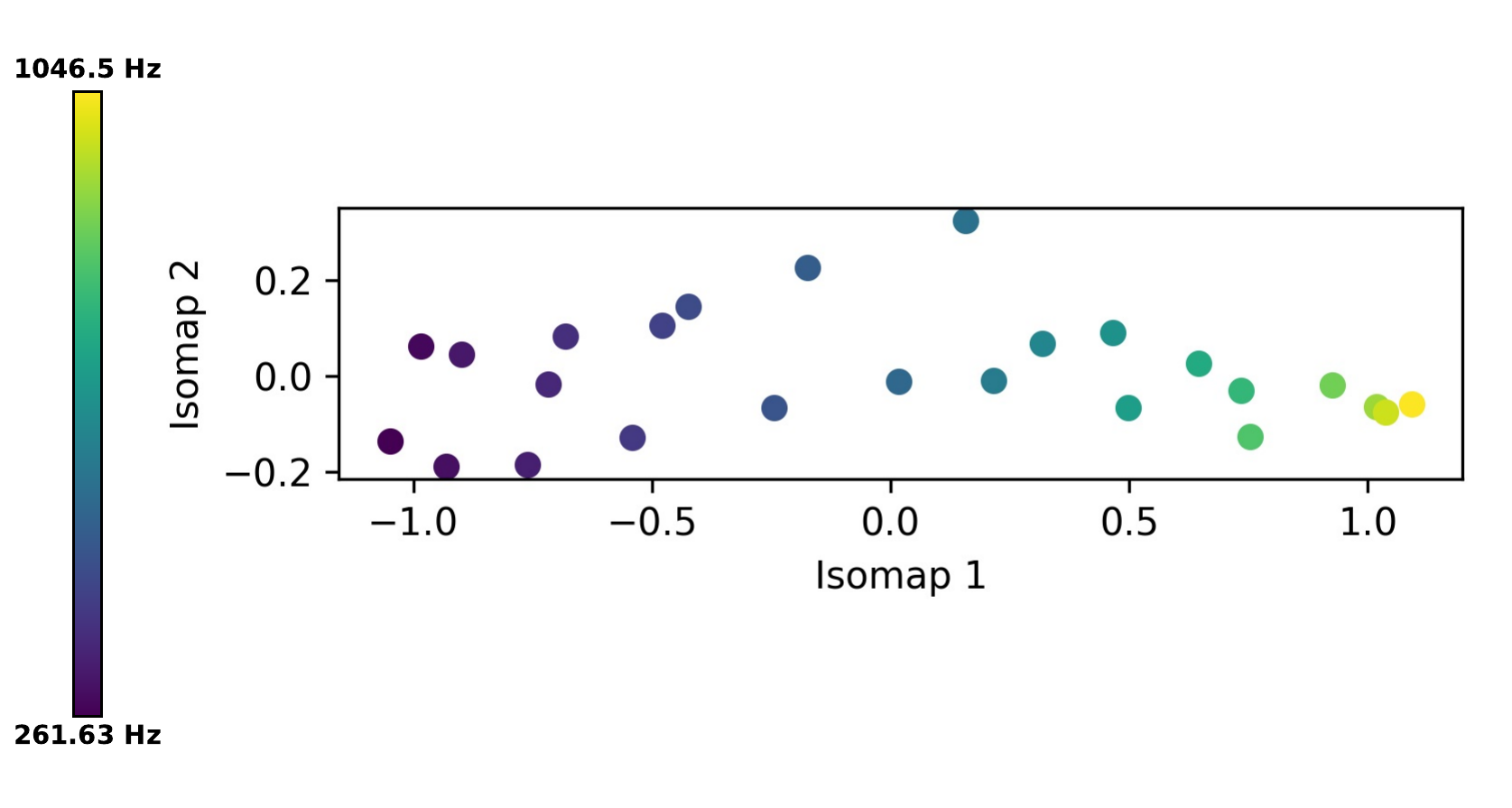} &
        \includegraphics[width=0.35\textwidth,height=0.25\textwidth]{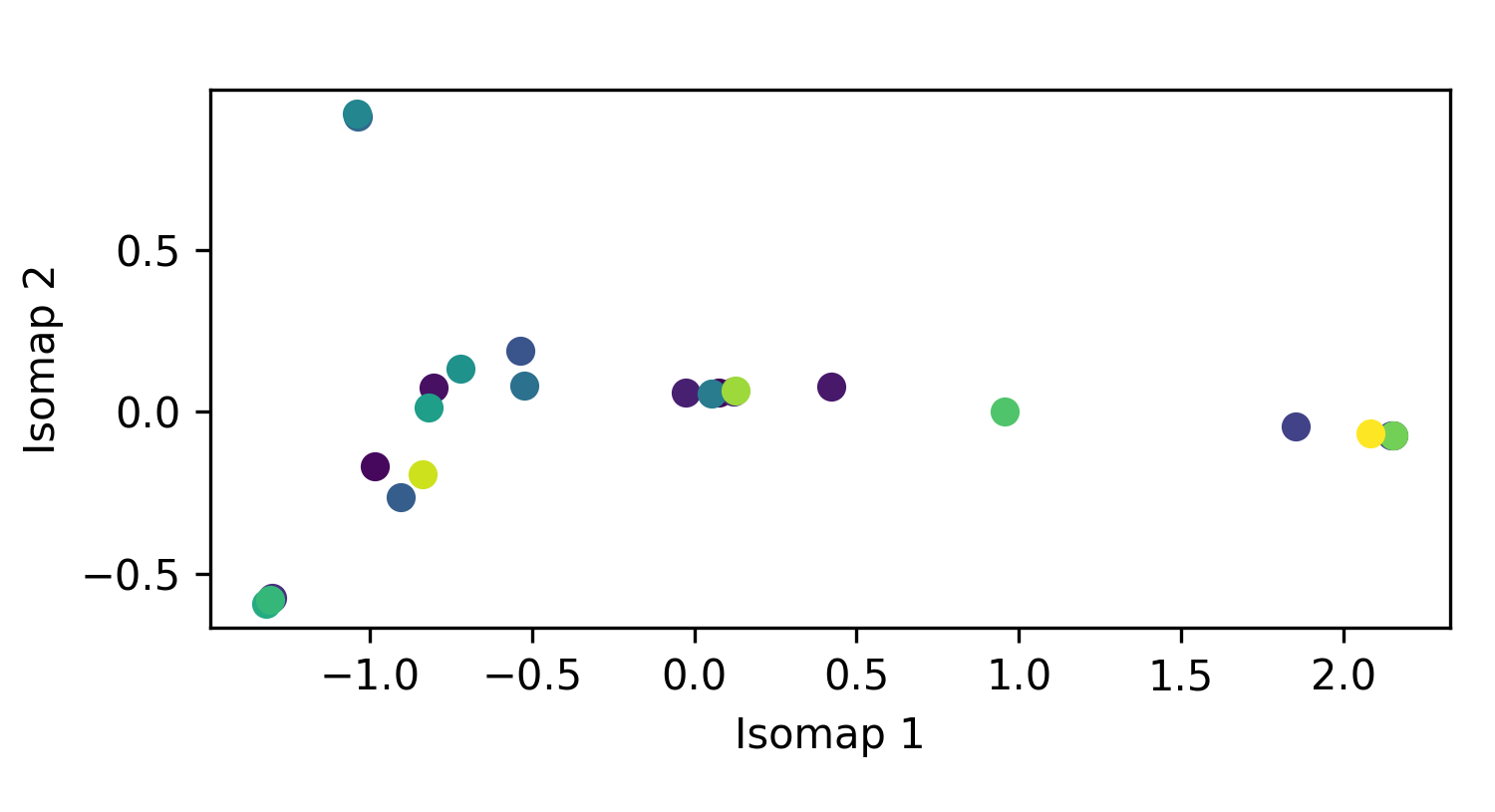} \\

        \includegraphics[width=0.35\textwidth,height=0.25\textwidth]{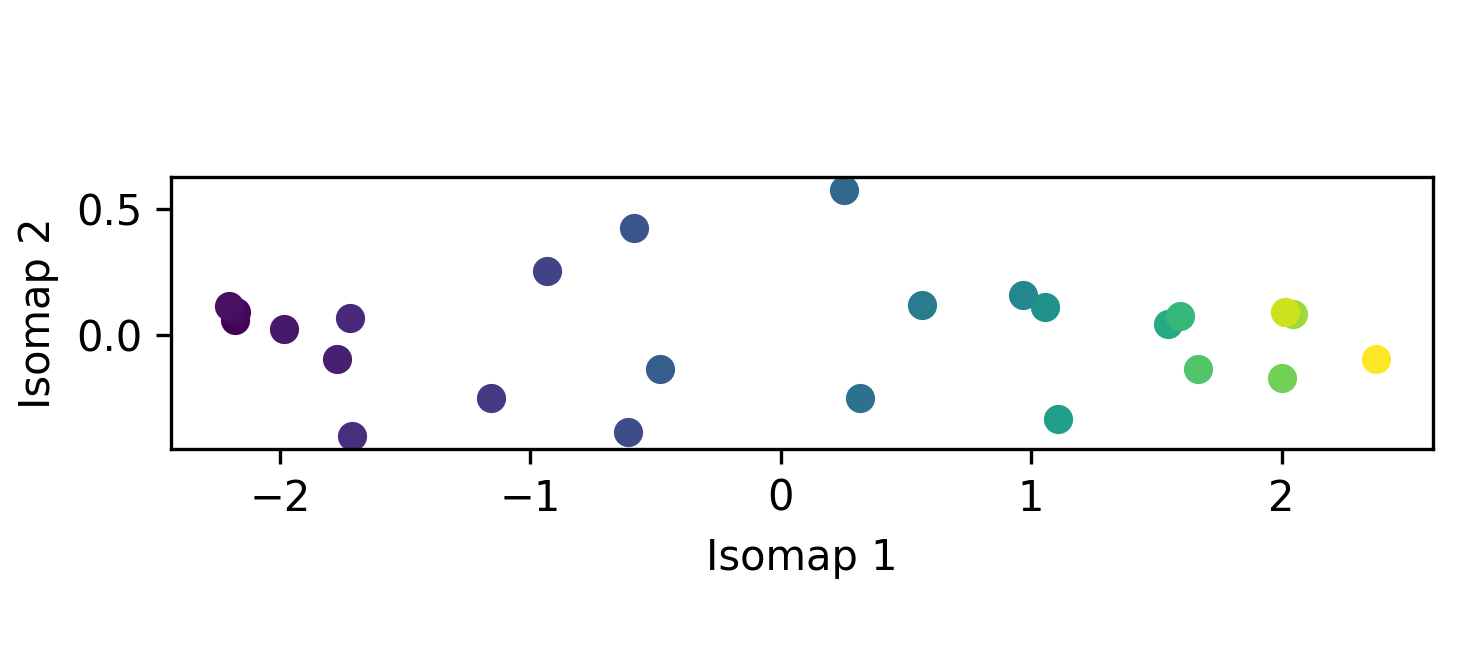} &
        \includegraphics[width=0.35\textwidth,height=0.25\textwidth]{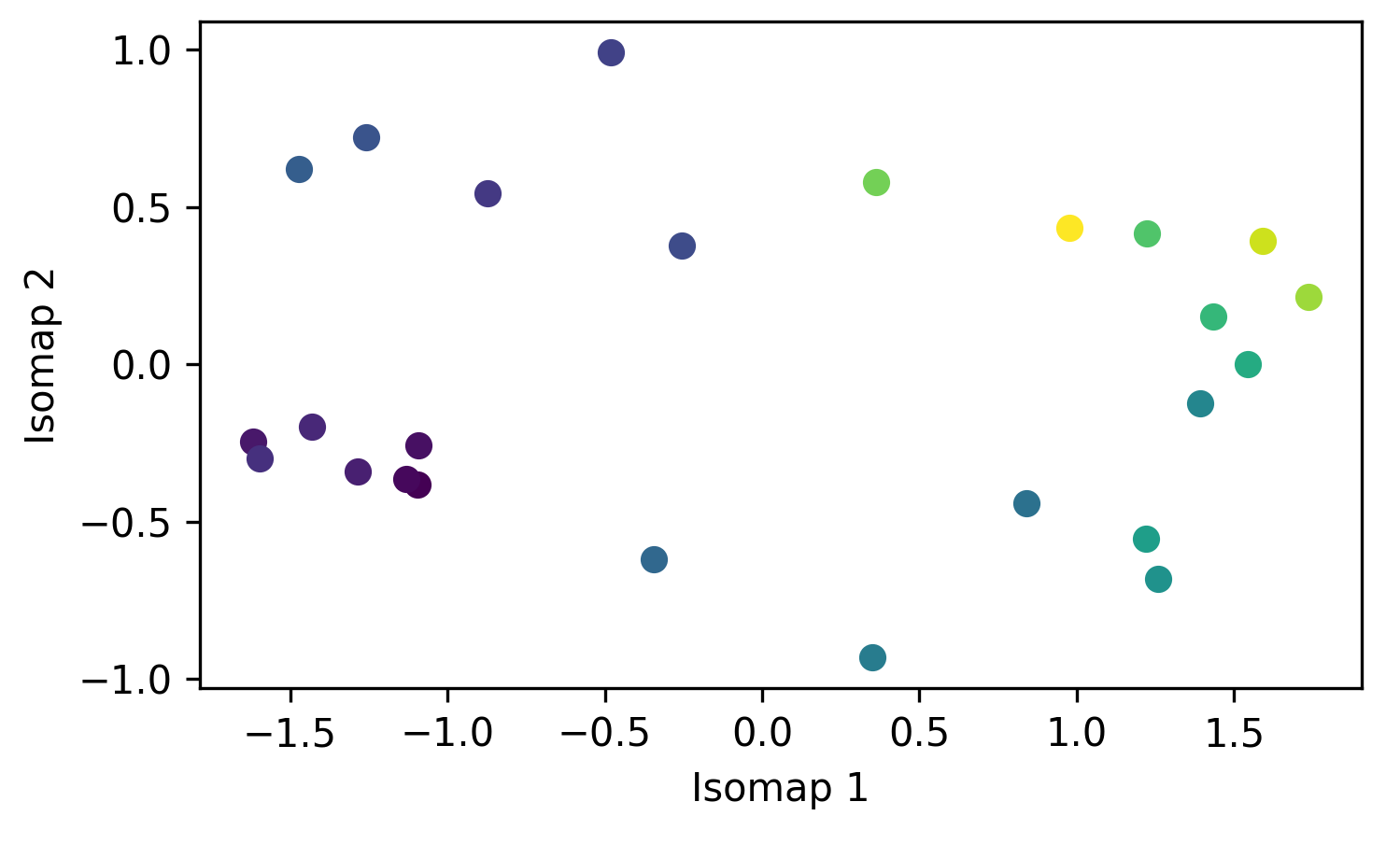}
    \end{tabular}

    \vspace{0.3em}
    \captionof{figure}{Layer-wise emergence of 2D Pitch geometry in Qwen-3-4B across model depth. Panels show (top-left) human perceptual geometry, (top-right) early-layer representation, (bottom-left) peak-alignment layer, and (bottom-right) final-layer isomap representation.}
    \label{fig:appendix_pitch_maps_2d_iso}
\end{center}

\begin{center}
    \setlength{\tabcolsep}{2pt}
    \begin{tabular}{cc}
        \includegraphics[width=0.35\textwidth,height=0.30\textwidth]{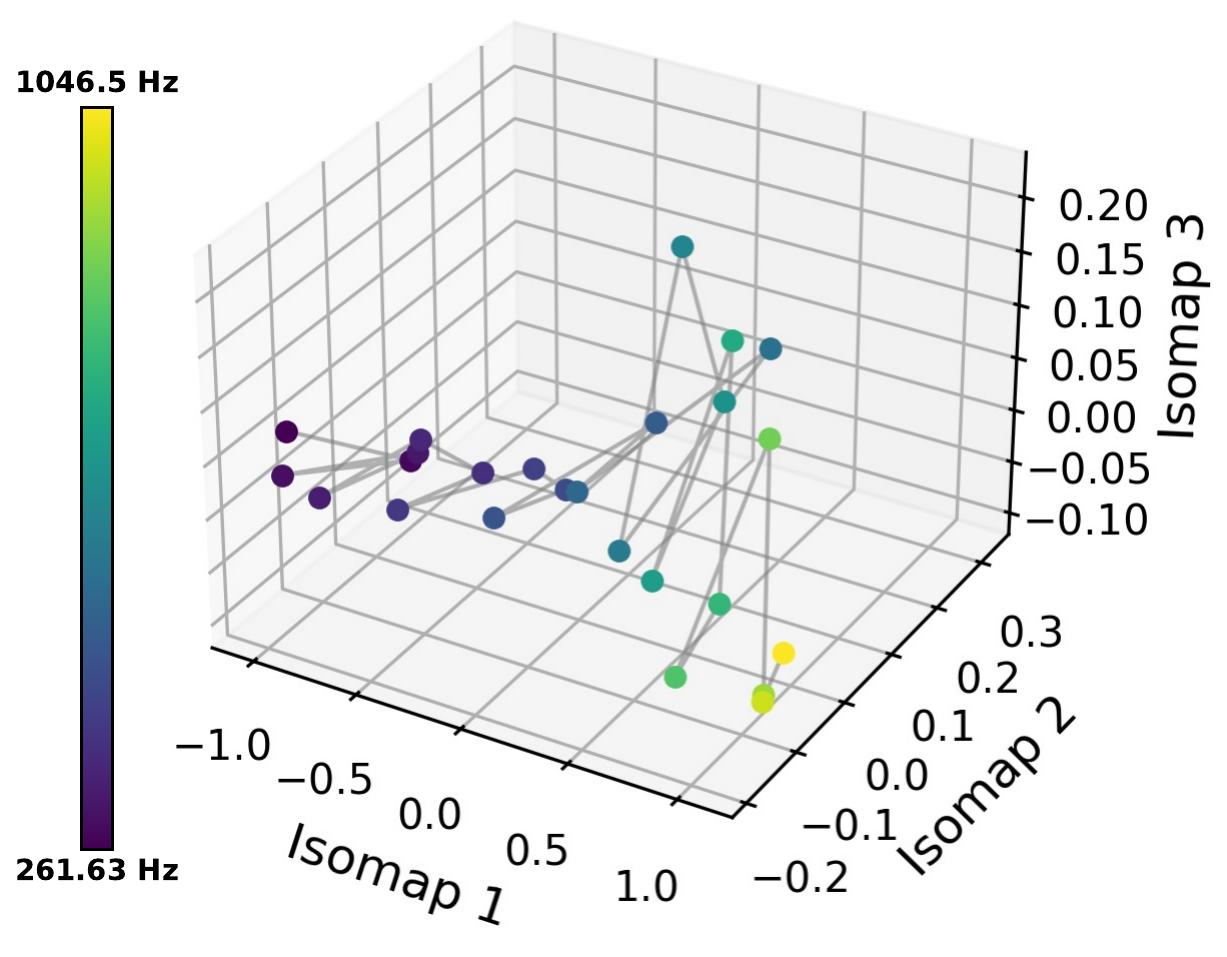} &
        \includegraphics[width=0.30\textwidth,height=0.30\textwidth]{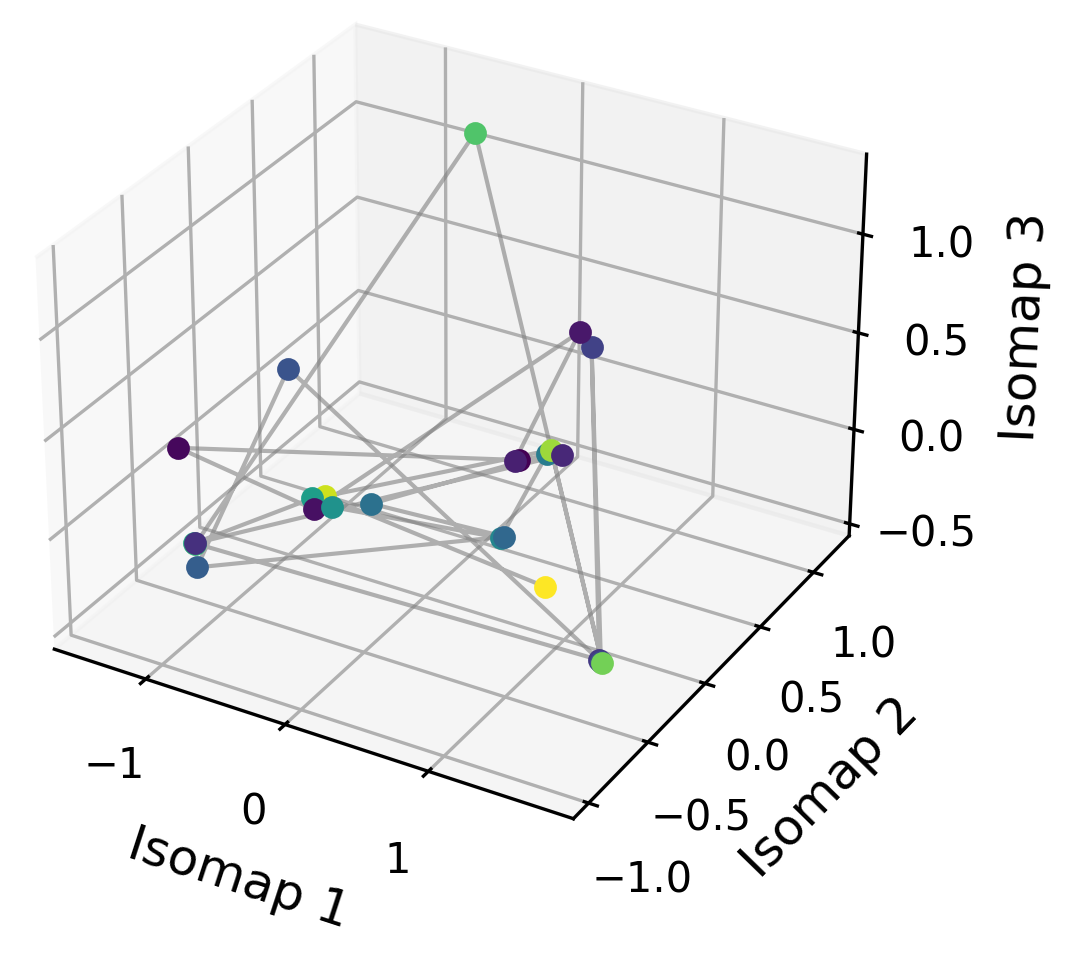} \\

        \includegraphics[width=0.30\textwidth,height=0.30\textwidth]{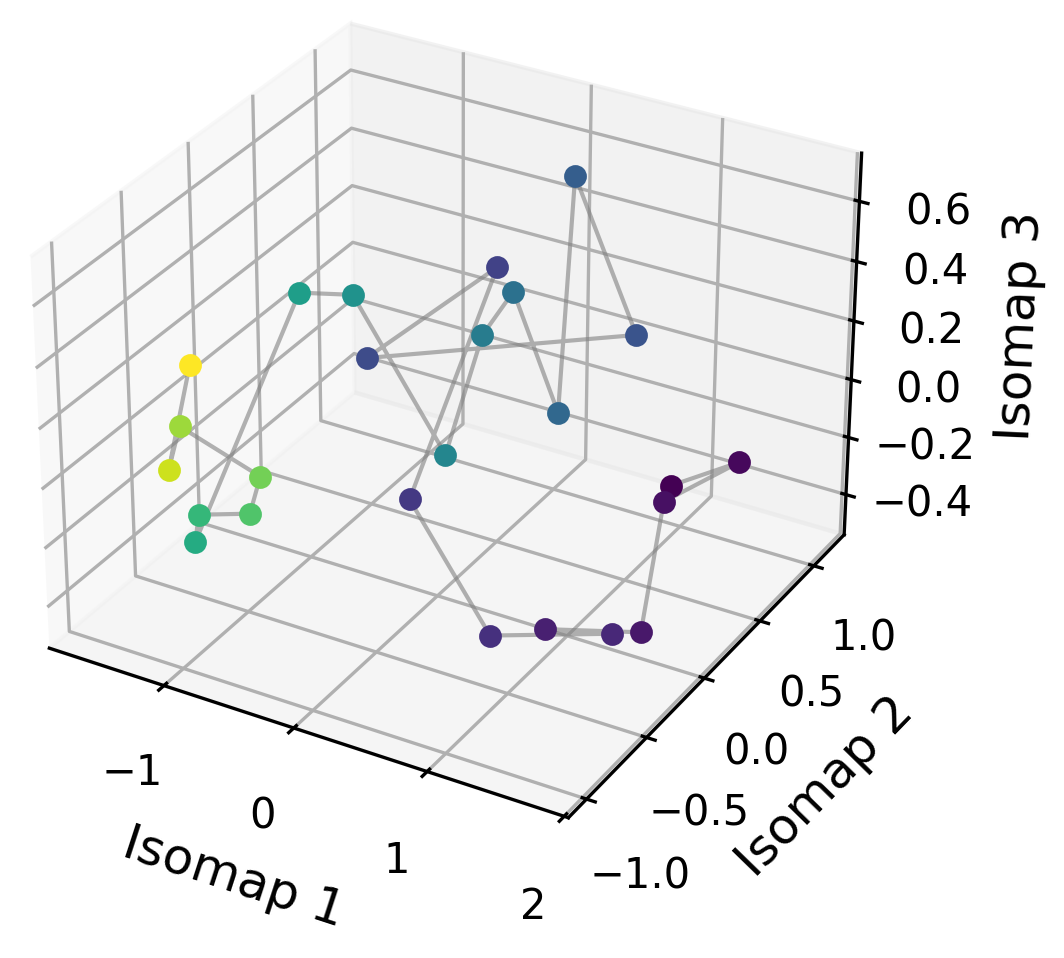} &
        \includegraphics[width=0.30\textwidth,height=0.30\textwidth]{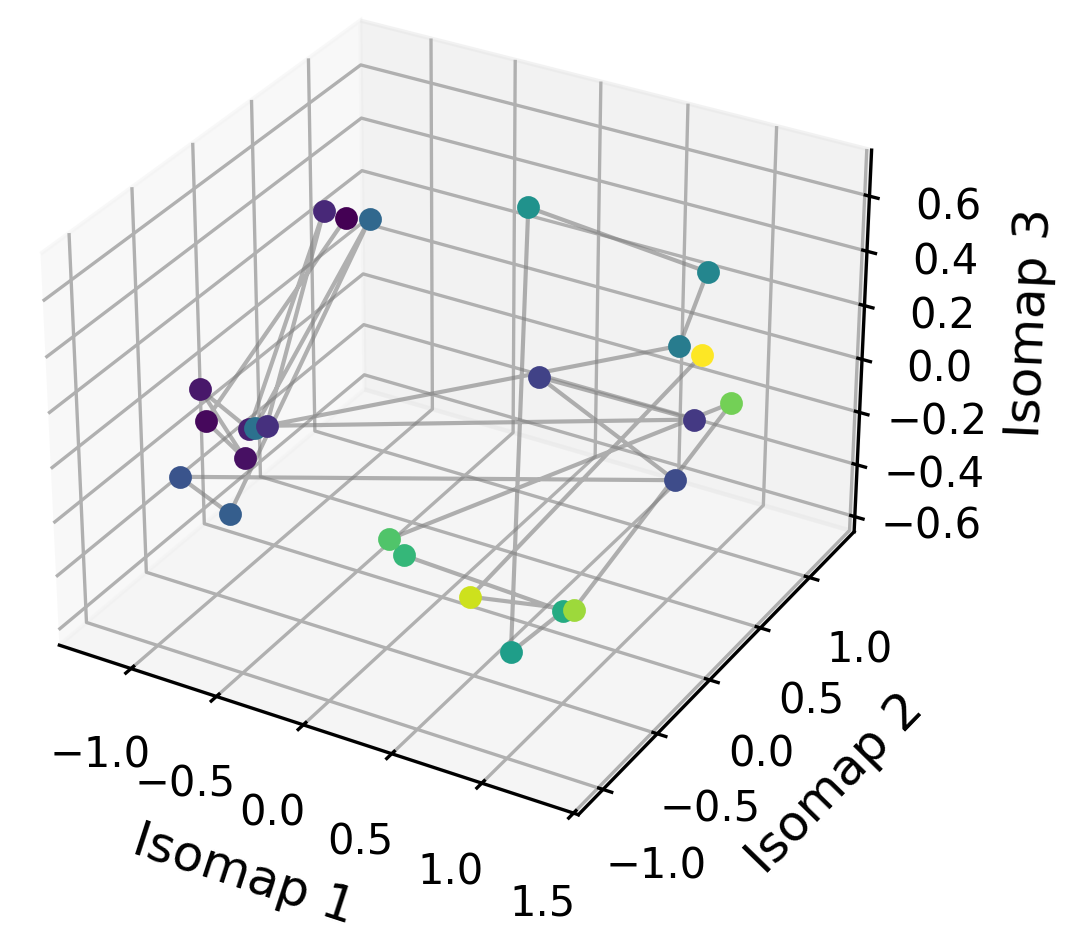}
    \end{tabular}

    \vspace{0.3em}
    \captionof{figure}{Layer-wise emergence of 3D Pitch geometry in Qwen-3-4B across model depth. Panels show (top-left) human perceptual geometry, (top-right) early-layer representation, (bottom-left) peak-alignment layer, and (bottom-right) final-layer isomap representation.}
    \label{fig:appendix_pitch_maps_3d_iso}
\end{center}

\subsection{Layer-wise Metric Profiles}

To complement the qualitative geometric visualizations, we report additional layer-wise alignment profiles for each stimulus domain.

\begin{center}
    \setlength{\tabcolsep}{4pt}
    \begin{tabular}{cc}
        \includegraphics[width=0.47\textwidth]{Figures/Color/rgpa-c-l8b.png} &
        \includegraphics[width=0.47\textwidth]{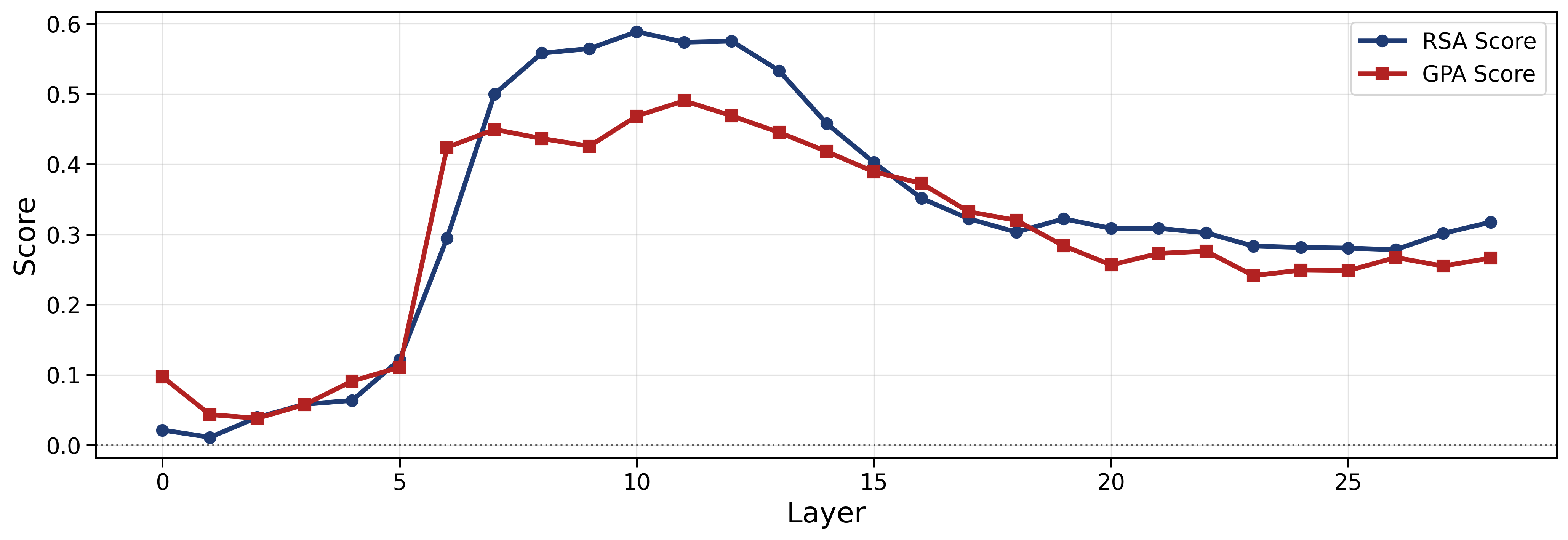} \\

        \includegraphics[width=0.47\textwidth]{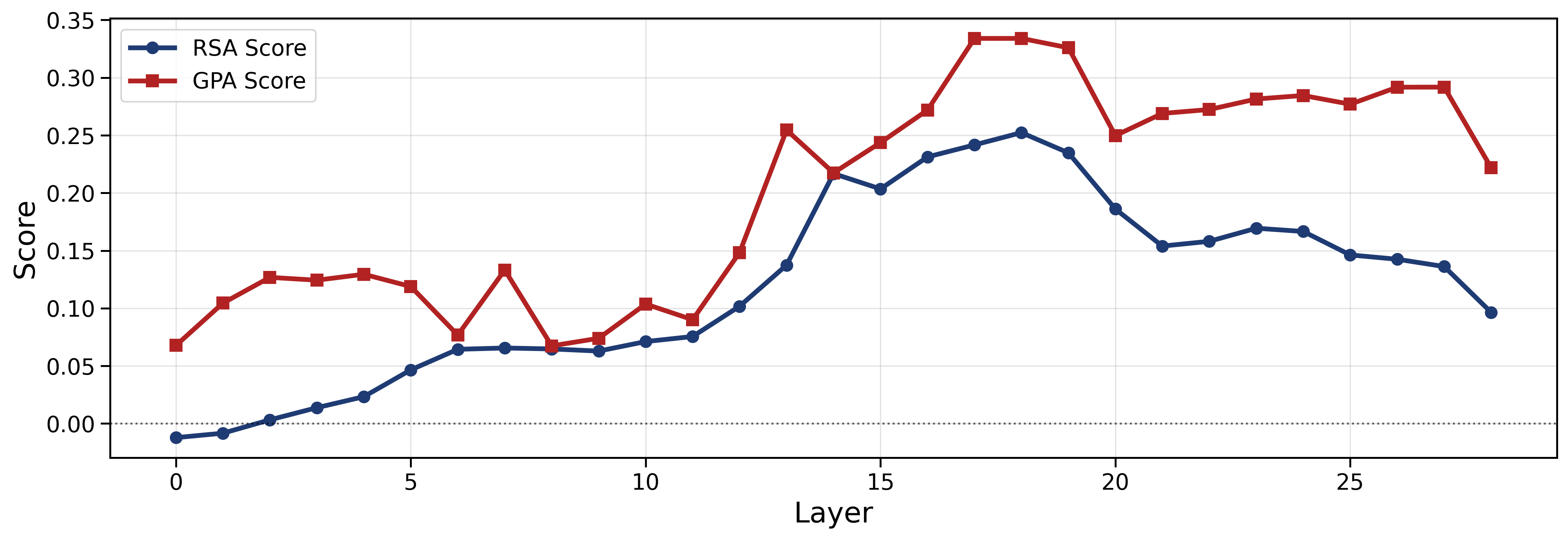} &
        \includegraphics[width=0.47\textwidth]{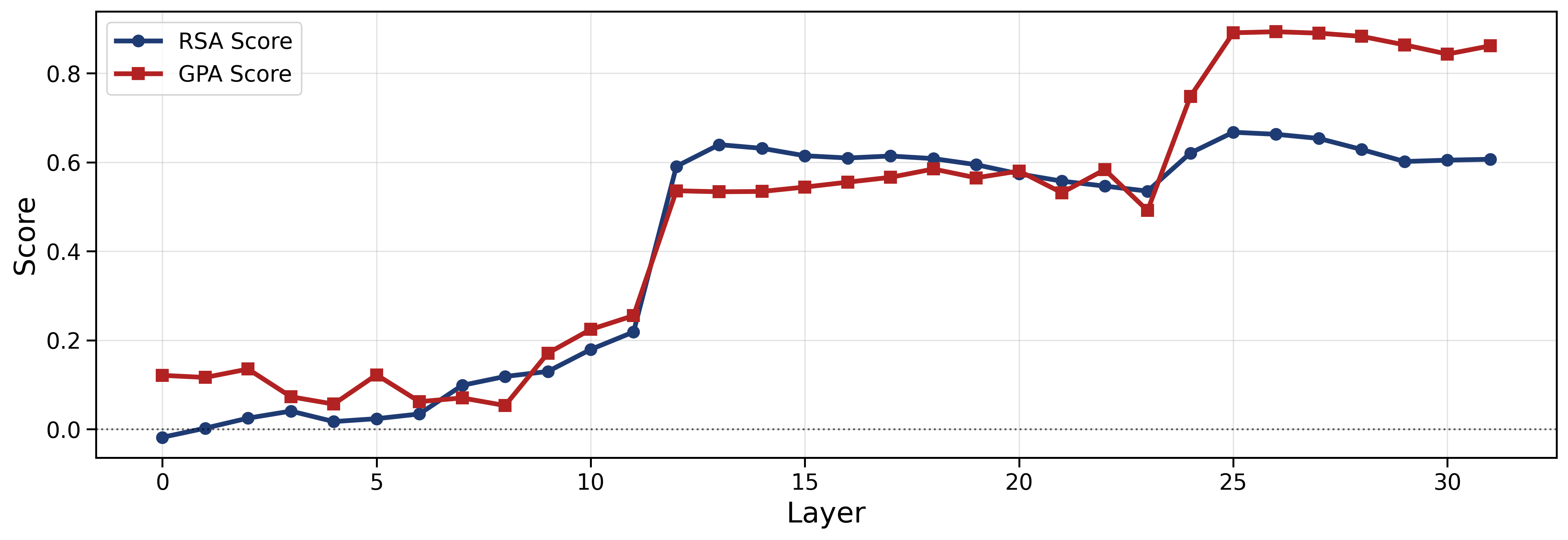}\\
    \end{tabular}

    \vspace{0.3em}
    \captionof{figure}{Layer-wise metric profiles for color across four models: (Llama-3-8B) model 1, (Llama-3.2-3B) model 2, (Gemma-7B) model 3, and (Qwen-3-4B) model 4.}
    \label{fig:appendix_color_metrics}
\end{center}

\begin{center}
    \setlength{\tabcolsep}{4pt}
    \begin{tabular}{cc}
        \includegraphics[width=0.47\textwidth]{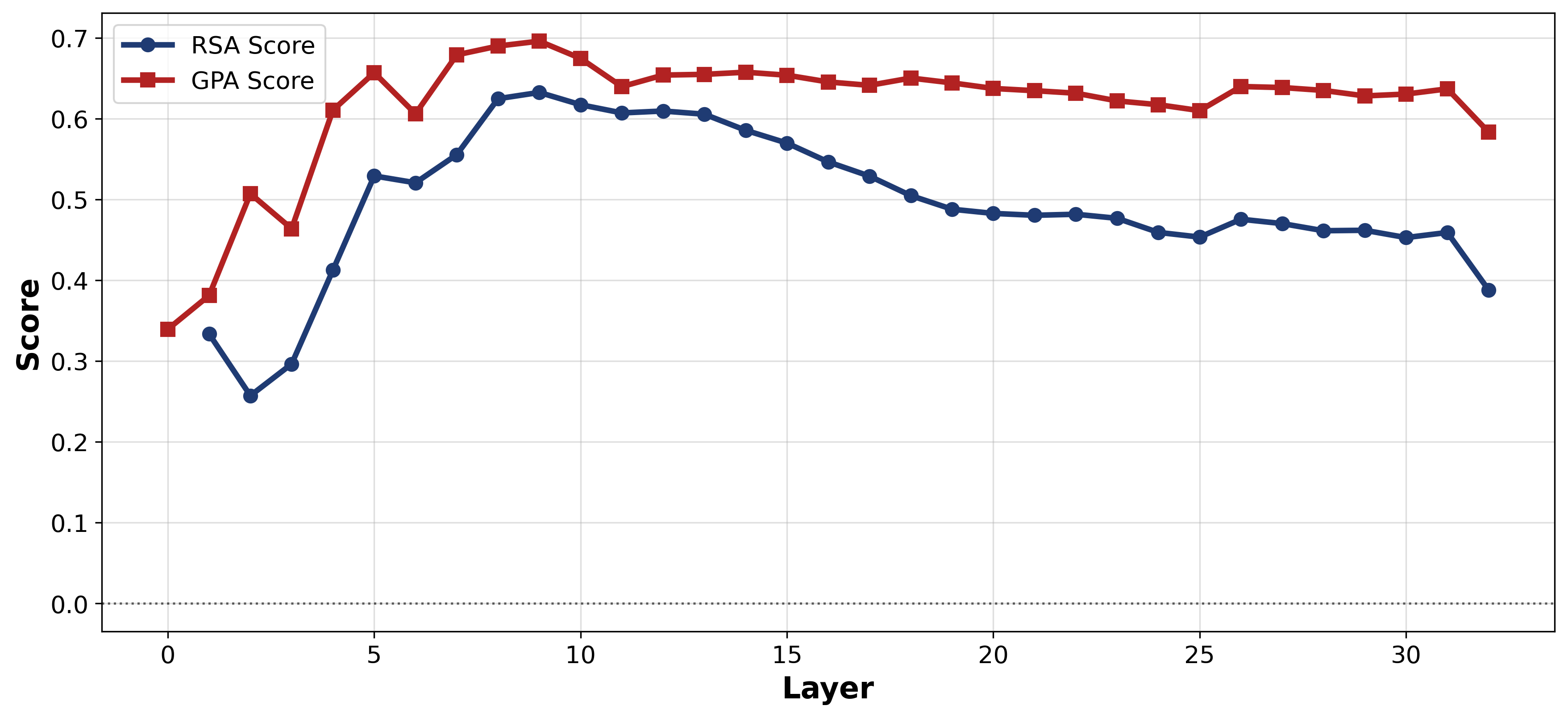} &
        \includegraphics[width=0.47\textwidth]{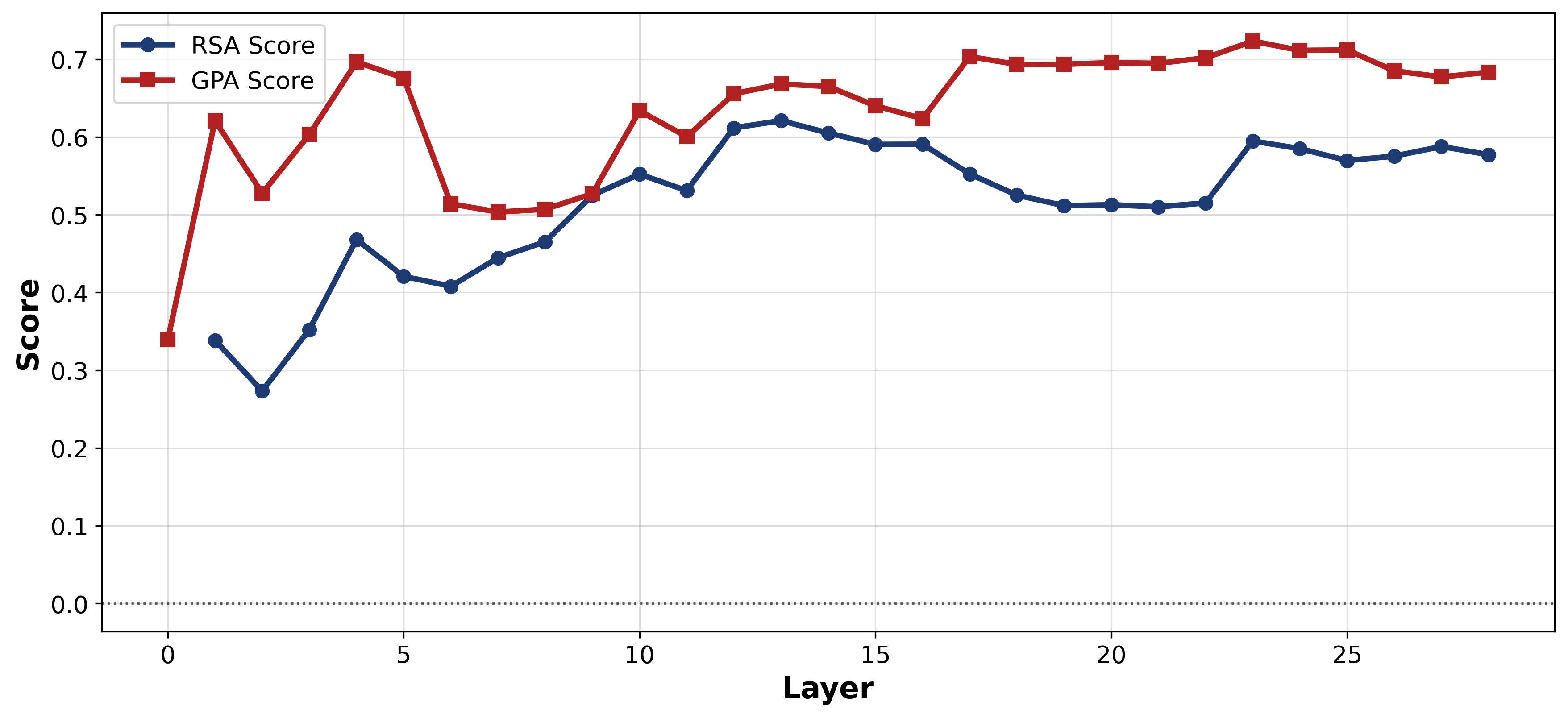} \\

        \includegraphics[width=0.47\textwidth]{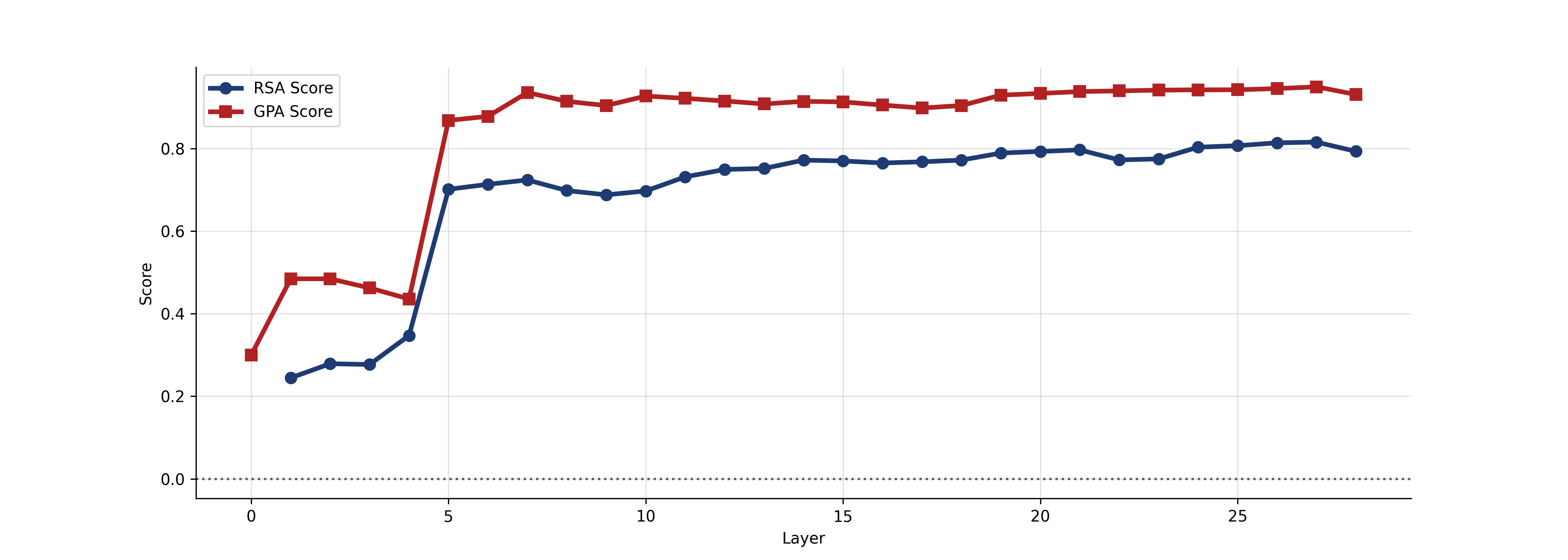} &
        \includegraphics[width=0.47\textwidth]{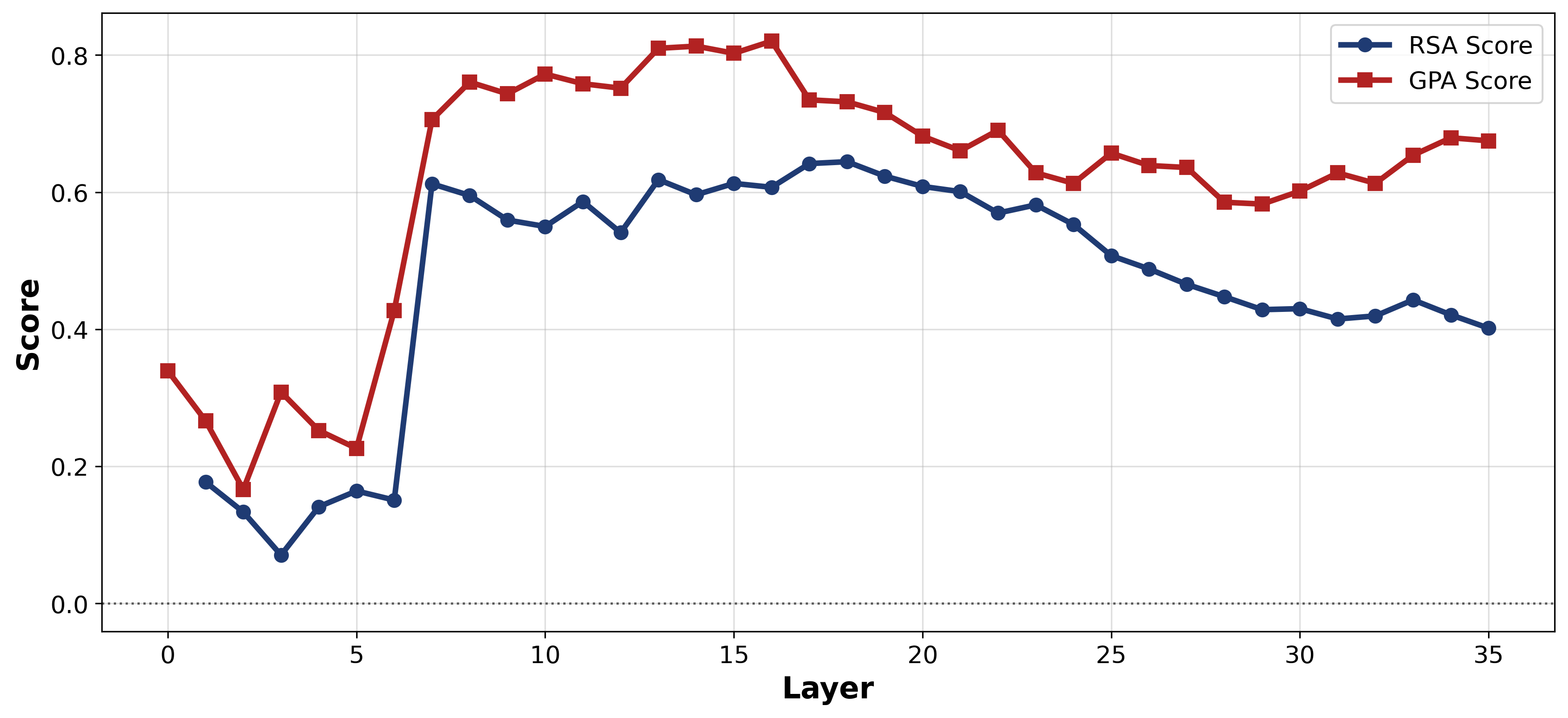}\\
    \end{tabular}

    \vspace{0.3em}
    \captionof{figure}{Layer-wise metric profiles for Emotion across four models: (Llama-3-8B) model 1, (Llama-3.2-3B) model 2, (Gemma-7B) model 3, and (Qwen-3-4B) model 4.}
    \label{fig:appendix_emotion_metrics}
\end{center}

\begin{center}
    \setlength{\tabcolsep}{4pt}
    \begin{tabular}{cc}
        \includegraphics[width=0.47\textwidth]{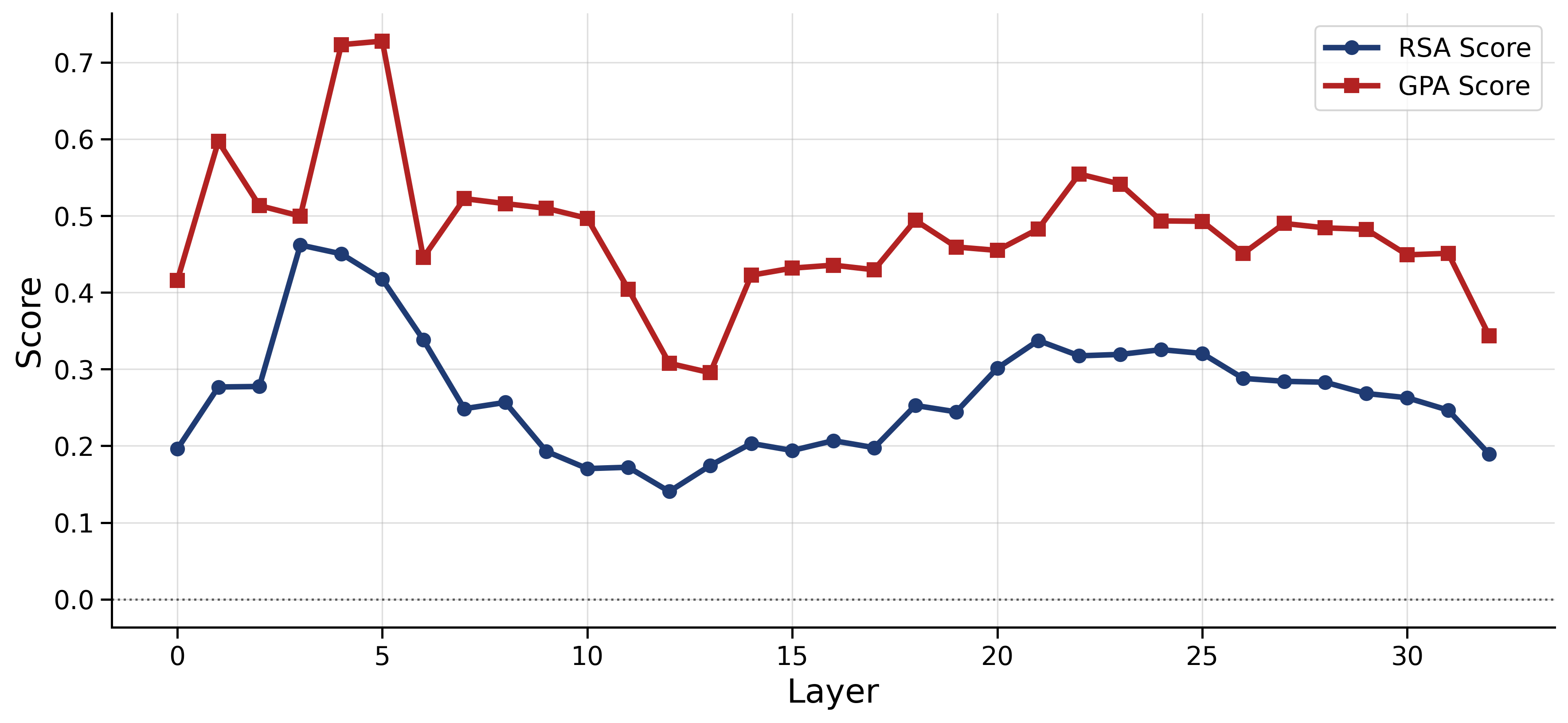} &
        \includegraphics[width=0.47\textwidth]{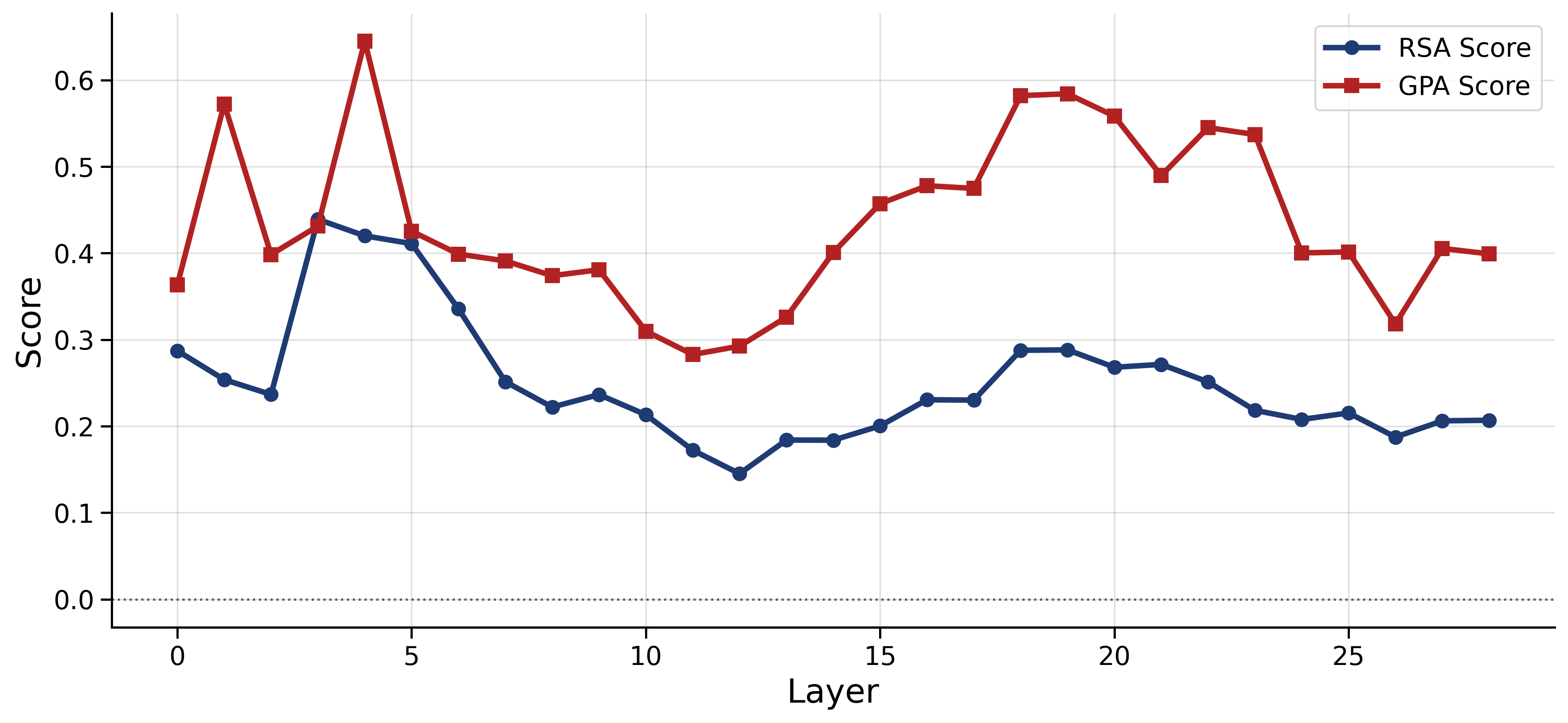} \\

        \includegraphics[width=0.47\textwidth]{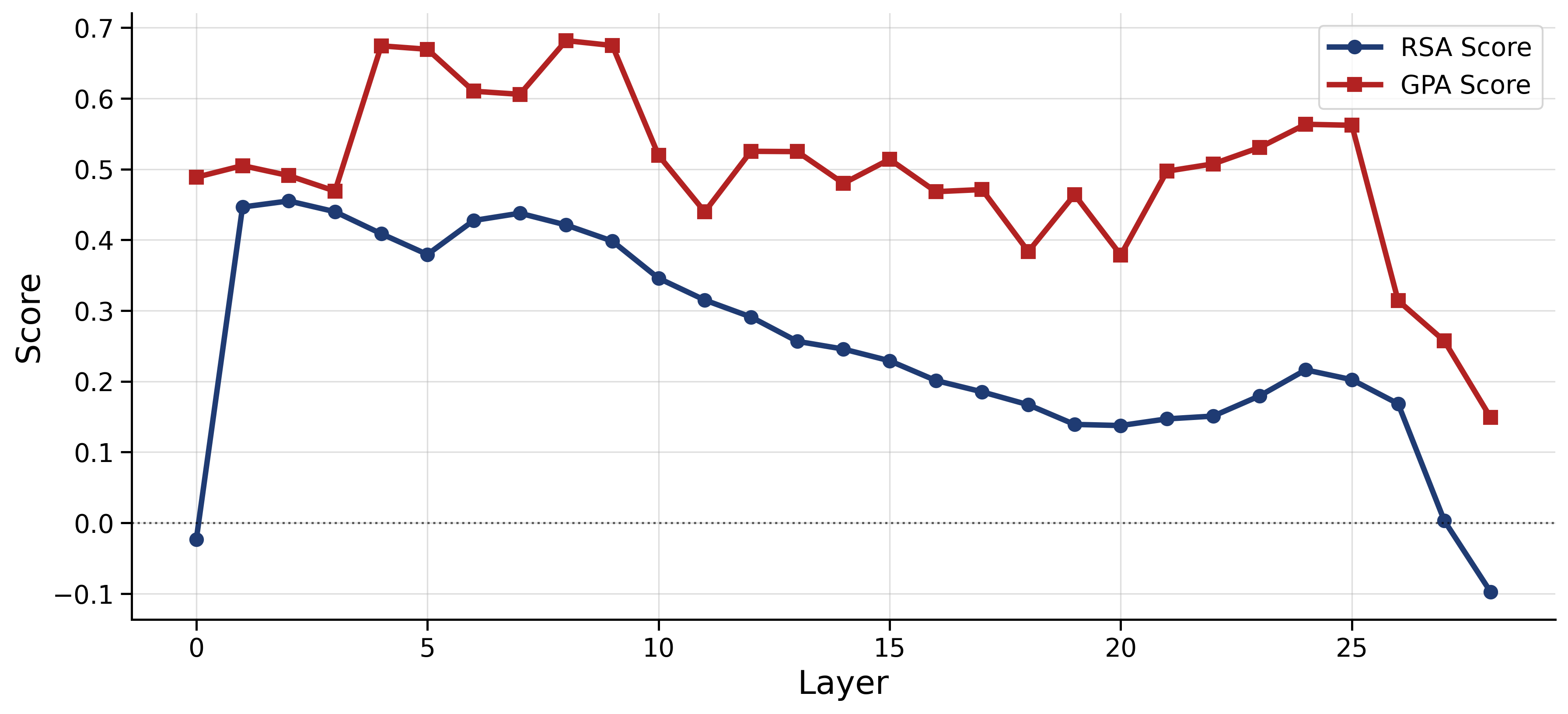} &
        \includegraphics[width=0.47\textwidth]{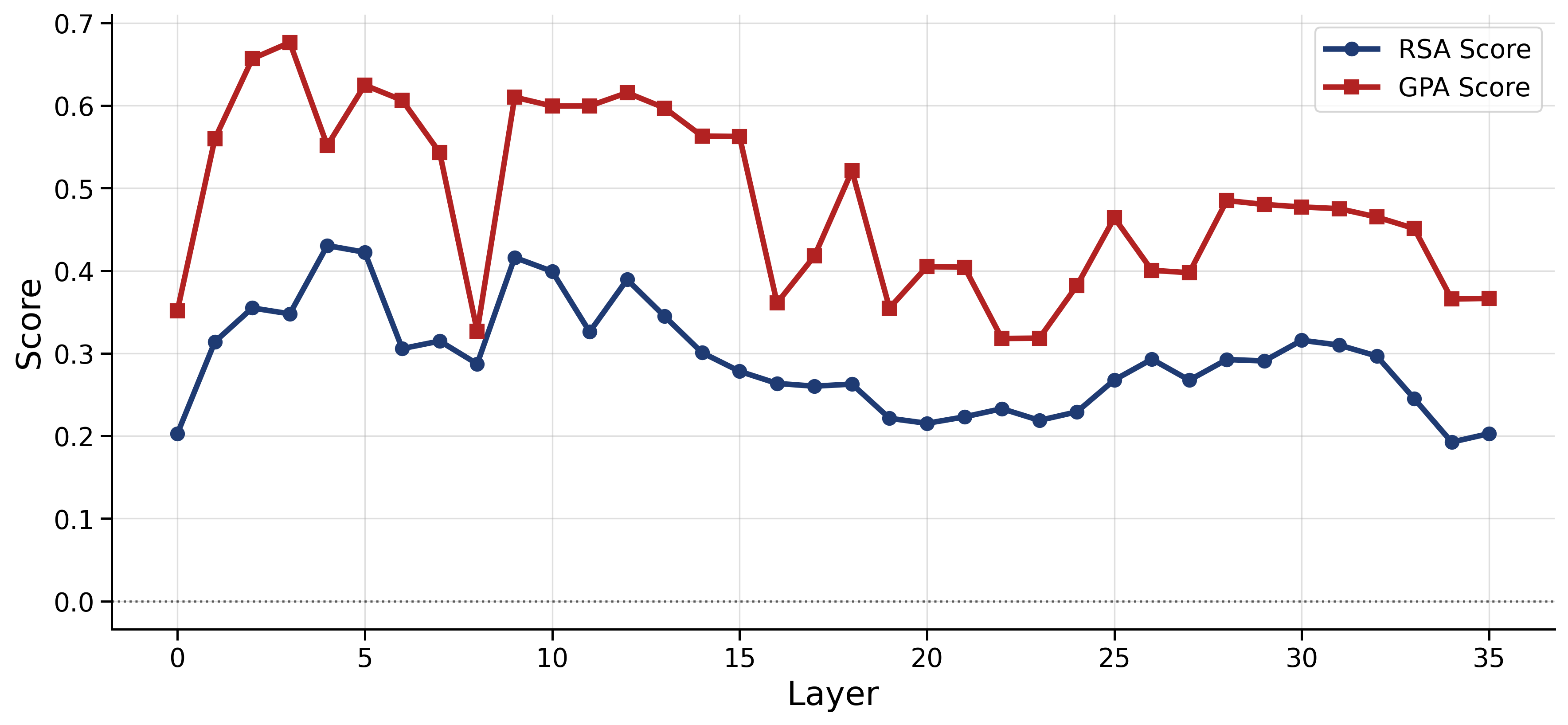}\\
    \end{tabular}

    \vspace{0.3em}
    \captionof{figure}{Layer-wise metric profiles for Taste across four models: (Llama-3-8B) model 1, (Llama-3.2-3B) model 2, (Gemma-7B) model 3, and (Qwen-3-4B) model 4.}
    \label{fig:appendix_taste_metrics}
\end{center}

\begin{center}
    \setlength{\tabcolsep}{4pt}
    \begin{tabular}{cc}
        \includegraphics[width=0.47\textwidth]{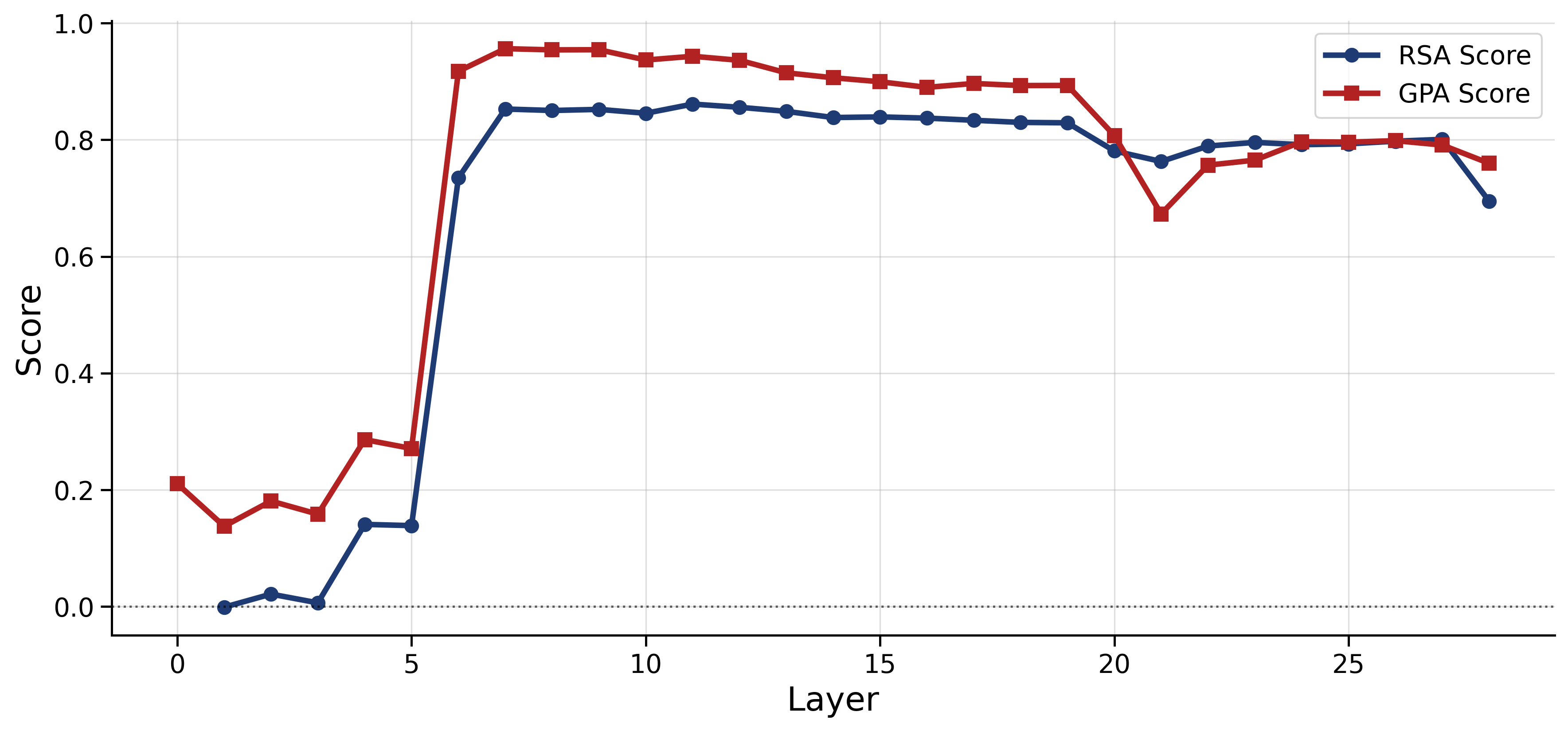} &
        \includegraphics[width=0.47\textwidth]{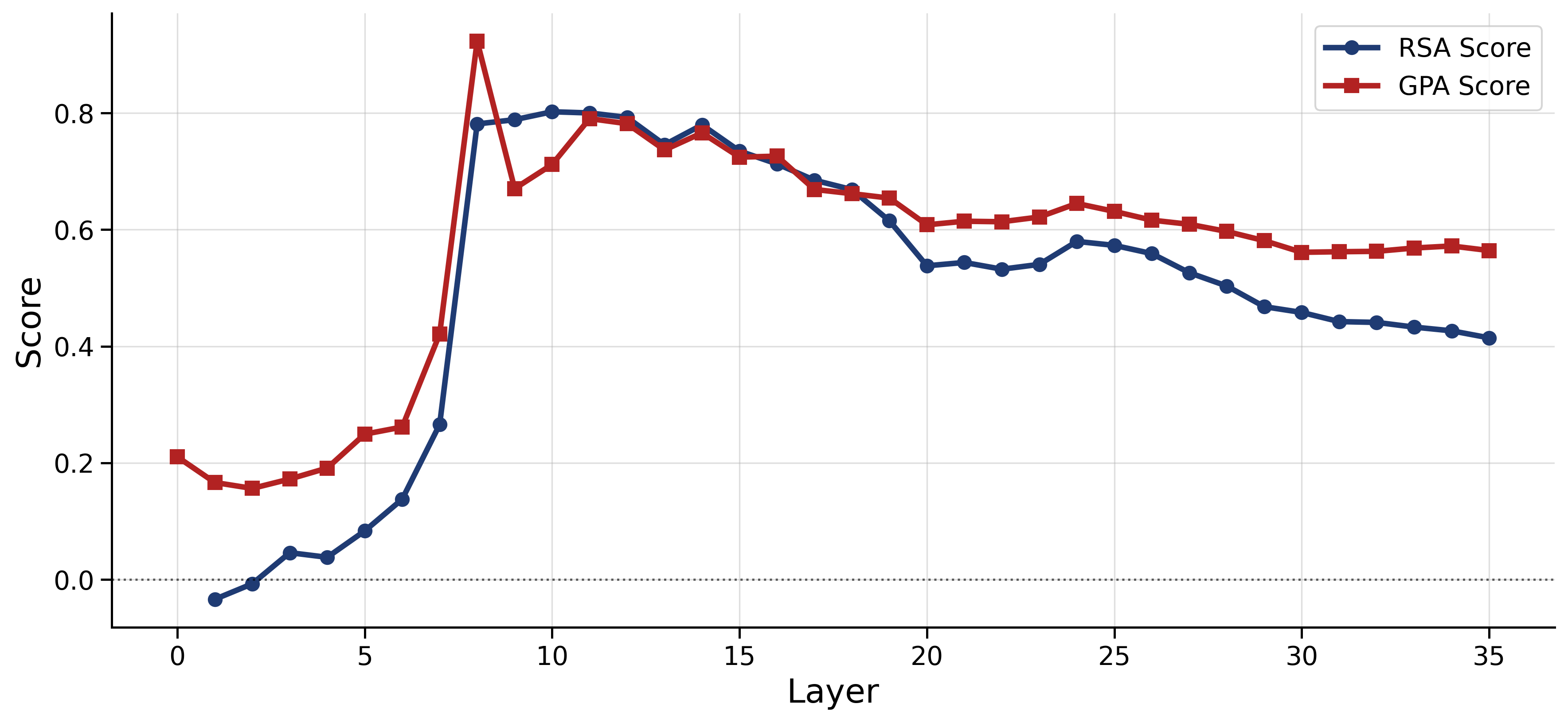} \\

        \includegraphics[width=0.47\textwidth]{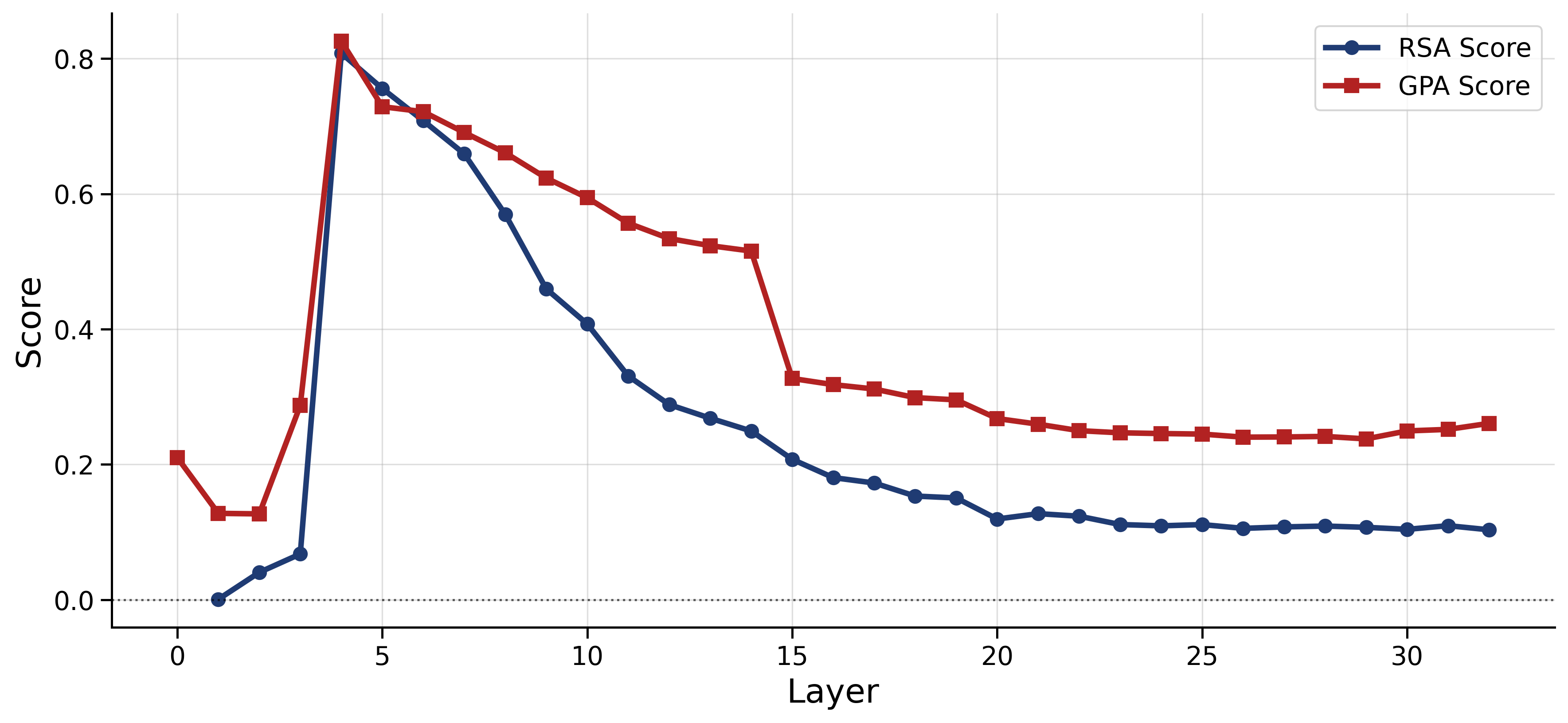} &
        \includegraphics[width=0.47\textwidth]{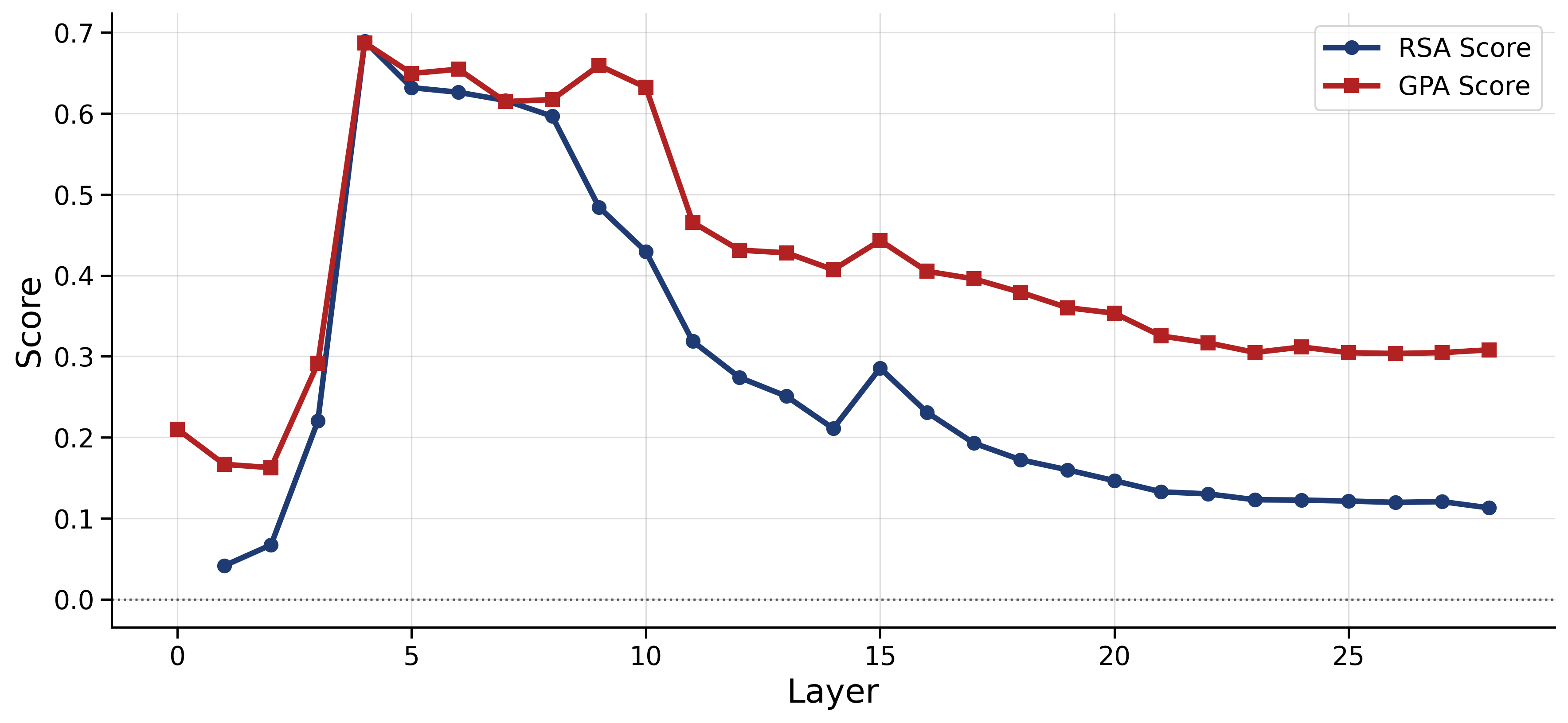}\\
    \end{tabular}

    \vspace{0.3em}
    \captionof{figure}{Layer-wise metric profiles for Pitch across four models: (Gemma-7B) model 1, (Qwen-3-4B) model 2, (Llama-3-8B) model 3, and (Llama-3.2-3B) model 4.}
    \label{fig:appendix_pitch_metrics}
\end{center}

\subsection{Bootstrap Confidence Intervals for Layer-wise Alignment}

To assess the statistical stability of the observed layer-wise alignment profiles, we estimate bootstrap confidence intervals for both RSA and GPA scores by resampling stimuli with replacement at each layer. For each perceptual domain and each layer, we perform 1000 bootstrap iterations. In every iteration, the stimulus set is resampled with replacement, the corresponding human and model dissimilarity submatrices are extracted, and RSA and GPA are recomputed on the resampled set. We report 95\% percentile bootstrap confidence intervals for both metrics, allowing us to evaluate whether the observed layer-wise trends are robust to the particular choice of stimuli rather than driven by a small subset of labels.

\begin{center}
    \setlength{\tabcolsep}{4pt}
    \begin{tabular}{cc}
        \includegraphics[width=0.45\textwidth,height=0.20\textwidth]{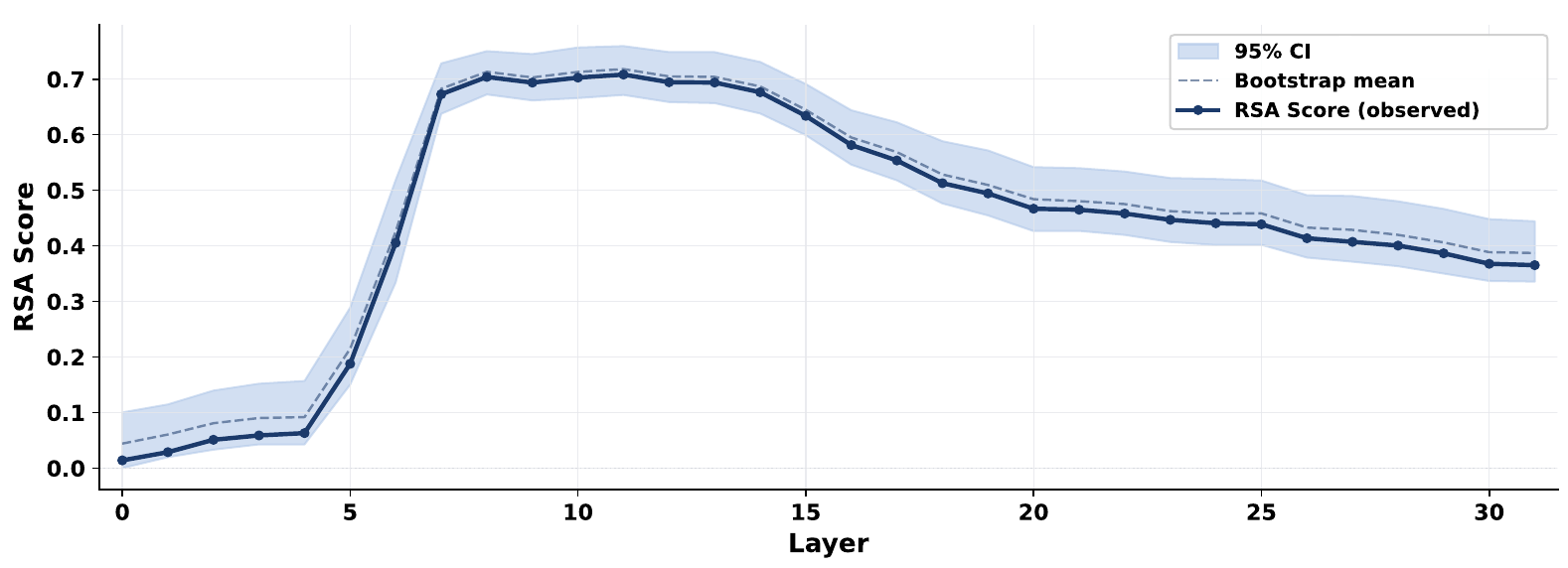} &
        \includegraphics[width=0.45\textwidth,height=0.20\textwidth]{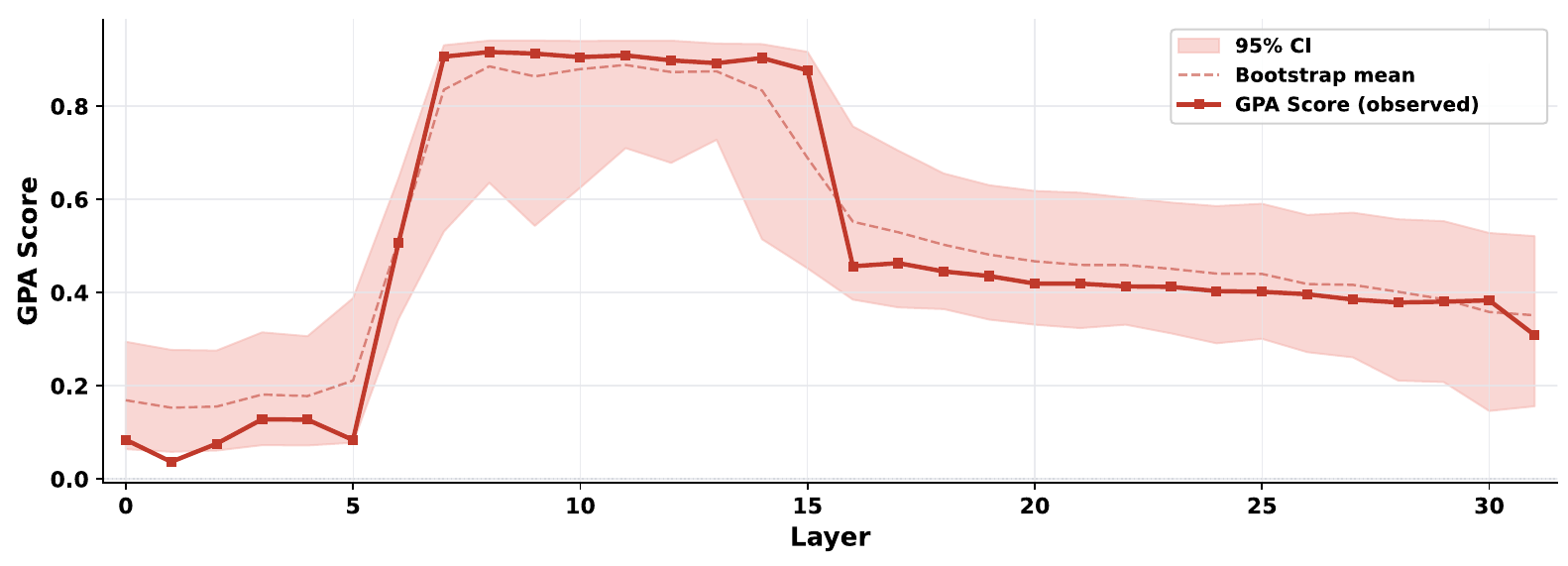}
    \end{tabular}

    \vspace{0.3em}
    \captionof{figure}{Bootstrap confidence intervals for color in LLaMA-3-8B, showing layer-wise RSA (left) and GPA (right) profiles with 95\% confidence bands.}
    \label{fig:appendix_color_bootstrap}
\end{center}

\begin{center}
    \setlength{\tabcolsep}{4pt}
    \begin{tabular}{cc}
        \includegraphics[width=0.45\textwidth,height=0.20\textwidth]{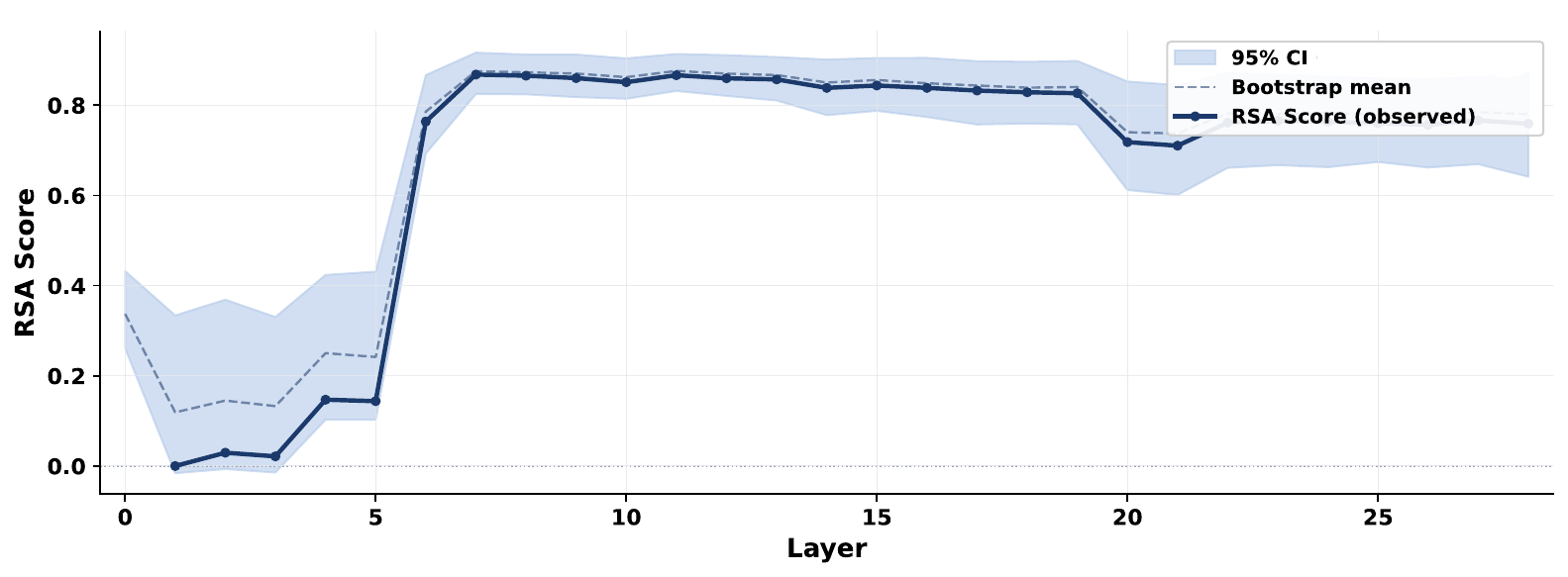} &
        \includegraphics[width=0.45\textwidth,height=0.20\textwidth]{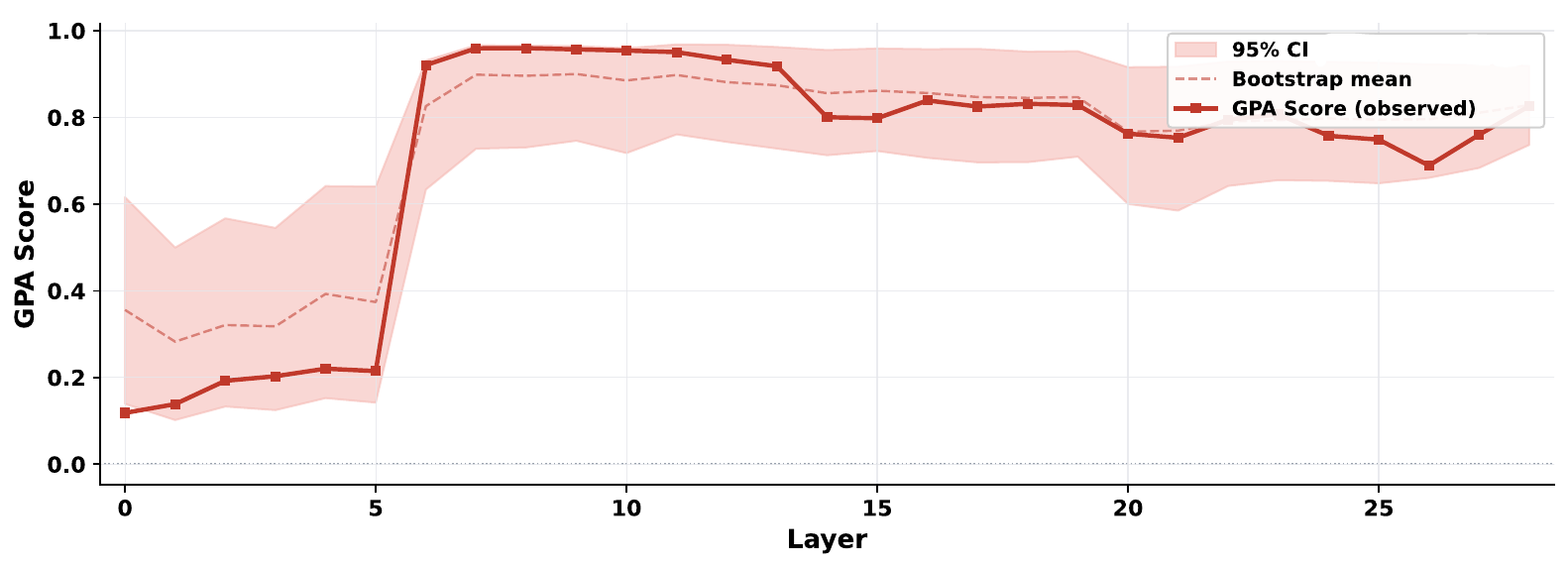}
    \end{tabular}

    \vspace{0.3em}
    \captionof{figure}{Bootstrap confidence intervals for Pitch in Qwen-3-4B, showing layer-wise RSA (left) and GPA (right) profiles with 95\% confidence bands.}
    \label{fig:appendix_pitch_bootstrap}
\end{center}

\begin{center}
    \setlength{\tabcolsep}{4pt}
    \begin{tabular}{cc}
        \includegraphics[width=0.45\textwidth,height=0.20\textwidth]{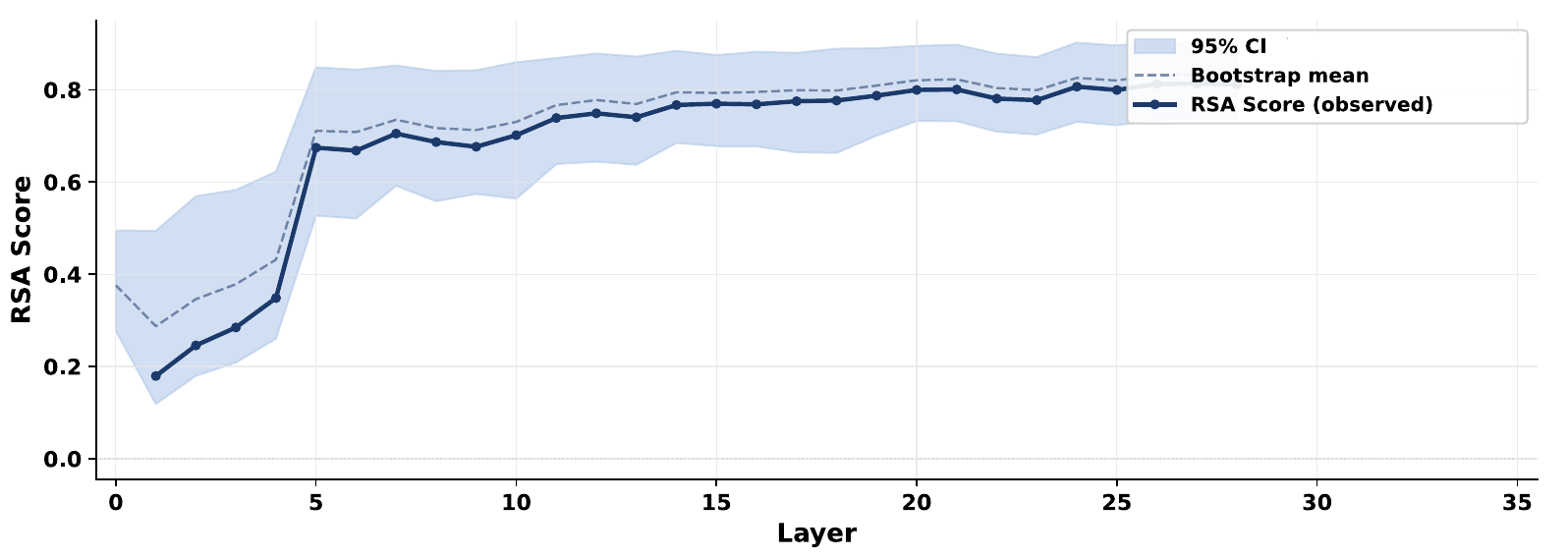} &
        \includegraphics[width=0.45\textwidth,height=0.20\textwidth]{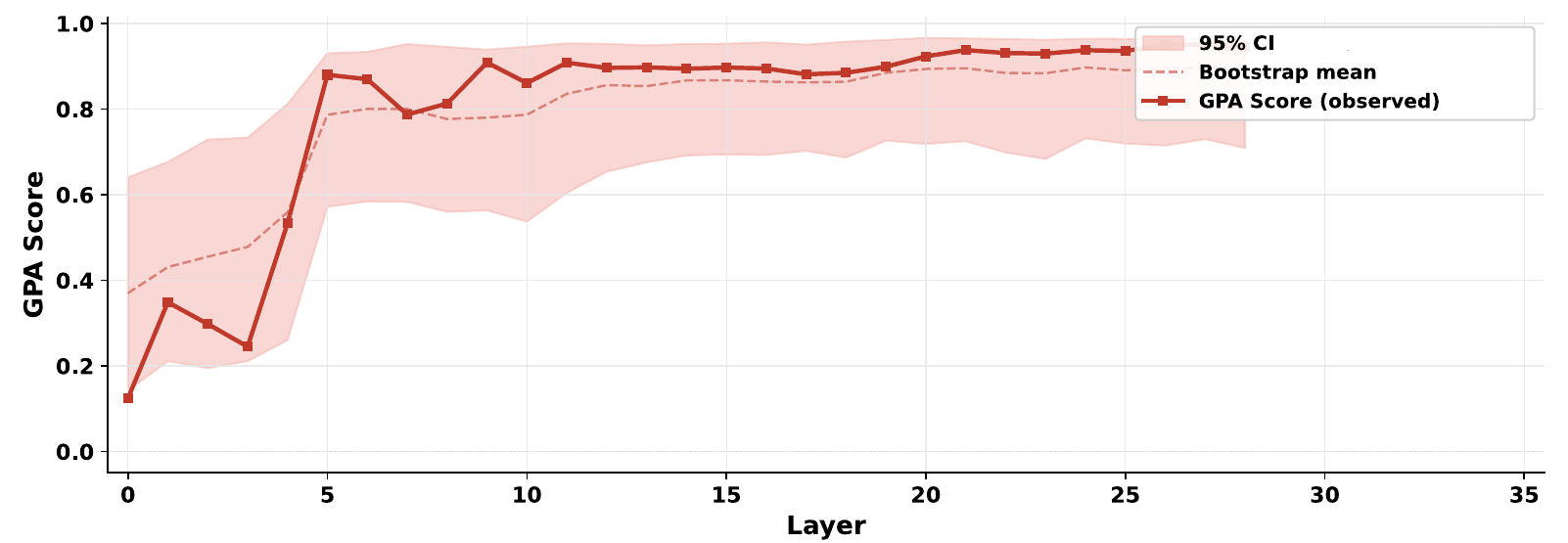}
    \end{tabular}

    \vspace{0.3em}
    \captionof{figure}{Bootstrap confidence intervals for emotion in Gemma-7B, showing layer-wise RSA (left) and GPA (right) profiles with 95\% confidence bands.}
    \label{fig:appendix_emo_bootstrap}
\end{center}

\begin{center}
    \setlength{\tabcolsep}{4pt}
    \begin{tabular}{cc}
        \includegraphics[width=0.45\textwidth,height=0.20\textwidth]{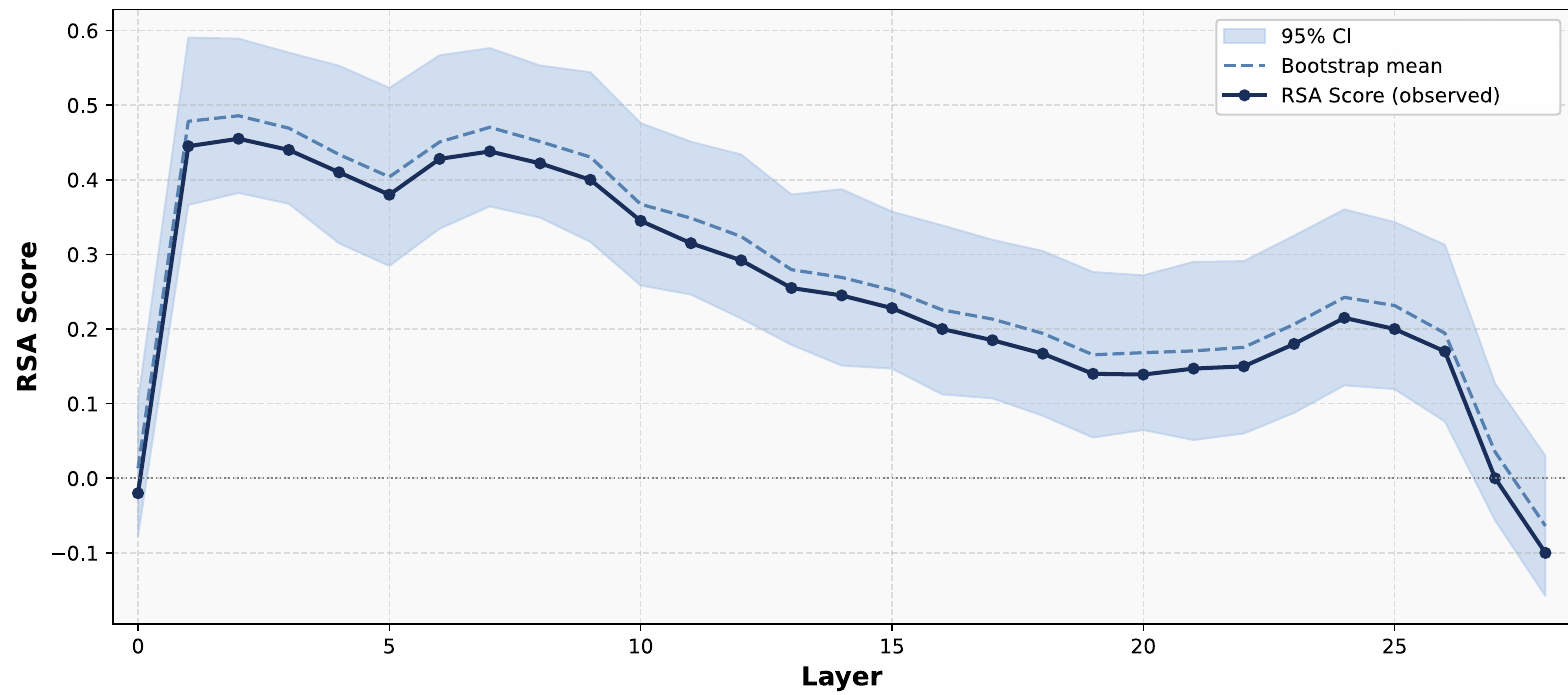} &
        \includegraphics[width=0.45\textwidth,height=0.20\textwidth]{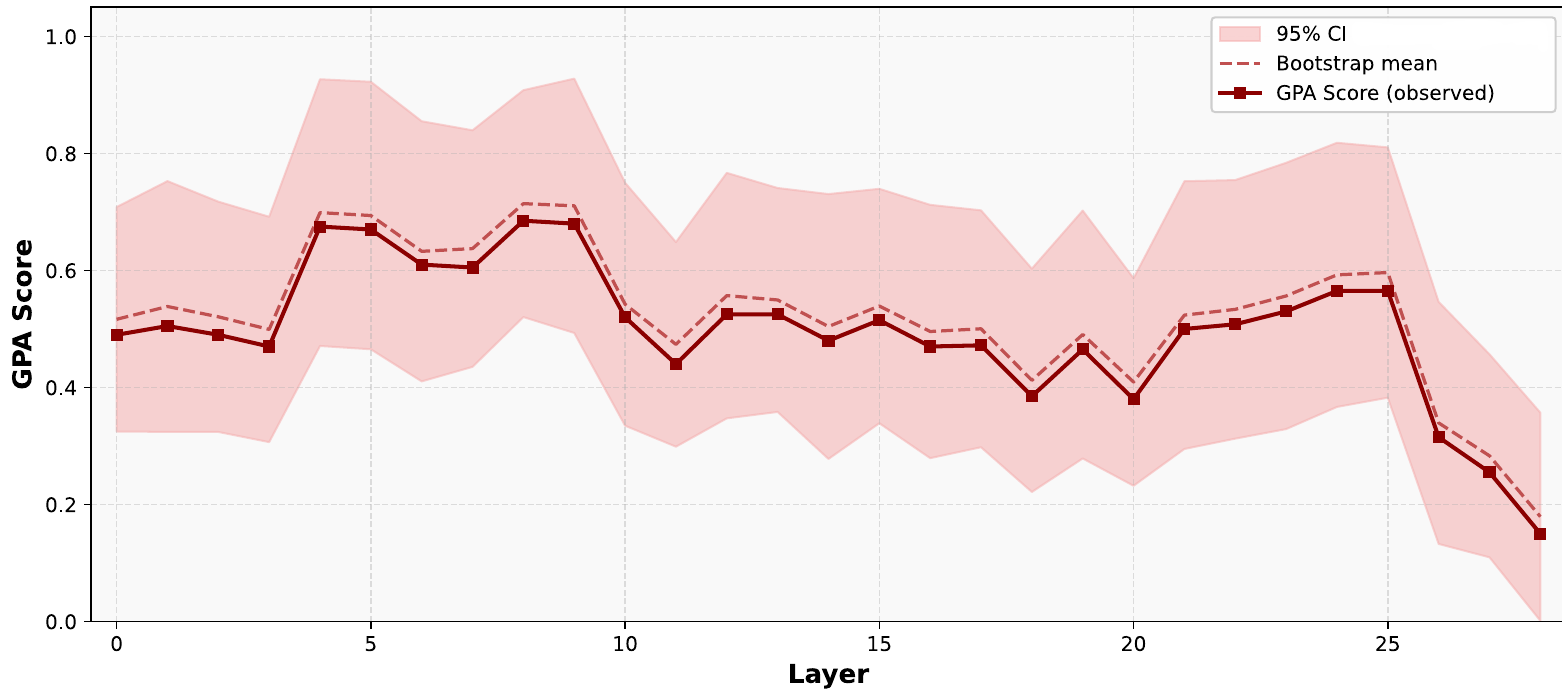}
    \end{tabular}

    \vspace{0.3em}
    \captionof{figure}{Bootstrap confidence intervals for taste in Gemma-7B, showing layer-wise RSA (left) and GPA (right) profiles with 95\% confidence bands.}
    \label{fig:appendix_taste_bootstrap}
\end{center}

\end{document}